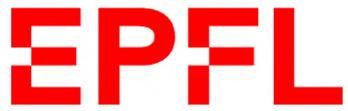

# Neural networks for semantic segmentation of historical city maps

*Cross-cultural performance and the impact of figurative diversity*

Rémi Petitpierre [remi.petitpierre@epfl.ch]

École Polytechnique Fédérale de Lausanne (EPFL), Switzerland

Digital Humanities Laboratory (DHLAB)

2020

# Table of contents

# Abstract

In this work, we present a new semantic segmentation model for historical city maps that surpasses the state of the art in terms of flexibility and performance. Research in automatic map processing is largely focused on homogeneous corpora or even individual maps, leading to inflexible algorithms. Recently, convolutional neural networks have opened new perspectives for the development of more generic tools. Based on two new maps corpora, the first one centered on Paris and the second one gathering cities from all over the world, we propose a method for operationalizing the figuration based on traditional computer vision algorithms that allows large-scale quantitative analysis. In a second step, we propose a semantic segmentation model based on neural networks and implement several improvements. Finally, we analyze the impact of map figuration on segmentation performance and evaluate future ways to improve the representational flexibility of neural networks. To conclude, we show that these networks are able to semantically segment map data of a very large figurative diversity with efficiency.

*Dans ce travail, nous présentons un nouveau modèle de segmentation sémantique des cartes historiques de villes qui surpasse l'état de l'art en termes de flexibilité et de performance. La recherche en traitement automatique des cartes est largement focalisée sur des corpus homogènes, voire sur des cartes individuelles, ce qui mène à des algorithmes peu flexibles. Récemment, les réseaux de neurones convolutifs ont ouvert de nouvelles perspectives pour le développement d'outils plus génériques. En nous basant sur deux nouveaux corpus de cartes, le premier centré sur Paris et le second rassemblant des villes du monde entier, nous proposons une méthode d'opérationnalisation de la figuration basée sur des algorithmes traditionnels de computer vision qui permet de mener des analyses quantitatives à grande échelle. Dans un deuxième temps, nous proposons un modèle de segmentation sémantique basé sur les réseaux de neurones, et y apportons plusieurs améliorations. Finalement, nous analysons l'impact de la figuration cartographique sur les performances de segmentation et évaluons les pistes futures d'amélioration de la flexibilité représentationnelle des réseaux de neurones. Pour conclure, nous démontrons que ces réseaux sont capables de segmenter sémantiquement des données cartographiques d'une très grande diversité figurative avec efficacité.*

# 1 Introduction

## 1.1 Definitions

*The purpose of this section is not to present all the meanings and semantic subtleties of the terms introduced, but only to clarify the use made of them in the frame of this work.*

**Map processing:** A sub-domain of the discipline called document processing, which focuses more specifically on the processing of maps, with the aim of automatically or semi-automatically extracting geographic information of a pictorial or textual nature.

**Convolutional neural network (CNN):** We will use this abbreviated terminology here to refer to fully convolutional deep neural networks, a widely used family of neural networks based on mathematical convolution wherein each mathematical neuron in one layer is linked to all neurons in the next layer, as described first by LeCun and Bengio [1], and adapted by Long et al [2].

**Semantic segmentation:** Process of making an image digitally intelligible by assigning a digital class to each type of object represented in the image, and to the corresponding pixels, according to a pre-defined ontology.

**Figuration:** Pictorial representation of an object. In the present case, the cartographic representation of a physical and geographical object and the way in which this object is depicted.

**RGB:** Red-green-blue, color channels.

## 1.2  Previous works

Among the techniques developed to segment map elements, many algorithms are based on color. Colored maps are particularly easy to segment because the raw data (RGB channels) can be, in some cases, almost linearly separable. According to Muhs [3], in map processing, "the color-based separation of distinct object layers from the map is considered an essential pre-processing task of cartographic image analysis, facilitating the subsequent recognition". Based on [4], he cites as an example [5] and [6], that use color thresholding to segment maps. Other studies add a second, more morphological step, which allows the expansion of areas that belong with a high degree of certainty to a color category, using region growing algorithms, such as recently Chiang and Knoblock [7], and Leyk and Boesch [8], [9], or relying on human feedback [10]. In deep learning, region growing is also used by Oliveira et al [11], who use a watershed algorithm to flood fill and semantize recognized polygons.

A reason why color is widely used to segment and semantize maps is that much of the research focuses on late $19^{th}$ or $20^{th}$ century maps which are usually colorized, such as the United States Geological Survey (USGS) maps, involved in numerous studies [7], [9], [12], [13]. The appearance of color in map printing processes is relatively late. Previously, the maps were generally printed in black only, and wealthier customers could sometimes have them hand painted to their liking. In 1838, Charles Knight patented a revolutionary new and affordable process for printing color maps [14]. From the middle of the $19^{th}$ century onwards, therefore, an increasing proportion of maps became colored. With more or less success in the superimposition of colors, at times approximate, which sometimes led to shifts between the edges of the buildings and the colorization.

In response to the limitations of color, some researchers have turned to segmentation based on texture, which allows to take into account uncolored maps, or even colored maps that are also using texture, on which color-based algorithms generally fail. Hatched areas are particularly targeted [15]–[17]. Other methods, such as the one proposed by Miao et al, focus on texture energy [18]. These approaches have met some success on textured maps [3], [19], [20]. Unlike color, however, textures have several degrees of freedom, including size and rotation. Their use for segmentation therefore requires additional parametrization, or predefined filter libraries.

Other research focuses on primary morphological features of the map, usually lines [18], edges, and sometimes closed polygons [21]. Map processing is generally confronted with the problem of incomplete edges, sometimes due to the figuration (e.g. dashed lines) or reproduction, and sometimes due to post-processing (e.g. Otsu thresholding). A lot of research is thus aimed at reconstructing lines and closing contours [22]–[25], to recover geometric elements of the map. The identification of these structures can also be used for pre-segmentation purposes, such as the guided superpixel method presented by Miao et al [26]. However, geometric reconstruction is very sensitive to information overlay, the latter being the norm in cartography. Methods had therefore to be developed to detect and eventually remove every disruption, such as the map grid [27], the background texture [18], the text [28], or the symbols [29]. The detection of these interfering elements generally requires a precise knowledge of their nature and their visual characteristics: size, shape, texture, color, etc [30]. However, their identification may be useful for other analyses of the map, such as optical character recognition (OCR). These approaches based on edges and geometric shapes recognition can be understood as a search for semantization of the geometry, but not of the mapped elements. Indeed, one of the problems of semantization is the ambiguity of the maps themselves. A rectangle can represent many different things, depending on the context: a building, a courtyard, a basin, etc. When possible and unambiguous, the morphological indicators can be used along with color for semantization [31]. Otherwise, the work of contextualization and in some cases of morphological hierarchization must be carried out by a human, or by strict and non-generic criteria [21], in order to be able to interpret the extracted geometries.

A very different approach aims to match recent raster or vectorized images with older maps, and to use the contemporary data as a reference for semanticizing the historical map as theorized for instance by Chiang [32]. In the past, figuratively simple maps have themselves been used as reference data for vectorizing aerial images [33]. Many shape recognition methods can be used to combine both data sources. Shaw and Bajcsy [34] use Hu's moments, for instance. The conflation can also be used to rectify the extracted data. In their work, Chen et al [35] present a method for automatic conflation of aerial images on vectorized maps, and extraction of road intersections, that works for regular urban forms. However, as promising as the use of correct contemporary data may seem, semantic segmentation based on recent ancillary data may do poorly when working with data prior to the 20$^{th}$ century, as shown by Uhl et al [36].

Early attempts to use neural networks for map processing have been reported by Ignjatic et al [37]. As early as 1996, Chen et al [38] used a simple Hopfield neural network [39] for rotated Chinese characters recognition. However, the parcel extraction was still done with traditional computer vision algorithms. A few years later Katona and Hudra [24] also used a back-propagation neural network on a similar problem of separated symbols recognition. This use of deep learning for symbol recognition and optical character recognition (OCR) has continued, as this technology is still used in more recent studies, which include the latest advances in the field [40], [41]. More recently, fully convolutional networks (FCN) developed by Long and Shelhamer [2] finally opened up new perspectives to the resolution of semantic segmentation problems. In 2017, Liu et al [42] were using their technology to segment house plans. On this basis, Ignjatic et al [37] suggested to apply a similar approach to segment map elements. The following year, Oliveira et al [43] will propose a generic tool for semantic segmentation of documents. They presented a successful application of this tool on maps in 2019 [11], by using a FCN to recognize the contours of parcels of the Napoleonic cadaster as well as their identification number. They achieved a precision of 0.557 and recall of 0.944, with more than 50% intersection over union (IoU) threshold, on extracting parcels from this figuratively homogenous dataset.

With the advent of deep learning, other methods have also been tested and published very recently, during winter and spring 2020. For instance, Uhl et al [36] proposed to use an encoder to classify map patches representing human settlements. The result was then be extended for semantic segmentation using a sliding window. On a 2-class problem, for figuratively homogeneous data from the USGS, they achieved a mean intersection over union (mIoU) of 0.42. Further results have been obtained by Chiang et [44], who managed to extract the railway network from some USGS maps, using a FCN with a modified PSPNet (pyramid scene parsing network) as encoder. On this 2-class problem and for figuratively homogeneous data, they achieved an IoU of 0.622. However, linear data, such as railroads, are considered more difficult to segment than areal data, such as human settlements. On a slightly different problem, namely the recognition of road intersections, Saeedimoghaddam and Stepinski [45] were able to match, or even exceed under certain conditions, the traditional computer vision state-of-the-art algorithms, using a region-based deep CNN.

The FCN have also demonstrated excellent performance in other fields, related to map processing. One of them is the semantic segmentation of satellite images [46]–[48]. Specific architectures have been developed for addressing the specificities of aerial images, such as spatial information loss and lack of context for very high resolution images [49]–[51]. New ways of calculating the loss have also been proposed, such as Liu's edge loss reinforced network [52].

## 1.3 Present challenges and potential applications

The creation of large digital databases on urban development is a strategic challenge, which could lead to new discoveries in urban planning, environmental sciences, sociology, economics, and in a considerable number of scientific and social fields. Digital geohistorical data can also be used and valued by cultural institutions, for instance in the form of a 3D model, as proposed by R. G. Laycock et al [53], and S. D. Laycock et al [54]. Finally, these historical data could also be studied to better understand and optimize the construction of new infrastructures in cities nowadays, and provide humanities scientists with accurate variables that are essential to simulate and analyze urban ecosystems [55]. For instance, comprehending the impact that extreme weather events and climate change might have on our cities could benefit greatly from big geodata of the past. As an example, Penny et al [56] recently demonstrated that at the end of the 13th century, the water distribution system of Angkor, one of the largest cities in the world at the time, had become so complex and so wide that large parts of it collapsed in the face of climatic events. The size of the network had made it fragile, exposing it to a phenomenon of cascading mechanical failure. Today, cities are getting bigger than ever, and even if we understand fluid mechanics somehow better, yet we still lack decision and simulation tools to understand the impact of cities on land, people, and environment. This is why we consider the mining of geodata from the past to be so essential. Segmentation and vectorization are crucial steps in this process, allowing to make the geometric geodata computationally usable and understandable.

When the data are of good quality, and when large homogeneous corpora are processed, it is possible to obtain very good segmentation results with traditional computer vision and decision algorithms, as in the works from Muhs et al [57], [58]. However, these algorithms are far from being generic. They are based in fact on a detailed knowledge of the treated map and its figuration. Consequently, they cannot be used on maps using a different figurative grammar than the one for which they were developed [59]. This implies that the industrial automation potential of these methods lies only in large and very homogeneous cartographic corpora. It was not until nationalism and the end of the 19$^{th}$ century that corpora of this size emerged, such as Cassini's map of France. At the level of the city, the first coordinated efforts of standardized surveys and uniformized representation accompanied the conquest of Europe by the French Empire, resulting in the Napoleonic cadaster. Aside from those corpora, hopes raised by traditional computer vision algorithms are low.

When the cartographic figuration grammar is close enough, it would be possible to adapt the parameters of some pre-existing algorithm. However, this pre-existing algorithm still needs to be known. Indeed, the specificity of such traditional methods caused many different algorithms to be developed, but all are aimed at a restricted academic audience. Moreover, the work required to obtain the code, to understand it, to compile it, to understand the various parameters and to adapt them to the map of interest, as well as the work of manually correcting the results make these tools unsuitable for use by most geoscientists and social scientists. Consequently, map vectorization is still largely manual, despite being extremely time-consuming. Indeed, the process of manual vectorization of a map can take several days or weeks. As a conclusion, we consider that semi-automation with traditional computer vision methods does not allow the quantitative leap expected for the creation of geohistorical big-databases. Large-scale data production would require a major productivity gain of several orders of magnitude, as well as the development of generic tools. In this context, fully convolutional neural networks could help to solve many problems, and to develop software that can generically process a very large number and diversity of maps.

It remains however to be seen what diversity of map figuration grammars these are capable of covering. Cognitive psychology is using the concept of "representational flexibility" to refer to a similar phenomenon. In 1990, Karmiloff-Smith [60] conducted an experiment in which she asked 4-6 years old and 8-10 years old children to draw a man, a house, and animals. In a second step, she asked them to redraw the same subjects, so that they cannot exist, i.e. imaginary ones. While older children easily adapt to the instructions and manage to modify their second drawing to make it absurd (for example, by adding a pair of eyes), almost half of the younger children are unable to do so. In particular, some changes, such as the insertion of new elements or change of orientation, which are common for 8-10 years old, hardly ever occur for younger children. Other changes conversely seem to be similarly mastered at both ages. Karmiloff-Smith concludes the following:

> "Development appears to involve reiterated cycles of representational change, from the simple running of automatized procedures, to redescriptions of internal representations specified as a sequentially fixed list, and then to internal representations specified as a structured yet flexibly ordered set of features, that is, a manipulable concept." [60, p. 79]

In a way, a similar learning challenge applies to neural networks. From rote learning of training examples, i.e. overfitting, the latter learn to recognize spatially rearranged, or modified versions of the information. In the end, the question is what types and amounts of changes neural networks are capable of apprehending in the maps presented to them and whether this will be sufficient to actually develop generic tools.

## 1.4 Research questions and objectives

Based on the aforementioned, we formulated one research question and four sub-questions.

1) **How do figuration impact CNNs performance when they are used for semantic segmentation of historical city maps?**
    i. How to measure the figurative diversity in a map corpus?
    ii. How does the representation of specific map elements, such as buildings, streets, water, or crops vary?
    iii. How robust are trained neural networks when facing high figurative variance?
    iv. What avenues might increase their representational flexibility? How can the composition of training set contribute?

We will also use advanced neural network technologies for semantic segmentation, in an attempt to equalize or even beat previous benchmarks. We believe that the use of these technologies in map processing will soon explode, and it is important for the validity of the results to remain at the forefront of research in terms of performance.

In this work we will hypothesize that figuration has three components: color, texture and morphology. This definition seems reasonable to us as long as it allows for the inclusion of the elements that are processed and processable by a neural network. Depending on the situation, a fourth component could be added, that of symbolic representation. However, a neural network is unable to perceive such a level of abstraction because it does not have a general culture beyond the scope of its problem. Let us take an example. In 2013, facetious cartographers hid a drawing of a marmot on the official map of the Aletsch Glacier from the Federal Office of Topography [61]. While a human could use this symbolic reference as insight to identify, between two maps, the one representing an alpine glacier, a neural network is incapable of doing so.

## 1.5    Structure of the thesis

The structure of the rest of the thesis is as follows. We will focus for this study on historical city plans and maps, which are a special kind of cartography. Urban geodata have interesting application potentials, as discussed above, and the objects figured are of a reasonable size, in proportion to their resolution. For this purpose, we will present the constitution of two corpora, the first consisting of a culturally homogeneous corpus of maps, French maps representing Paris, and the second aiming to bring together a heterogeneous assemblage of city maps from all over the world. Secondly, we will constitute a training set that will properly allow us to teach neural networks. Thirdly, we will seek to operationalize map figuration, in an attempt to answer research questions 1.i and 1.ii. Fourthly, we will use a neural network-based model to segment maps. We will present several improvements compared to the state-of-the-art in map processing, in order to achieve superior performance. Fifthly, we will study the comparative performances of the neural network, facing heterogeneity in a corpus, to answer research question 1.iii. Finally, we are going to present the ways of improving map processing networks in terms of flexibility, to answer question 1.iv.

# 2   Corpora

## 2.1   Definitions

**Western:** There are many definitions of the limits of the western world. In this text we will adopt the reduced definition of Huntington [62], which includes Scandinavia, Western and Central Europe, bounded to the east by the Baltic countries, Poland, Slovakia, Hungary and Croatia, as well as four non-european countries: United States, Canada, Australia and New Zealand.

## 2.2   Paris corpus

The Paris corpus was gathered during a 6-month internship at the Bibliothèque nationale de France (*French national library,* BnF). The aim of the internship was to geolocate and georeference 1515 maps of Paris, using automatic methods combining deep learning and automatic geometry recognition. Half of the corpus came from the collections of the BnF, while the other half came from the collections of the Bibliothèque historique de la Ville de Paris (*Historical library of the City of Paris,* BHVP). 90% of the maps had a scale comprised between 1:25'000 and 1:2'000. The BnF maps covered a period from 1760 to 1949, while the BHVP maps covered a period from 1787 to 1994. However, most maps were published between 1800 and 1950. The overall yearly distribution is visible in **Fig. 2.2.1**.

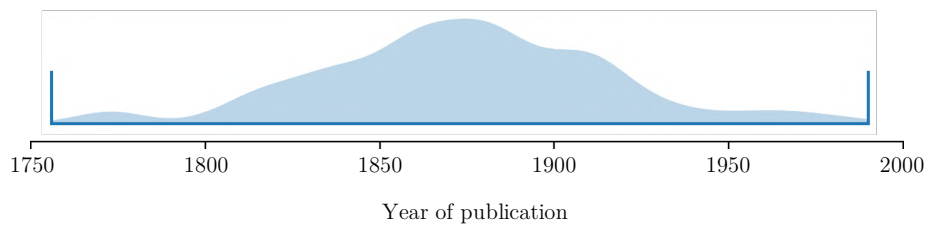

**Figure 2.2.1** Time distribution of the Parisian corpus, by year of publication

As part of this work, 330 1000x1000 pixels map patches were cut out and annotated for semantic segmentation, with a three-color code indicating: 1) the road network; 2) the background (map frame); and 3) the cartographic content excluding the road network. Two-thirds of these patches originated from BnF maps and one-third from BHVP maps. The reason for this repartition was that the cartographic figuration of the BHVP subcorpus seemed to be more homogeneous, since the vast majority of the maps in this subcorpus were published by a restrained number of entities, notably the administration of the city of Paris and the Prefecture of the Seine, often for internal use, e.g. for urban planning purposes. The BnF, for its part, collected maps published, printed or distributed in France, by private publishers or public organisms. However, this does not translate into greater cultural heterogeneity since 98,5% of those maps have been published by French cartographers from the Paris region. Thus, one can consider that the Paris corpus is culturally homogeneous and figuratively diverse. This figurative diversity is visible in **Fig. 2.2.2**. It can be seen that despite the homogeneity of the subject, many challenges persist: differences in scales, codes, figurative grammars, etc. The most arduous aspect of the Parisian corpus undoubtedly remains in the great density of information. Many layers of information are superimposed on the maps, as we will exemplify in the following sections of this chapter, at the same time as for the second corpus.

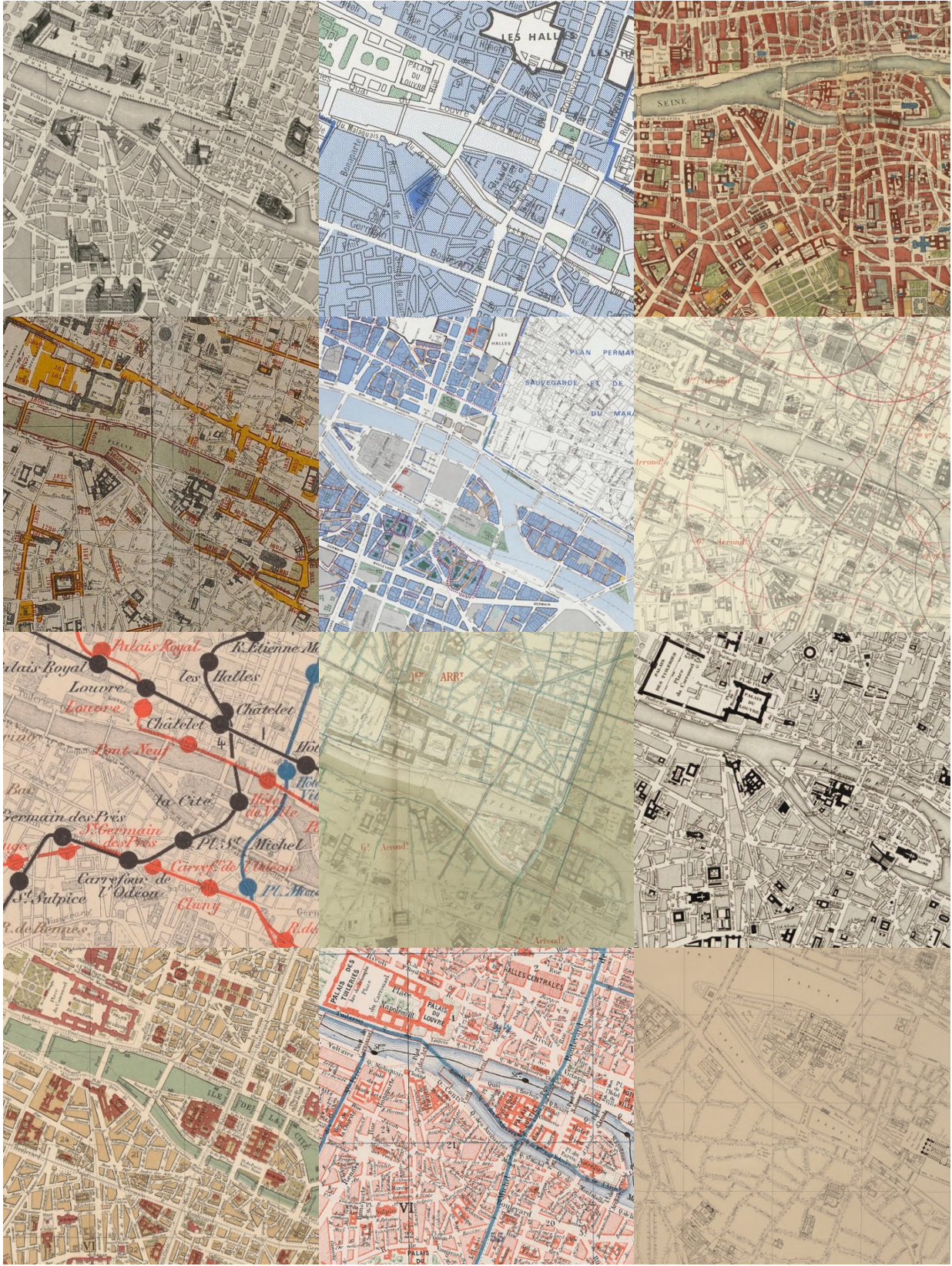

**Fig 2.2.2** Excerpt of maps from the Paris corpus [63]–[74]

## 2.3 World corpus

The second corpus studied in this work is the culturally more heterogenous World corpus. The latter gathers many maps published in different countries and representing cities from all continents. The aim is to obtain a sample, if possible balanced, of urban mapping at the global level. This corpus and its comparative performance with the Parisian corpus are essential to understand the possibility of generalizing learning, and using learning transfer, to solve semantic segmentation problems on maps coming from diverse cultures and representing various cities.

**Methodology to constitute the World corpus**

The constitution of a corpus of maps on a world scale obviously raises many methodological questions. Trade-offs had to be made, although we tried to proceed as methodically as possible. The first stage of our approach relied on literature and cartography manuals. We used three different reference manuals in cartography: *Map cities: Histoires de Cartes*[1] [75], *The City in Maps: urban mapping to 1900* [76], and *Histoire des villes: étude cartographique de l'urbanisme de la Renaissance à la moitié du XXe siècle*[2] [77], all three of which seek to explore and present the history of the urban map. We narrowed the period studied from 1700 to 1950. The first reason for selecting these time boundaries is the geometric accuracy of the maps. Topographic surveys had reached a relatively correct accuracy in the 18th century. On the other hand, the proportion of maps in aerial view became significant. Many older maps preferred an oblique projection, or an illustration of the city from a real or imaginary point of view. On the other hand, from 1950 onwards, cartography underwent profound changes, linked in particular to new chemical printing techniques and then, from 1960, to the digital transition of the field [78], [79][3]. Along with that, in the second part of the 20th century, the development of satellites, as well as the growing number of airplanes, allowed to keep a geographical trace of the past by other means, and we therefore consider that the period 1950-2020 is not a priority for research in historical map processing. Let us also note that the time 1700-1950 overlaps very largely with the period covered by the Parisian corpus, which facilitates the comparison. Only maps in aerial view, or that include only a few illustrations, such as monuments were selected. This also applies to the Paris corpus.

---

[1] Map stories
[2] History of cities: cartographic study of urban planning from the Renaissance to the middle of the 20th century
[3] Cited by [37]

Firstly, we focused on collecting extra-western maps representing Asian, Middle Eastern and African cities. Then Slavic, Latin American and Oceanian cities. Finally, North Americans and Europeans. The goal was to balance the representation of these world regions in the final corpus. Both in cartography textbooks and in the online collections of heritage institutions, Western maps are by far the most numerous. Starting with the rarest maps ensures that the more frequent categories will not be over-represented.

The first step of the corpus harvest therefore began with searching for extra-western maps in cartography reference textbooks and downloading those available online. In a second step, the goal was to complement them with other extra-western maps available online, which were not specifically referenced in the literature but could allow to balance the corpus and complete it. The main access point used for this was OldMapsOnline, which allowed quick visualization of the maps available for a city in several large collections such as the British Library, the Harvard Library or the *Institut Cartogràfic i Geològic de Catalunya* (Cartographic and Geological Institute of Catalonia, ICGC). In a third step, we tried to complete the most obvious "holes" of the corpus, this time with a specific and systematic search directly in the databases of heritage institutions.

Thereafter, western maps were added to the corpus, until a mass of maps similar to extra-western maps was obtained. In general, the examples found in the manuals were sufficient to obtain a balanced representation, even if some under-represented areas or countries have also been completed by a targeted search in the databases.

**Harvested metadata**

Only a few metadata were collected: the title of the map, the year of publication, the heritage institution, the internet address where the map can be downloaded, the city represented and the current country in which it is located, the countries or nationalities of the people involved in the geometric field surveys, in the publication or reproduction of the map. In addition to these data, the metadata of the Paris corpus included the name of the publisher, the language of the map, the subject tags, the format, the address of the record, the catalogue number, and sometimes the scale.

**Quantitative analysis**

In the end, the collected corpus is made up of 256 maps from 32 different databases (**Fig. 2.3.1**). Eight heritage institutions account for 83% of the corpus. Only a few institutions are located outside the West. This does not necessarily reflect a methodological bias because institutions in Western countries are also probably better funded and better equipped to digitize their documents. Consequently, they represent a higher proportion of the maps available online. On the other hand, cartographic production from the colonies were partly repatriated to the mother nation and this contributed to the concentration of maps in Western collections. The corpus maps held by non-Western institutions came from the Historic Cities collection (Israel-Palestine, 7 maps), the Art Research Center of the Ritsumeikan University (Japan, 4 maps), the Colombian National Library (4 maps), the Kalakriti Archives (India, 2 maps), the University of Cape Town (1 map), the Archives from the City of Almaty (Kazakhstan, 1 map), Virtual Shanghai (China, 1 map), and the Arif Hasan Collection (Pakistan, 1 map).

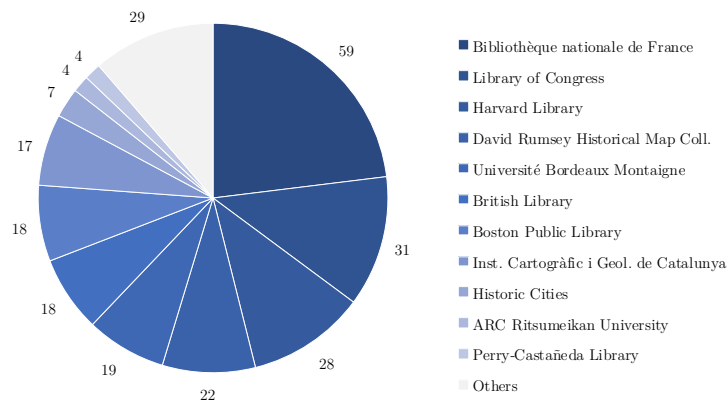

**Figure 2.3.1** Source collection of the maps found in the corpus world

One of the most important collections of digitized historical maps [80, p. 23] does not appear in our corpus: the collection of the Univerzita Karlova (*Charles University*) in Prague. This is due to the fact that their server was visibly down for several weeks when the corpus was gathered, in March 2020.

The 256 maps in the corpus represent 182 different cities in 90 countries around the world (**Fig. 2.3.2**). These cities were classified in 11 different regions (**Fig. 2.3.2**). The most represented countries are the United States (23), China (15), India (13), Japan (10), France (8), Russia (8), Australia (6), Canada (6), Argentina (6), and Italy (6). Four cities are represented by 4 or more maps: Beijing (6), as well as Buenos Aires, Calcutta, and Prague (4 each). 16 other cities are represented by 3 maps.

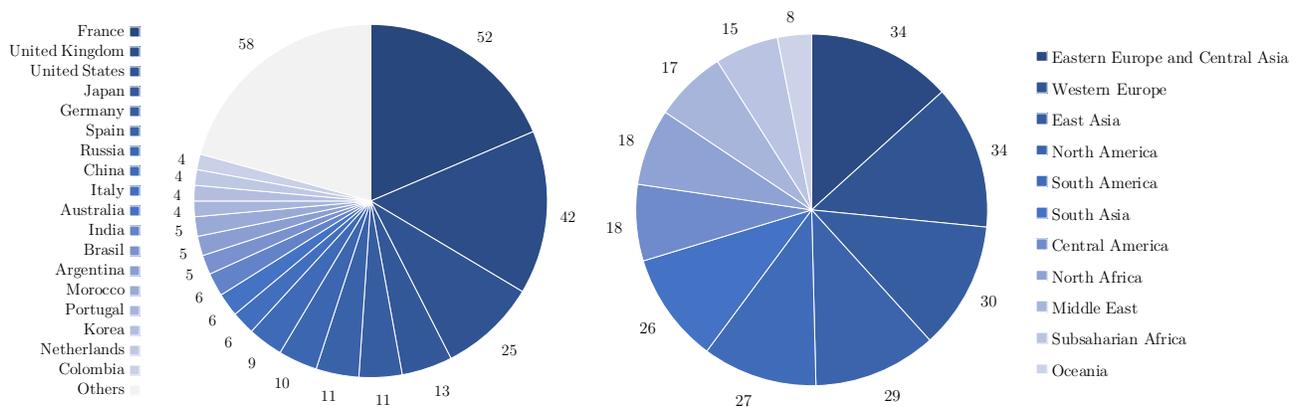

**Figure 2.3.2 (left)** Country of publication of the maps in the corpus world. There may be more than one country of publication for a single map, when the publication is the result of a partnership. **(right)** Distribution of maps by world regions

Three areas remain clearly under-represented in the final corpus (**Fig 2.3.3**). They are sub-Saharan Africa (excluding South Africa), Central Asia and the region Indonesia-Philippines. However, although some of these areas are densely populated today and are home to huge metropolises, their demographic transition is still far from complete and, in many cases, had not yet begun in 1950, i.e. at the later time boundary of the corpus [81]. The population of sub-Saharan Africa, for example, was only 186 millions in 1950, compared to 1.02 billion in 2017 [82]. In addition to this massive population increase, the cities growth is also supported by a strong rural exodus. These data reveal a sub-Saharan Africa which, before 1950, was very little urbanized and exploited mainly for its natural resources. This may reflect low interest in mapping sub-Saharan African cities over the time boundaries covered by the corpus.

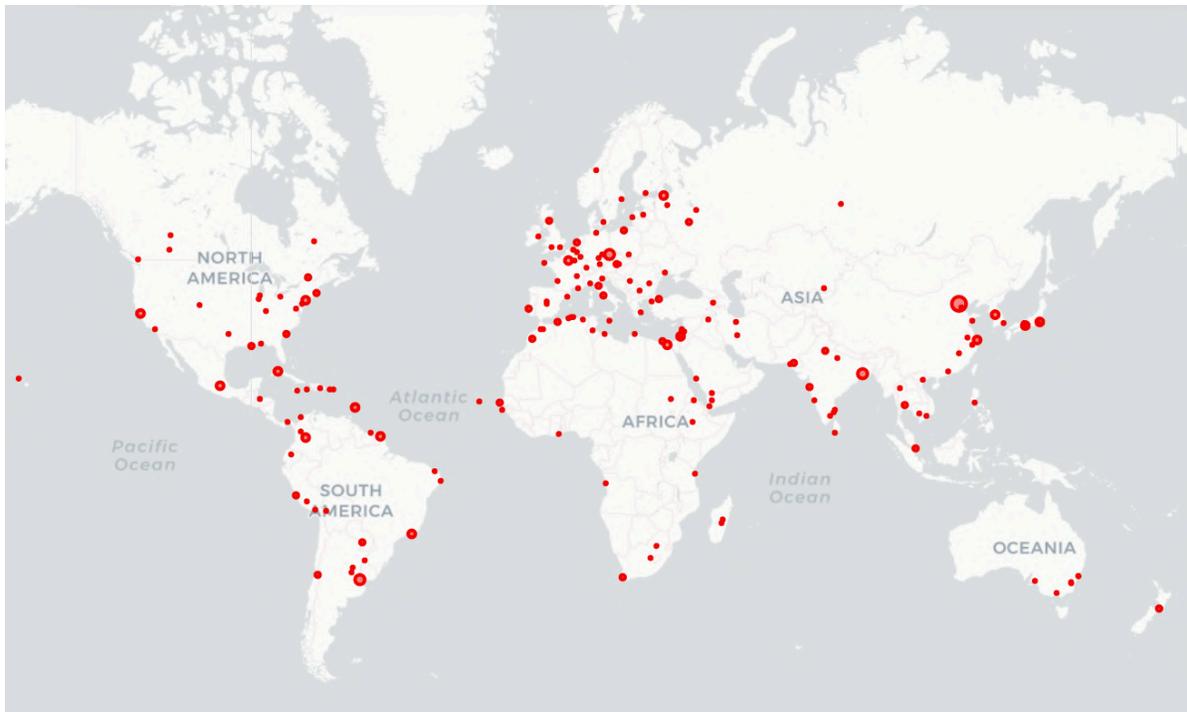

**Figure 2.3.3** Visualization of the corpus on the world map. The radius of the circles represents the number of maps per city

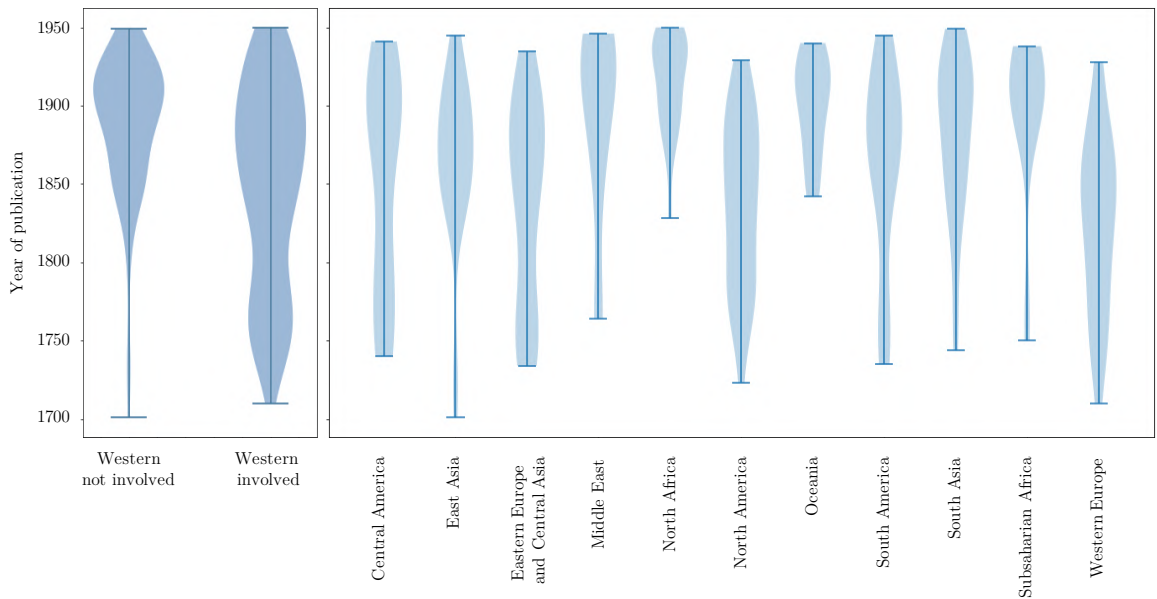

**Figure 2.3.4 (left)** Year of publication, depending on whether or not a Western country was involved in the publication process. **(right)** Year of publication, according to the region of the mapped city

The maps were printed or published in 56 different countries (**Fig. 2.3.2**). However, most of the maps (208) have at least one Western country implicated in its publication. **Fig.**

**2.3.4** shows very clearly that the independent publication of maps by non-Western countries began only timidly, starting in the 19th century. All non-Western regions were relatively synchronous in the date of publication of their first urban map, although it started slightly later in Africa. The publication of Oceanian maps is more recent. This may reflect a small bias in the corpus, due to the modest representation of this region (see **Fig. 2.3.3**).

**Qualitative analysis**

We have seen that the world corpus is very varied. This diversity is also visible individually on the corpus maps. A small number of examples have been grouped in the following pages to illustrate it. In **Fig. 2.3.5**, we notice the differences in scale between the *Plan [...] of Edinburgh* by J. Craig [83] and the *Bangkok town plan* from the War Office [84]. Urban blocks can measure from several hundred pixels to only a few pixels long.

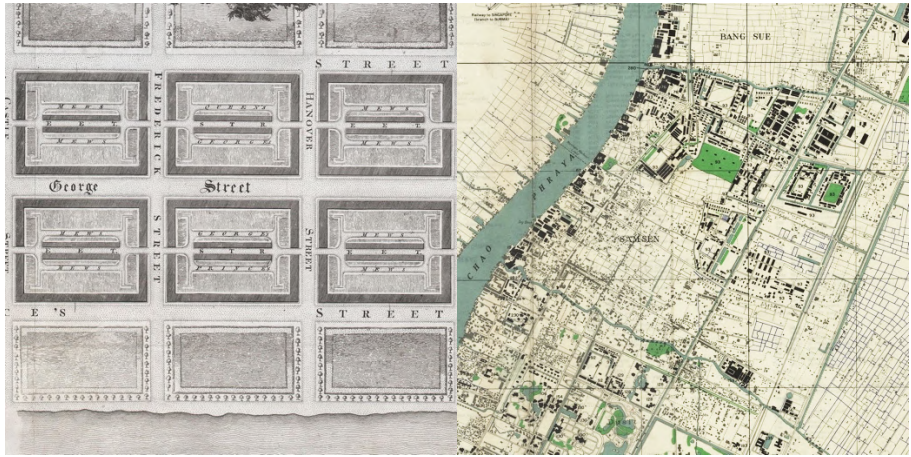

**Figure 2.3.5** Scale differences [83], [84]

This phenomenon can be amplified by the scanning resolution, since the actual scale of the map is usually unknown. These scale differences can be problematic, as can the presence of small objects, which are known to be difficult to detect in CNN semantic segmentation of aerial images, for instance. Indeed, small objects occupy a small area of the image and tend to have a low impact on loss, and thus on gradient and learning. The cross entropy loss that will be used later has the possibility to be adjusted with classes weights, which can partially correct this bias, according to Kampffmeyer et al [46]. Besides, very high resolution, or very large scale images could also be problematic, if one again draws the parallel with satellite imagery [49]–[51].

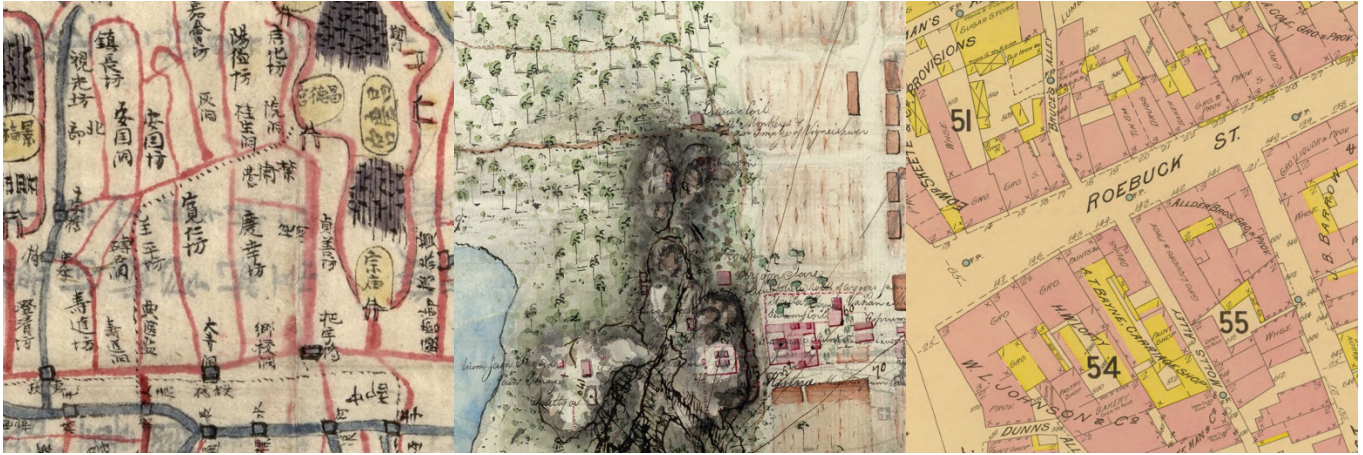

**Figure 2.3.6** Level of formality of the map. From left to right: 1) highly informal [85]; 2) rather informal [86]; 3) highly formal [87]

In the corpus, the figuration of the maps is sometimes not very formal (**Fig. 2.3.6**). In addition to geographic formalism, figurative grammar can also vary greatly across maps and cultures. In **Fig. 2.3.6**, on the left, one can see a map of Beijing that adopts a common Chinese figuration grammar, where the streets are represented as wires. On the contrary, on the right, in an insurance plan of Bridgetown, Barbados, the grammar relies on the use of numerous symbols, colors, crosses and bars to indicate various information about buildings. Every symbolic abstraction is different, but they are all equally valid and have not been a criterion for inclusion or exclusion in the constitution of the corpus. As we are seeking to explore the limits of CNNs, the presence of such maps in the corpus is deliberate.

Monuments are frequently deviating from the established conventions of representation. A particular importance is given to their enhancement, as on the left of **Fig. 2.3.7** in this tourist map of the *Monuments and interesting sights in Recife* [88], or on the right, in the *New Topography of Rome* by Giovanni Battista Nolli, used as a reference by many architects [89]. The figuration of monuments in perspective [90] is also very common in Parisian maps. The main limitation of this is the occlusion of the streets and the difficulty to estimate the building's footprint.

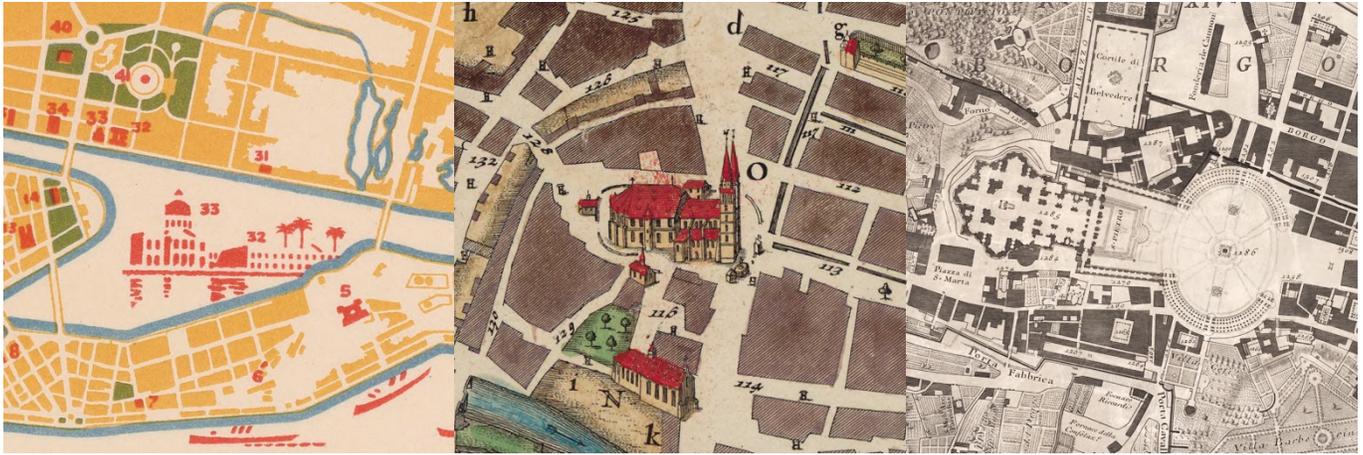

**Figure 2.3.7** Non-trivial figuration of the monuments, from left to right : 1) stylized drawing [88] ; 2) perspective drawing [90] ; 3) representation of the interior architecture [89].

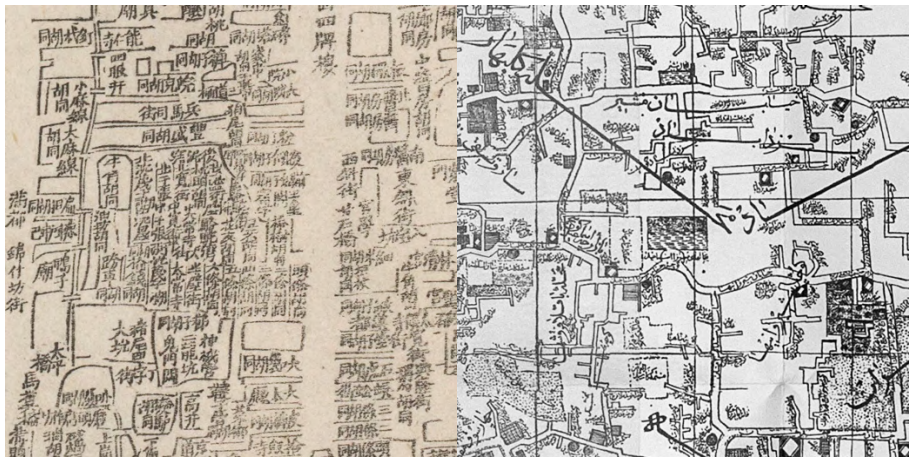

**Figure 2.3.8** Different use of writing. On the left [91], the Chinese characters can be used as a texture, due to their regular shape. On the right [92], names written in the Persian alphabet can be lengthened to occupy a represented space and calligraphic words can be intermingled.

Sometimes, cultural specificities can result in absolutely surprising map grammars, as in **Fig. 2.3.8**. On the left [91], the Chinese characters are used as a texture, due to their regular and rectangular shape. This case raises interesting questions about map processing methods that decouple semantic segmentation and OCR, including our own. On the right [92], names written in the Persian alphabet, an Arabic-inspired alphabet, can be lengthened to occupy a represented space, such as a neighborhood. Moreover, here also, calligraphic words can also be intermingled to a specific shape. As in Arabic culture, calligraphy is prized in Persian culture and the resulting interweaving of words could make the recognition of text areas more difficult for CNNs.

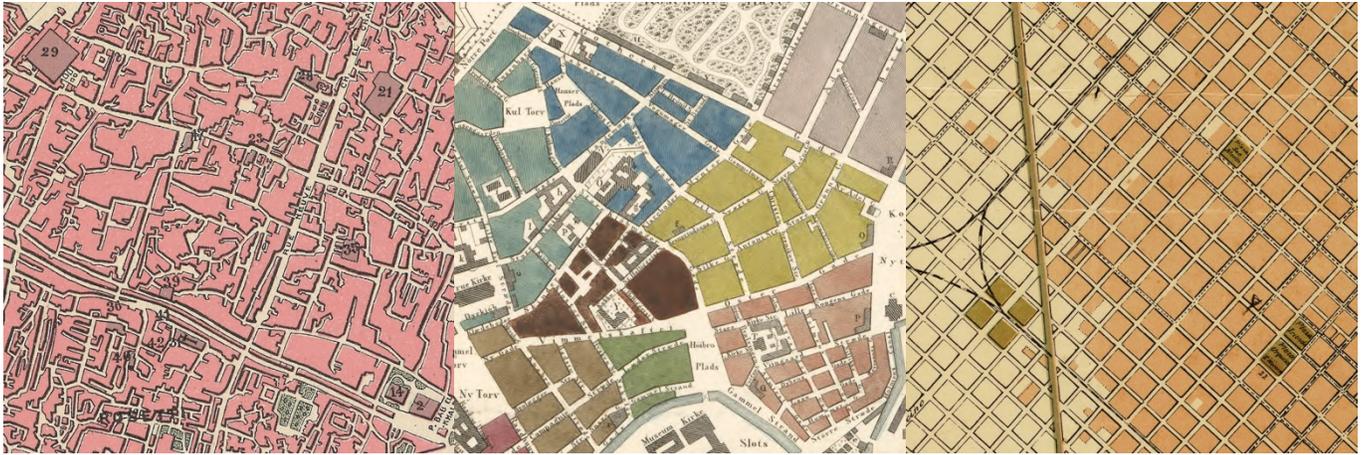

**Figure 2.3.9** Regularity of urban form. From left to right: 1) highly irregular [93]; 2) mixed [94]; 3) highly regular [95]

Cultural differences are also visible in the urban planning of the cities themselves. In **Fig. 2.3.9**, one can see on the left a map of Cairo, typical of Islamic urban planning, organized in a labyrinthine form, which makes the city easier to defend against potential assailants, with only a few wider streets that allow for circulation. In the centre is a map of Copenhagen showing a more typically European urban form. On the right finally the city of Rosario, Argentina, built according to a regular plan. All the maps in the world corpus were classified according to these three categories: regular, irregular, and mixed urban form: 95 were classified as regular, 68 as irregular, and 93 as mixed. For the Parisian corpus, the urban form is considered to be mostly mixed.

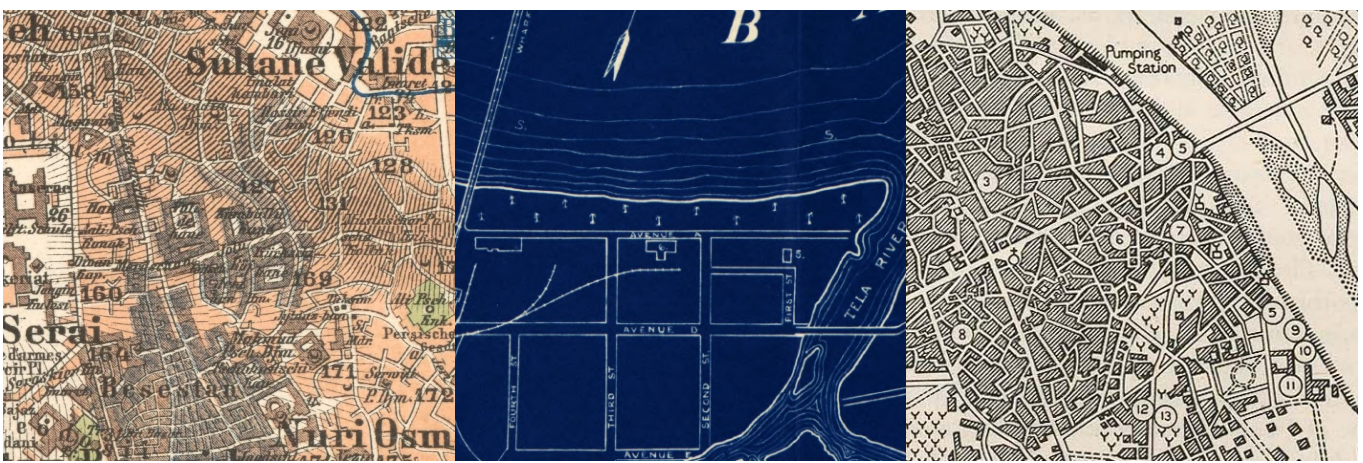

**Figure 2.3.10** Figurative specificities. On the from left to right: 1) field shadowing [96]; 2) blueprint-style map [97], 3) overlapping circles [98]

In addition to cultural differences, the maps in the corpus also present many additional difficulties of their own for the CNNs. In **Fig. 2.3.10**, one can see a sample of major complications for computer vision problems. On the left, field shadowing produces noise that can hamper object recognition. In the center, color inversion is confusing for edges recognition. On the right, the number circles are occluding the map. Occlusion is a difficult computer vision problem that is far from being solved.

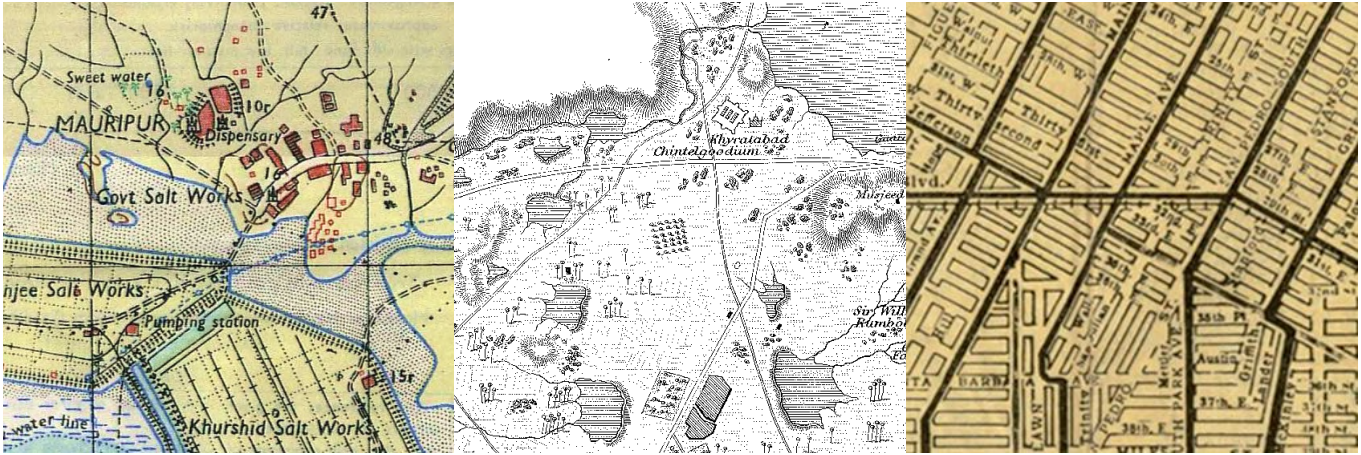

**Figure 2.3.11** Digitization imperfections. From left to right: 1) poor color scanning quality and brightness jumps [99]; 2) loss of information due to image binarization [100]; 3) low resolution [101].

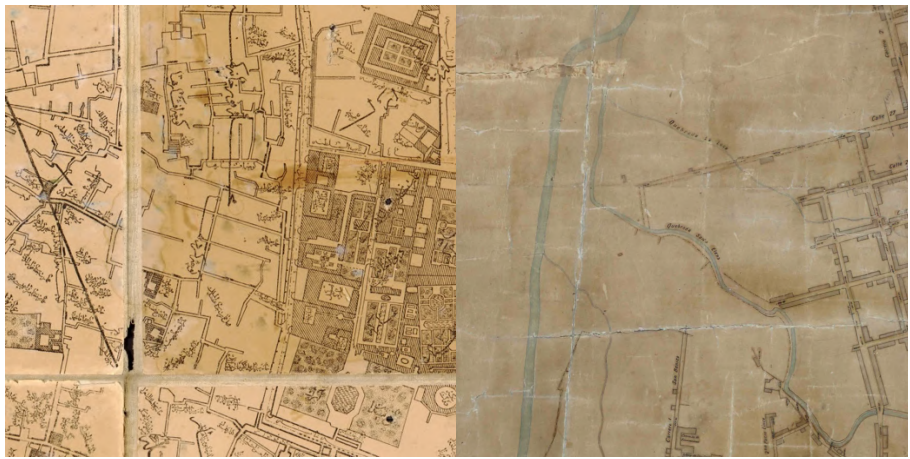

**Figure 2.3.12** Damaged maps or alterations due to conditioning. On left [102] the paper is damaged by dampness, light toning, stains and fabric canvas, on the right [103] by tears and visible traces of intervention at the top left of the image

The particular difficulty of the corpus also stems from the differences in digitization quality, which varies a lot (**Fig. 2.3.11**) and can produce digitization artifacts, loss of information, or low-resolution images. Resource disparities between heritage institutions are also visible on the state of conservation of the map itself, as in **Fig. 2.3.12**, where maps in a poor condition can be seen. All these elements can impact the clarity of the image and could significantly complicate the segmentation problem for the neural network.

The few examples presented in this section allow one to see both the extreme difficulty of the corpus, but also its richness, diversity and beauty. The map is also an aesthetic object, prized by museums and heritage institutions. The extraction of the information they contain also allows to better understand and enhance the value of these digitized documents.

## 2.4 Training sets

Neural networks need training examples that they can learn from. The training examples are like small memory cards with the question on one side and the answer (ground-truth) on the other. In this work, the training sets consisted of patches randomly cut out of the maps. For the Parisian corpus, 330 patches were cut from 330 maps randomly selected from the whole corpus. The position of the patch on the map is also totally randomized. For the world corpus, which contains only 256 maps, 256 patches, one per map, have been cut to make up the training set. Then, 49 additional patches were randomly cut from 49 randomly drawn maps.

For each corpus, the patches were divided into a training set and a validation set. This ensures that the neural network does not simply memorize the solution from the training examples, but that it is able to generalize and give a correct answer when presented with a map patch that has not been used for training. For the Paris corpus, in general, 300 patches of 1000x1000 pixels were used for training and 30 for validation. For the world corpus, 256 1000x1000 pixels patches were used for training and 49 for validation. This slight difference in repartition is due to the greater subject diversity of the world corpus, which required a slightly larger test set to ensure a valid evaluation.

As explained, each training sample is composed of an image (the question) and a label (the answer). The label comes from the attribution of each pixel to a class. This process is called annotation. Each semantic class corresponds to a type of object found in the images. Together, the classes form an ontology, which gives meaning to the map data. The annotation work was done iteratively, first annotating 3 classes, then expanding to 4, then 5 classes. This allowed the data from the first iteration to be exploited for many experiments while the ontology was being extended to include the last two classes, and thus make better use of the time allotted for this work. The total annotation time is estimated to 8 weeks, of which 4 weeks were done in the frame of this master project. The patches were annotated using Adobe Photoshop CS6 software and saved in the raster lossless PNG format.

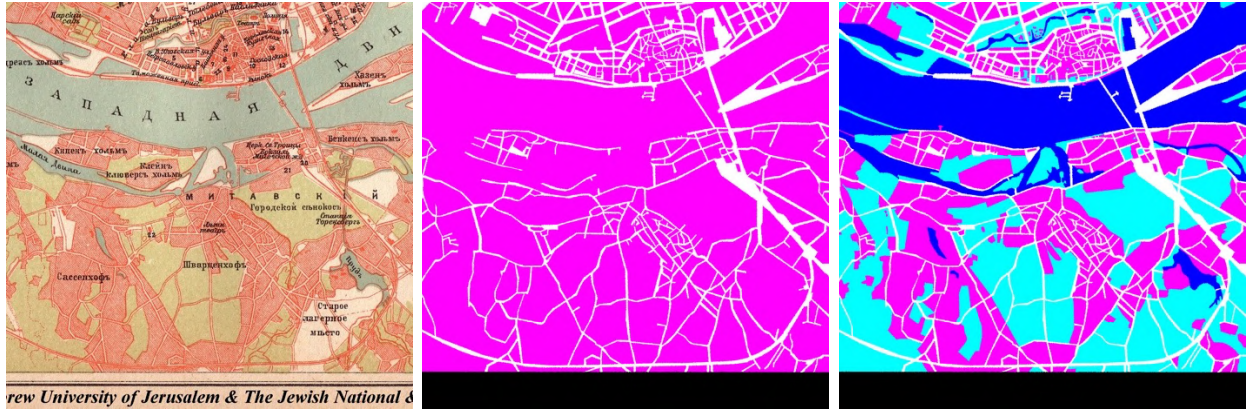

**Figure 2.4.1** From left to right: 1) Patch of a plan of Riga, 1889 [104]; 2) 3-classes label from the left map patch (black: frame, white: road network, magenta: map content); 3) 5-classes label from the left map patch (black: frame, white: road network, magenta: blocks, blue: water, cyan: non-buillt)

The initial ontology consisted of 3 classes (**Fig. 2.4.1**, center):

1) The frame (black), i.e. the non-mapping part of the document. It contains the frame as well as the illustrations, illuminations, surrounding texts and medallions;
2) The road network (white). This class includes roads and streets as well as railroads and bridges;
3) The map content (magenta), i.e. all the cartographic content of the map, which does not belong to the road network.

This ontology was then extended to 5 classes (**Fig. 2.4.1**, right), keeping the two first classes unchanged and separating the 3$^{rd}$ class into 3 new classes:

3) Blocks (magenta), city blocks, detached houses, walls or anything similar to buildings;
4) Water (blue), including rivers, canals, bodies of water, reservoirs, lakes, and sea;

5) Non-built (cyan), which in fact includes all non-aquatic unbuilt land, including wasteland, meadows, crops, forests, but also urban unbuilt land, such as parks, enclosed squares, and inner courtyards.

These 5 classes that will be called frame, road network, blocks, water, and non-built, to simplify, allow to include all the elements of the map, in theory. The proposed classification assumes that the map represents the environment at an instantaneous moment, and on a single terrain layer. This assumption is due to the segmentation model chosen (classification), which allows each pixel to be assigned only one class. This assumption can be considered reasonable, since we were analyzing a single ink thickness. However, it is sometimes infringed.

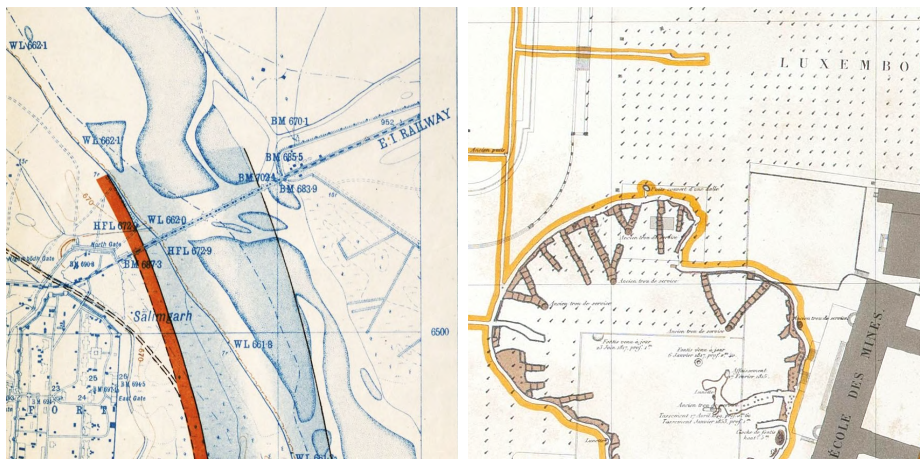

**Figure 2.4.2 (left)** Patch of a map of Delhi, the Survey of India Office, 1913 [105]. **(right)** Patch of the underground Atlas of the city of Paris ordered by the Baron G. E. Haussmann, sheet 6, 1859 [106].

Let us take an example, where the immediacy of the moment is not respected. The map in **Fig. 2.4.2 (left)** shows an area of the Jumna River in Delhi. It is marked *HFL 672·9*, which corresponds to the High Flood Level of 5th August 1908. The darker blue area corresponds to the part of the river "which generally contain water" [105]. These measurements were from the morning of the 26th April 1912. Two temporally disjoint pieces of information were therefore superimposed on the map and a subjective interpretation of the "correct" class is therefore necessary. This map patch also contains a second example of disruption: the great red boulevard which is superimposed on the rest if the map information. This boulevard, which was planned as part of the construction of the Imperial

Delhi, the planned new capital city of British India, did not exist at the time the map was made. While this major planning work provided a convenient historical opportunity to map the territory, the resulting cartographic representation is temporally ambiguous. In this case, this information is particularly confusing because the road section represented will in fact never be built, probably abandoned in a later version of the project. Thus, what is represented on this map is a possible future that never existed…

Other times, the principle of representing a single layer of terrain is not respected. This is the case of the underground Atlas of the city of Paris [106], (**Fig. 2.4.2**, **right**). The tunnels of the underground quarries in the south of the city, some of which were transformed into ossuaries, or catacombs, from the 18$^{th}$ century onwards, were also superimposed on the surface elements. Other examples of common multilayer representation include metro lines or sewerage networks.

If urban mapping contains so many of these examples, it is because it is closely linked to the idea of spatial planning and used as a tool for this purpose. For the urban map additionally, the represented environment is, by its very nature, a vertical structuring of space, where underground pipelines, overhead networks and multi-storey buildings are intermingled, whereas the map seeks to project the environment on a horizontal plane, leading to the hybridization of these dimensions.

**Table 2.4.3** Proportion of areas covered in the different corpora and sets by the 5 annotated classes.

| Corpus | Set | Frame | Road netw. | Blocks | Water | Non-built |
|---|---|---|---|---|---|---|
| World | train | 0.2219 | 0.1226 | 0.1749 | 0.1212 | 0.3594 |
| World | eval | 0.1851 | 0.1140 | 0.1783 | 0.0728 | 0.4499 |
| Paris | train | 0.2358 | 0.2079 | 0.3381 | 0.0262 | 0.1919 |
| Paris | eval | 0.1659 | 0.2036 | 0.3601 | 0.0448 | 0.2256 |

**Table 2.4.4** Proportion of areas covered in the different corpora and sets, in the 3-classes ontology.

| Corpus | Set | Frame | Road netw. | Map content |
|---|---|---|---|---|
| World | train | 0.2258 | 0.1234 | 0.6507 |
| World | eval | 0.1806 | 0.2041 | 0.7009 |
| Paris | train | 0.2358 | 0.2056 | 0.5586 |
| Paris | eval | 0.1659 | 0.2014 | 0.6328 |

Looking at the proportion of the area occupied by the different classes in **Tab. 2.4.3**, it can be seen that the defined ontology allows a relatively balanced distribution. This helps to contain disparities between classes during neural network training and generally prevents the need to employ subterfuges, such as adding weight coefficients when calculating loss or doing post-processing. Logically, the intermediate 3-classes ontology (**Tab. 2.4.4**) was more inequitable, as some classes of the complete ontology were still grouped at the time. The slight differences in frame and road network classes between **Tab. 2.4.3** and **Tab. 2.4.4** are due to minor corrections made during the extension to 5 classes.

The Parisian corpus has a much higher areal proportion of road network compared to the world corpus. This difference in surface area may partly be due to the Haussmanian works and the successive street widening works that took place in Paris during the 19$^{th}$ century. Moreover, the Parisian maps depict a more urban environment compared to the other maps. Indeed, one can also notice in **Tab. 2.4.3** that blocks and road networks occupy a larger share of the Parisian maps, while non-built occupies a larger share of the world corpus maps.

There are several reasons for this. First, over the period of study, Paris is one of the largest and densest cities in the world. Many maps in the world corpus portray colonial cities, which were, especially in the early 18$^{th}$ century, still very modest in size. The countryside was very close to the city and appears sometimes on the map. In addition, the use of the urban map varies. Many maps from the Parisian corpus had an administrative and planning purpose. In particular the maps from the BHVP collections, which were generally drawn up by the departments of the city of Paris. The tourist map also appeared very early in Paris, probably earlier than in most other cities, even capitals. Indeed, between 1855 and 1947, Paris hosted no less than 8 universal and specialized exhibitions, all attracting what were the first crowds of visitors. 27 maps of the corpus are therefore devoted to universal exhibitions, according to the subjects catalogued by the BnF, and 63 additional maps bear the explicit mention "guide" in their title. Both the administrative and tourist maps only aim to represent the urban landscape as such, which explains its preponderance on the map.

Conversely, world corpus maps, which are regularly colonial maps, as we have seen in **Fig. 2.3.2**, often aim to represent, in addition to the city, surrounding resources such as arable land, forests, and quarries. They may also have been drawn up for military and strategic reasons, to defend areas that were part of a colonial empire. An example is the *Plan of the* *city of New York in North America: surveyed in the years 1766 & 1767* ([107], **Fig. 2.4.5**), modestly dedicated to "His Excellency Sir Henry Moore, Baron, Captain General and Governor in Chief In and Over His Majesty's Province of New York and the Territories depending thereon in America". In the title alone of this map, published 8 years before the American War of Independence, one can see the stakes of imperialism and the struggle for physical and symbolic control of the territory in which cartography was engaged. Although this map of the then Little Apple is truly an urban map, the city occupies only a very small area, at the very tip of the island of Manhattan. Conversely, land and sea access routes, the Hudson River estuary, nearby resources and agricultural land are widely represented. This map is a good illustration of the observed high representation of non-built and water in the corpus. Indeed, water is also more present (**Tab. 2.4.3**), probably due to the numerous maps of coastal cities that are part of it (**Fig. 2.3.3**).

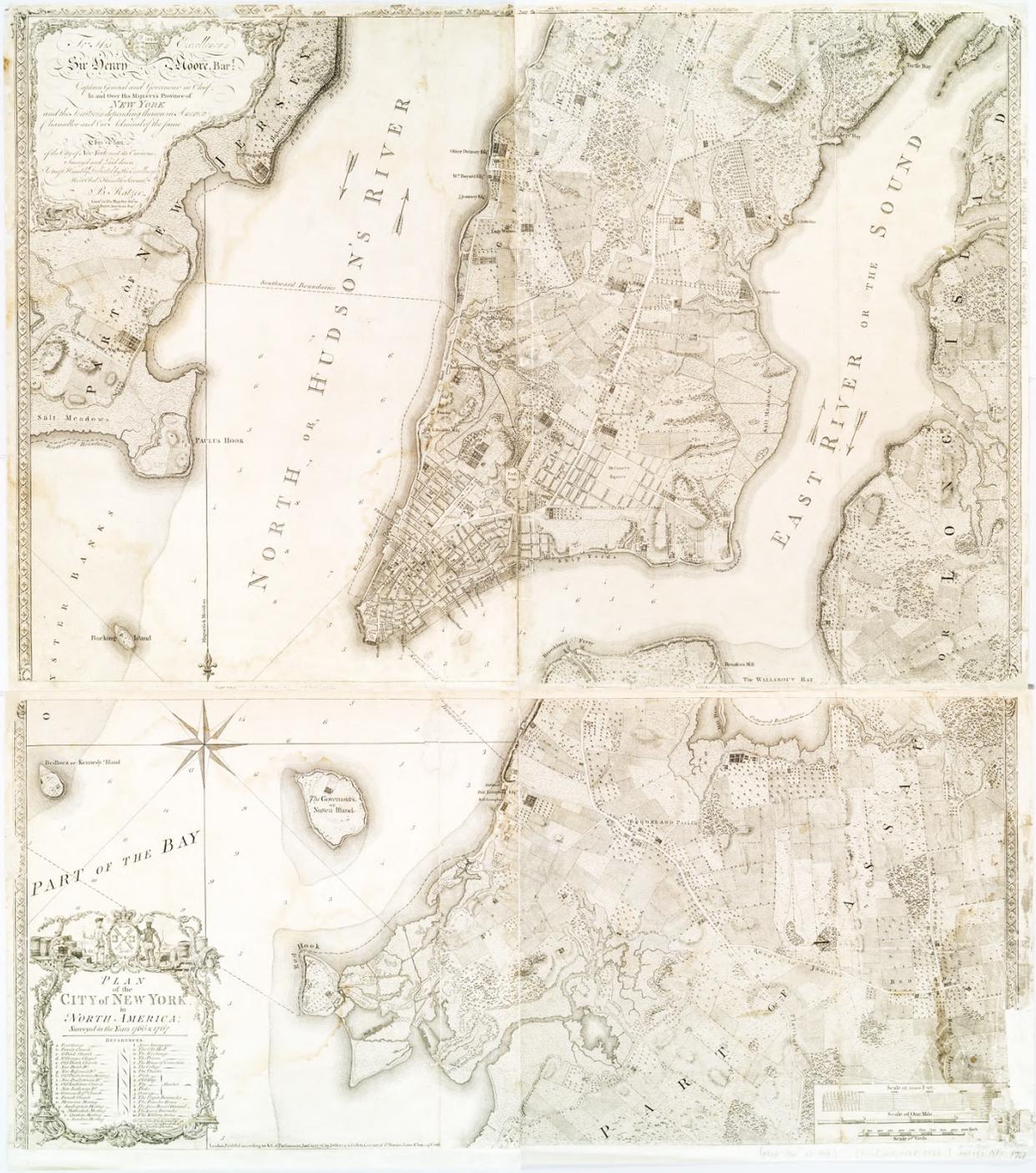

**Figure 2.4.5** Plan of the City of New York in North America, surveyed in the years 1766 & 1767. London. Published according to Act of Parliament [...] by Jefferys & Faden [107].

## 2.5 Pre-processing

To be trained, neural network should ideally receive patches of constant size, here 1000x1000 pixels. Four maps of the world corpus being smaller than this patch size, they were preprocessed with a CNN-based super-resolution algorithm. Such CNNs allows to artificially augment the resolution of an image. A generic software, named Bigjpg [108], [109], was used, setting the parameters to use the network trained on artwork type images (vs photo), with an upscaling factor of 2, without noise reduction. The impact of this preprocessing step will be analyzed later in this thesis. The result is judged to be visually satisfying (**Fig. 2.5.1**). Noise is present in the super-resolved image, but seemingly only when it is also present in the original image.

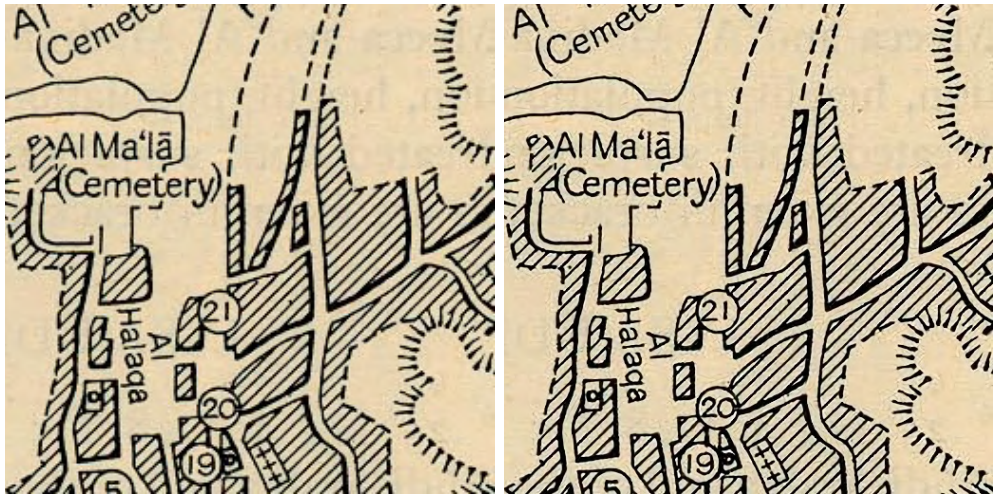

**Figure 2.5.1** Map of Mecca. From Western Arabia and the Red Sea. Great Britain. Naval Intelligence Division, 1946. [110] **(left)** sample from the original image. **(right)** 2x super-resolved sample.

Another type of preprocessing, based on the grabcut algorithm [111], has been tested. Grabcut is a graph-based algorithm, initially developed for foreground extraction, that may allow to correct manual annotations by comparing them to map patches, and by bringing the contours of the annotated elements together with the edges in the image. However, the impact on performance was found to be negative, as the mean IoU for the "corrected" pixels dropped significantly. Consequently, this approach was dismissed.

# 3 Figuration

## 3.1 Definitions

**Operationalization**: The process of making a concept mathematically intelligible by translating it into variables and operators.

**Texel**: Basic unit of texture, which can be repeated many times to cover a surface

## 3.2 Operationalization of figuration

In this chapter, we will first try to make cartographic figuration, i.e. representation, measurable. We will base ourselves on methods developed and used by Uhl et al [12], [36], [112], which offer visual-analytical approaches in map processing. In a first step [112], Uhl et al proposed using color-histogram based moments as visual descriptors. Such descriptors are very simple. The idea is to take each RGB channel of the image separately and draw a 256-bins histogram of it. In a second step, four distribution metrics: mean, standard deviation, skewness and kurtosis are calculated for each of the three RGB channels. The mean is used to transcribe the hue and the color value, the standard deviation indicates the contrasts. The skewness is a descriptor of the asymmetry of the color distribution, while the kurtosis is a flattening coefficient of the curve and therefore also allows to describe the contrasts. More recently [36], Uhl et al. used local binary patterns (LBP, [113]) as texture descriptors. The principle of LBPs is to list the neighboring pixels $p_i$ distant from a pixel $p$ by a radius $r$ (here $r = 2$) and to encode them as 1 if the value of $p_i$ is greater than the value of $p$, and 0 otherwise. In essence, LBPs are therefore invariant to value, i.e to the brightness of the image. They are made invariant to rotation by a bit-wise shift that maximizes the number of pixels encoded as 0 at the beginning of the radial sequence. The final LBP histogram is reduced from 17 to 12 bins. It is worth noticing that LBP descriptors are symmetric (see **Fig. 3.2.1**).

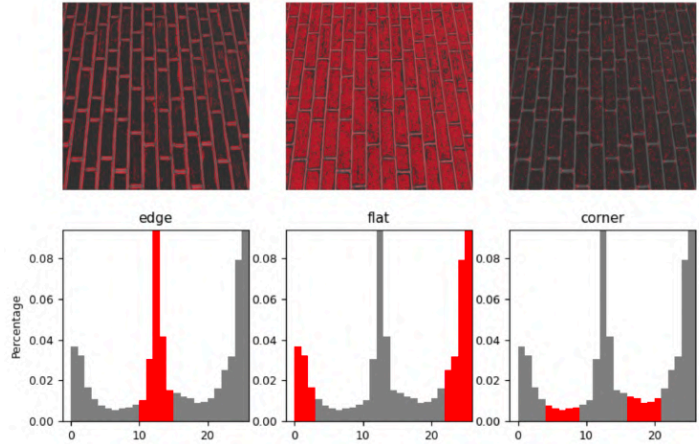

**Figure 3.2.1** Intensity of LBP descriptors (25 bins) on a sample image with a paved path texture. Image taken from Skimage documentation [114]

As the characterization of hatched textures in map processing is particularly important, a third type of features was added to color descriptors and LBP: the histogram of oriented gradient (HOG, [115]). Unlike LBP, HOGs are specific descriptors of orientation. The scale at which they work is also different, as HOGs are not pixel-wise descriptors, but rather describe textual orientation. In hatched areas, they can differentiate between horizontal, vertical or diagonal lines, for instance. HOGs are also powerful morphological markers. In computer vision, they are used with scale-invariant feature transform (SIFT) [116] to match visual keypoints. The principle of HOGs is based on the characterization of the orientations of the intensity gradient on a cell of the image. A simple calculation of the gradient allows to highlight value differences, in particular due to the local presence of edges. For each cell (groups of pixels), the local gradient is computed in every direction. These orientation gradients are grouped in a histogram, which is then binned to keep, in the present case, only a resolution of 24 orientations. For the specific case of the treated problem, these 24 orientations are then grouped anew with regard to the angle they describe in 5 superbin categories: vertical ($\pm\pi$), horizontal ($\pm\pi/2$, $\pm 3\pi/2$), diagonal ($\pm\pi/4$, $\pm 3\pi/4$), regular oblique ($\pm\pi/6$, $\pm 2\pi/6$, $\pm 4\pi/6$, $\pm 5\pi/6$), and irregular oblique (all other orientations).

To sum up, there are 29 features in total: 12 color descriptors, 12 LBP descriptors, and 5 HOG descriptors. For LBP, the grayscale images had to be binarized beforehand by Otsu's thresholding method [117], because of high noise. The **Fig. 3.2.2** depicts a colored plan of the city of Mocha (Yemen). To demonstrate the effectiveness of our figurative metrics, the texels of this map were rearranged according to their figuration. For this purpose, each color channel was first normalized, such as:

$$x' = \frac{x - \mu(x)}{\sigma(x)} = \frac{x - \bar{x}}{\sqrt{\frac{1}{n}\sum_{i=1}^{n}(x_i - \bar{x})^2}}$$

Where $x'$ is the normalized color channel matrix, $x$ is the original one, $\mu$ is the mean operator and $\sigma$ is the standard deviation. Subsequently, this map was cut into 50x50 pixel patches, from which the 29 visual features were extracted, and then normalized according to the L1 norm:

$$x' = \frac{x}{\|x\|_1} = \frac{x}{\sum_{i=1}^{n}|x_i|}$$

For visualizing the result, we proceeded as in [112], by first projecting the map patches features to a 2-dimensionnal space by t-distributed Stochastic Neighbor Embedding (t-SNE, [118]), and then forcing this projection to a grid with rasterfairy algorithm [119], in order to prevent overlapping of the visualized patches. For dimensional reduction, the state-of-the-art hierarchical density-based clustering (HDBScan, [120]) was also tested on various maps, but did not show better results than t-SNE. The example case of the Plan of the city of Mocha is shown in **Figs. 3.2.2** and **3.2.3**. We did not detect a kind of maps on which the methodology presented was ineffective, thus this example is considered representative of the performance. As one can witness, the rearrangement of map patches is very efficient, which demonstrates the high descriptive potential of our feature set to operationalize figuration.

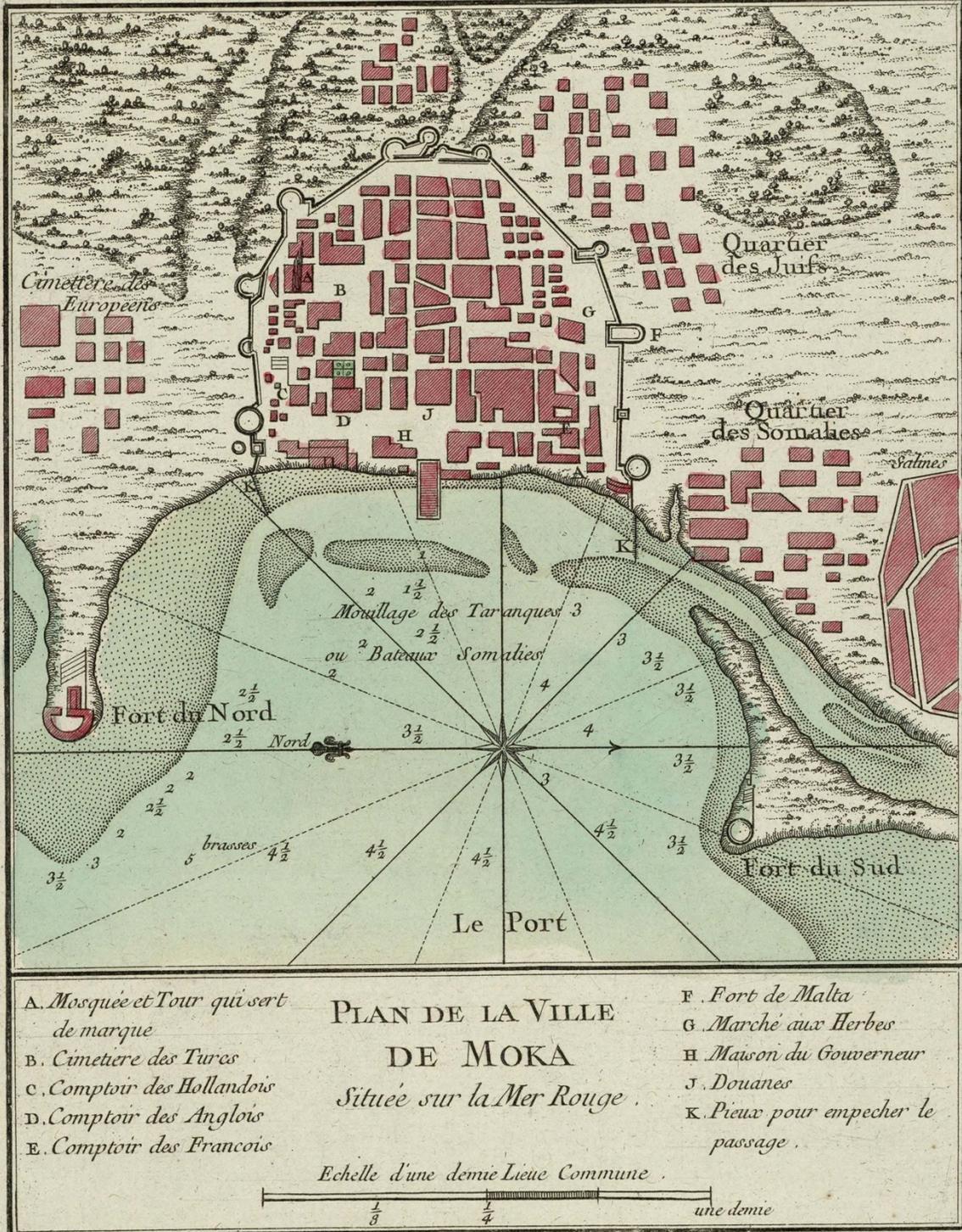

**Figure 3.2.2** Plan de la Ville de Moka (*Plan of the City of Moka*, Mocha, Yemen), in *Le Petit Atlas Maritime* [121, p. 20]

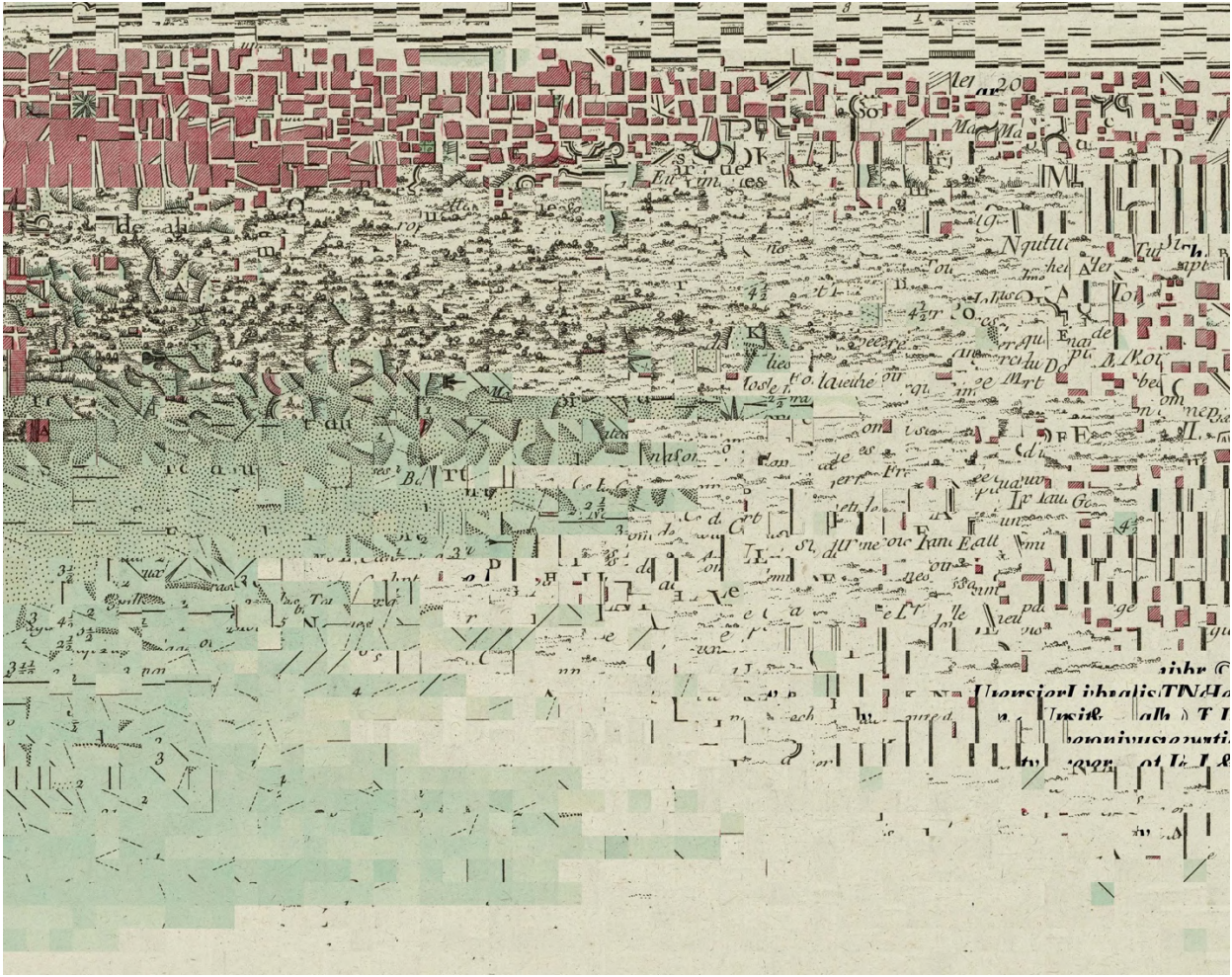

**Figure 3.2.3** t-SNE projection of the map patches' figuration descriptive features (from map in **Fig. 3.2.2**)

We can now build on this technology to investigate the first research subquestion: i) How to measure the figurative diversity in a map corpus? For each feature, we tried to determine how much the distribution varies, in other words, how flat or sharp the distribution is. This allowed to understand if the figuration is concentrated around peaks in the distribution, i.e. if there are typical and repeated figuration schemes, or not. To make the visual results more intelligible, the LBP features were grouped in the semantic categories shown in **Fig. 3.2.1** and the color-related features were slightly simplified by summing the 3 color channels for the distribution descriptors (standard deviation, skewness, kurtosis), so that those were computed on the image value. The number of simplified features was then 14 instead of 29, which facilitates visualization and interpretation. The **Fig. 3.2.4** shows an example of LBP outcome after grouping the channels in the 3 semantic categories mentioned above. One can also notice the limits of LBP here, which tends to assimilate very tight hatching as flat textures. The scale therefore influences the LBP features, which is one of the admitted limitations of the method.

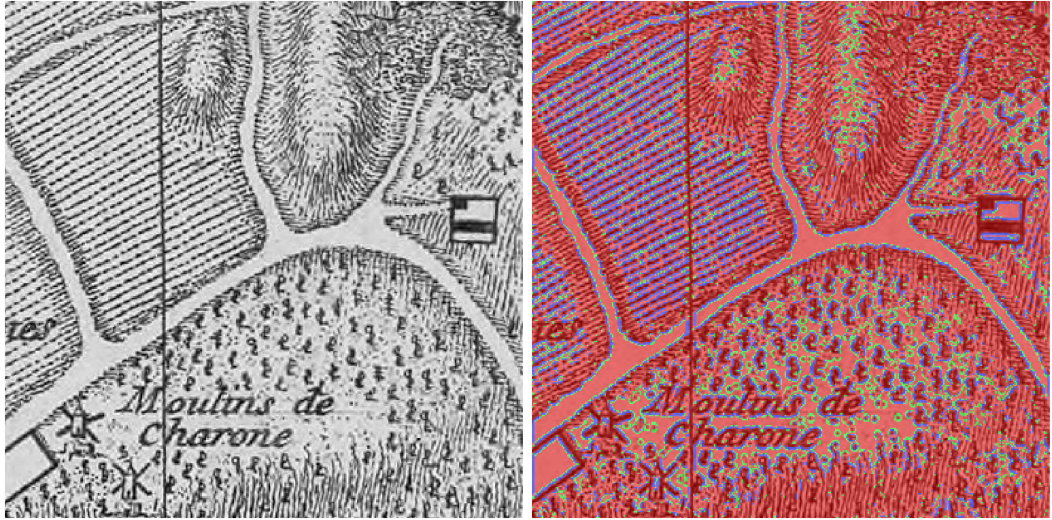

**Figure 3.2.4** Example of extraction of the 3 LBP semantic categories of features. **(left)** original grayscale image. **(right)** grayscale image on which the LBP semantic categories of features are superimposed. red = flat texture, blue = edges, green = corner

We are trying to identify repeated figurations, which appear as peaks when studying the features histograms. The aim is to calculate the local acuity of the distribution curve. To that end, for each feature, the flattening coefficient of the histogram curve, or kurtosis (from the Greek κύρτωσις, curvature), was computed:

$$Kurt(x) = \left(\frac{x - \mu(x)}{\sigma(x)}\right)^4$$

In the present case, the distributions can be multimodal. This is particularly the case when the figuration is very repetitive, as each repeated texture or color will produce a peak in the histogram of the corresponding feature. It is therefore necessary to separate the different modes in the distribution and to calculate the kurtosis separately on each of them. To separate the distribution modes, a low-resolution histogram with 32 bins was created. The histogram was smoothed with a Savitzky-Golay filter [122] of width 3 and polynomial degree 1. The local minima $m_i$ were identified on a window of width 3 and the histogram was split between each mode. Then, a κ-coefficient was computed as the weighted sum of the kurtoses on each mode, where the weight $w_i$ corresponds to the proportion of the population contained in the mode $i$:

$$\kappa(x) = w_0[Kurt(x_j), x_j \leq m_1] + \left[\sum_{i=1}^{n} w_i Kurt(x_j), m_{i-1} < x_j \leq m_i\right] + w_{n+1}[Kurt(x_j), m_n < x_j] \quad (1)$$

The κ-coefficient was computed, using texels of 50x50, on all samples from the world and the Paris corpora, as well as for the USGS map used by Chiang et al. [123] and shown in **Fig. 3.2.6**. As the number of samples varies for these three corpora and as the value of the κ-coefficient can vary according to the size, the bigger sample sets were randomly downsampled, without replacement, to the size of the smallest sample set (here the USGS set). The κ-coefficient was then computed 5000 times for various downsampling schemes and for each feature, the median κ-coefficient was retained as an estimator of the real κ-coefficient. The bias of this recalibration is below ±3.54% for the world, and below ±1.95% for Paris, with a confidence of 95%. The results are plotted in $\log_{10}$ scale, in **Fig. 3.2.5**.

The median and mean overall κ was also computed on each dataset. For this non-visual result, all the 29 features were included. The mean κ is 6.06 for the world, 7.75 for Paris and 102.87 for the USGS map. This means that the diversity of both corpora studied in this work is massive compared to the USGS map, which is often the focus of map processing research. It is possible to analyze these differences more precisely by looking at **Fig. 3.2.5**. The values away from the center of the circle indicate acute histogram modes, thus a great figurative convergence, or homogeneity. On the contrary, the values close to the center of the circle indicate a wide variety of existing figurations and a rich figurative diversity. In essence, one could say that the kurtograph indicates the relevance of considering a feature, as it indicates their power of characterization of the figuration, or the figurative convergence. Then, if the feature is considered characteristic, the associated histogram can be used to determine the way in which this feature is distinctive.

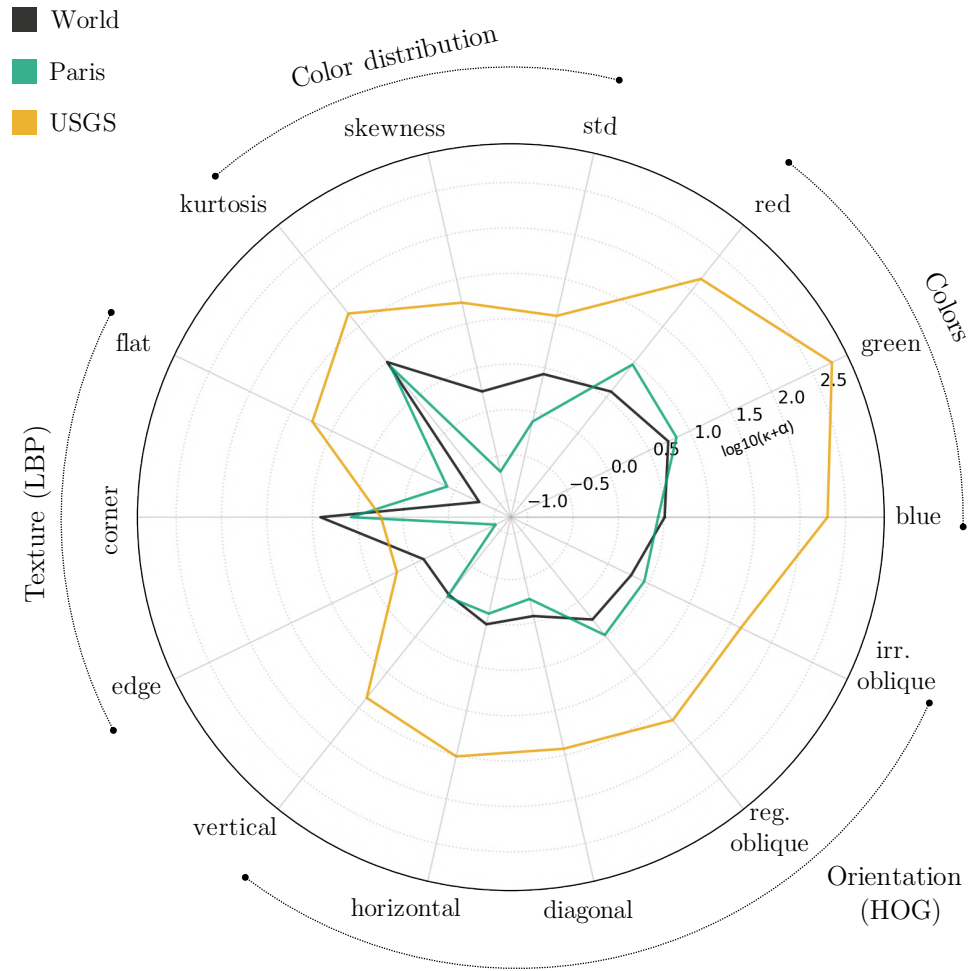

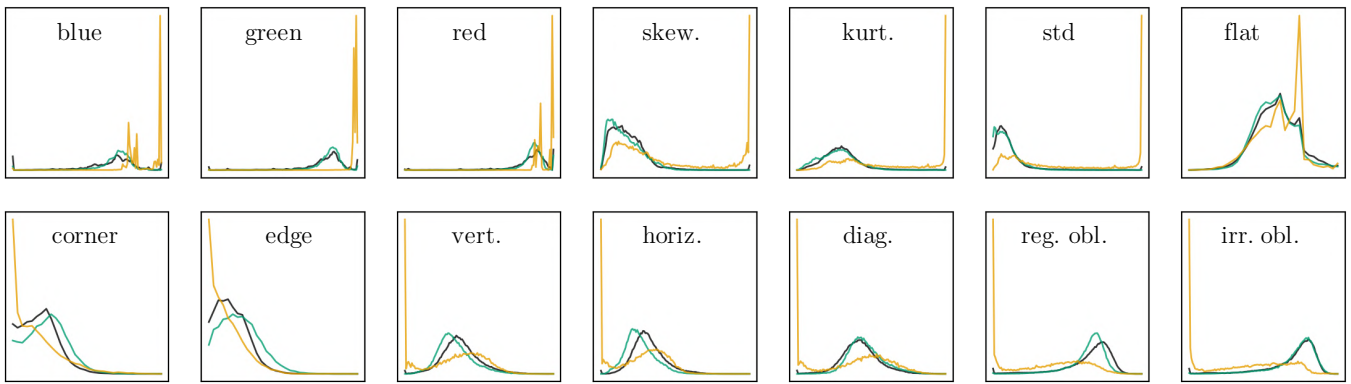

**Fig. 3.2.5** Radial kurtograph of figuration, comparing the $\kappa$-coefficient (see Eq. 1) of the visual features' distributions between the corpus world, Paris and the USGS map used by Chiang et al. [123], and detailed figurative features histograms. $\alpha \ni \kappa + \alpha \geq 0.1 \; \forall \; \kappa$

**Fig 3.2.6** USGS Map of Bray quadrangle [124] used by Chiang et al [123]

One can note that the great figurative convergence of USGS maps is mainly expressed in color: the $\kappa$ for the average RGB channel is up to 100 times higher for the USGS map, compared to both other corpora. The USGS color distribution is characterized by a constancy in sharp contrasts, expressed in high standard deviation and kurtosis, and high skewness, which can be explained either by differences in value or in saturation. The textures are mainly characterized by a lesser number of corners and edges textures (cf. **Fig. 3.2.4**), compared to the other corpora, with many patches containing none of those textures at all. However, those features are also present, and the kurtogram reflects this variety. There are also clear and sharp peaks of flat textures.

Additionally, the HOG features show a high $\kappa$-coefficient. This is due to a bimodal distribution of the USGS HOG features, with a half peak centered in 0. Counter-intuitively, this may indicate that the orientation of the lines is more clearly marked on the USGS map, and that there is therefore little confusion with other HOG orientations. In contrast, the HOG features of the world and Paris corpora exhibit a unimodal distribution. The Parisian corpus differs mainly from the world corpus by a greater diversity of edges and corner textures extracted by LBP. Those textures might indeed be more present in the Parisian corpus. The reds also seem more constant, in the contrary of the skewness of the color distribution.

For the USGS map, the results are easily explained when looking at **Fig. 3.2.6**. The colors are highly coded and respect a clear palette. This is probably the reason why so many studies involving USGS maps focus on color segmentation. The very acute peaks that can be observed in the LBP descriptors seem to be caused by the grey and homogeneous frame. The high acuteness and the sharpness of contrasts is partly due to the fact that this map did not have to go through a scanning step. Indeed, it was probably directly transformed into a raster image from its digital version. The sharpness of the image also reinforces the descriptive character of the HOGs. For the rest, the lower homogeneity of the edges and corner textures is probably due in part to the brown contour lines, which create very different texture patterns.

Concerning the Paris corpus, the homogeneity of the red tones is probably not due to the existence of a "Parisian red" used in cartography. In fact, when looking directly at the samples concerned, one can see that this is more likely due to the homogeneity of the color of the paper produced by Parisian paper mills (see **Fig. 3.2.7**).

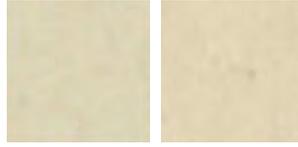

**Figure 3.2.7 (left)** Average color of paper of the Paris corpus. RGB (221, 212, 182) **(right)** Average color of paper of the World corpus. RGB (227, 213, 180)

In short, the main elements that emerge from this analysis are above all a much higher figurative diversity for the two corpora studied in this work compared to the maps traditionally used in map processing. While the analysis on the USGS map easily allows to isolate peaks of characteristic figuration, the cartographic grammars used within each of our two corpora are protean. This extreme diversity in figuration may make the depicted elements more difficult to classify.

To conclude, we made progress in understanding the figurative variance of corpora, and established a congruent metric, which was the primary objective of this experiment. Previous graphs gathered all the elements of the map together in a mashup. Thus, they were an effective means of verifying the existence of a diversity of figurations. However, to better understand the figuration conventions themselves, it is necessary to focus on semantic elements. The next section will investigate this avenue that might help to address the second research question, which stems from the first: ii) How does the representation of specific map elements, such as buildings, streets, water, or crops vary?

## 3.3 Analysis of the cartographic figuration

To study the representation of specific cartographic elements, the 50x50 patches are linked to the annotated labels of the 1000x1000 patches from which they are derived. When the annotation of a small patch indicates that at least 75% of its surface area belongs to a same single class, the patch is kept and considered as belonging to this class. Otherwise, if the label is more undecided, the patch is discarded for this experiment.

For each of the two corpora, 200 random examples from each class are shown in **Fig. 3.3.1**. Before attempting to analyze the second research question from a quantitative perspective, it is possible to provide some qualitative answers from this intermediate result. Apart from the paper color (**Fig. 3.2.7**), few differences are observed for the map patches representing the frame class. In both cases, they are characterized by a flat texture, similar to the color of the paper. Text, e.g. lists of street names, is also regularly present. The frame may also contain the scanner background or the fabric canvas hem. Decorative elements, such as floral patterns, are also regularly found in this class.

The representation of water differs more between the two corpora. Parisian maps seem to make more frequent use of hatching and less frequent use of color. Blue is unsurprisingly used on both sides. However, the palette seems slightly more extensive for world maps, ranging from azure blue to olive green, through grey, turquoise and Sienna yellowish brown. It is more restricted for Parisian maps, where it is concentrated around shades of sky blue. The Parisian hatching also seems to be more regular, compared to the world corpus.

The representation of Parisian buildings and blocks is sober. It is using diagonal hatching or flat texture, and few colors. For world maps, the colors are more vibrant, with a variety of shades. Flat and colored textures seem to be frequent. Hatching is also used here, with greater variations of oblique, or even with horizontal orientation. Differences in the regularity of the urban fabric can also be seen.

The representation of the non-built seems to use various but recurring codes for the Parisian maps. For example, dotted textures, small aligned circles, traitillated, or even traitillated hatching. Flat textures are also used, sometimes with a green or greenish coloration. For world maps, the representation seems more disparate. A grid texture is sometimes used, as well as wavy hatching. Flat textures are also widely used.

The representation of the road network seems relatively similar between the two corpora. The presence of text and parallel lines is evident, with perhaps a little more text on the Parisian side. Angles and rows of trees may also appear on either side. Red transverse lines are also regularly present, especially on the Parisian side, perhaps to represent boundaries or tramway lines.

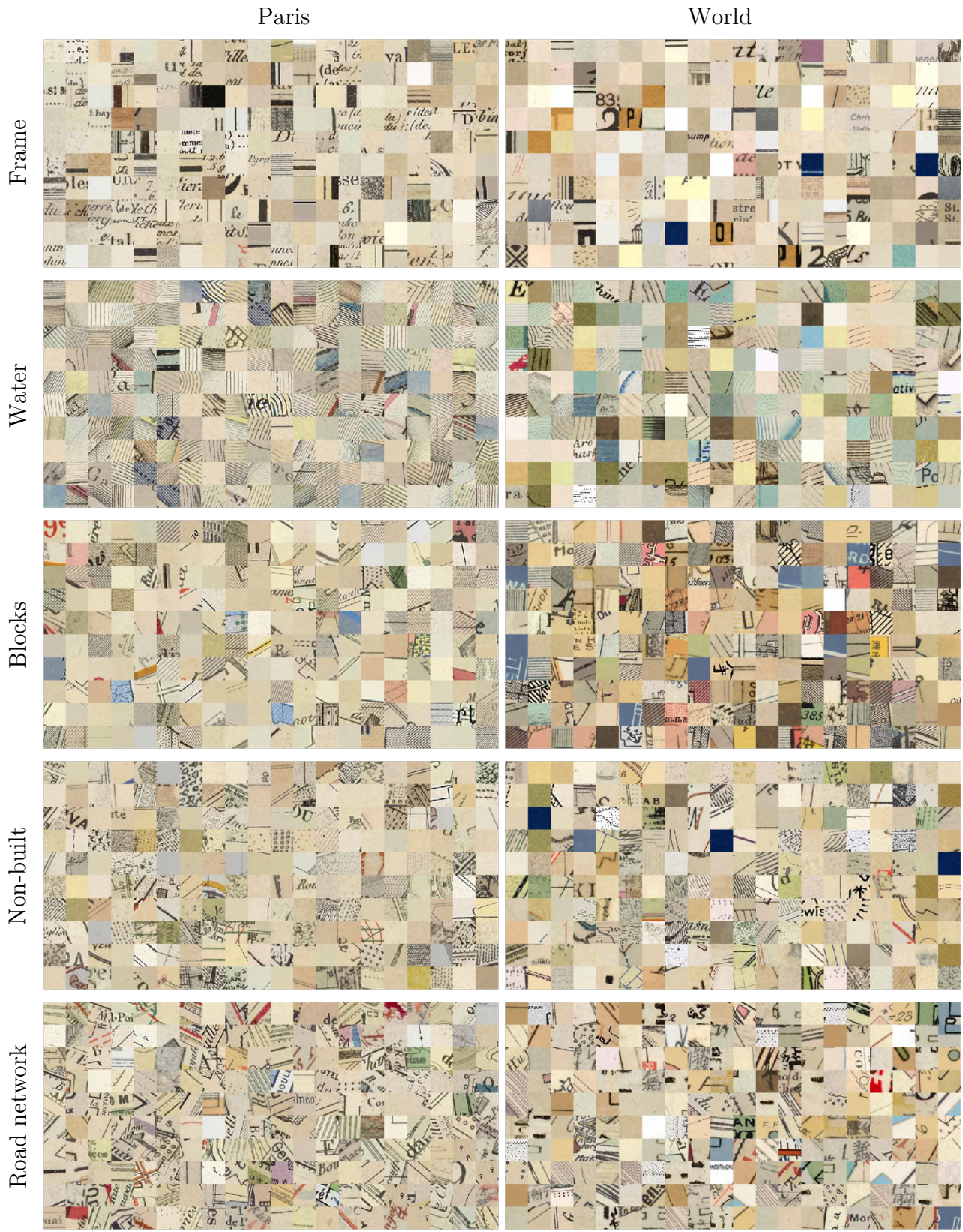

**Fig. 3.3.1** Image patches extracted from Paris and world corpora

In a second step, one can analyze these whole sets quantitatively, using the method developed in the previous section. Again, the size of each set is regulated by a repeated and random downsampling, over 5000 trials. The kurtographs are represented for the world corpus in **Fig. 3.3.3** and for the Paris corpus in **Fig. 3.3.4**. The mean and median values of the $\kappa$ class coefficient are detailed in **Tab. 3.3.2**. The overall recalibration bias is below $\pm 2.50\%$ for the world, and below $\pm 2.30\%$ for Paris, with a confidence of 95%. The bias per class is below $\pm 2.26\%$ for the world, and below $\pm 3.03\%$ for Paris, with the same confidence.

**Table 3.3.2** Mean and median values of the $\kappa$-coefficient per class and overall for all minipatches, including unassigned ones

| Corpus | Metric | **Overall** | Frame | Water | Blocks | Non-built | Road network |
|---|---|---|---|---|---|---|---|
| World | median | **4.19** | 4.73 | 14.41 | 6.17 | 3.13 | 5.41 |
| World | mean | **9.10** | 9.09 | 15.22 | 9.81 | 7.72 | 9.86 |
| Paris | median | **4.43** | 6.19 | 9.74 | 8.65 | 6.35 | 4.98 |
| Paris | mean | **8.06** | 9.50 | 12.85 | 11.79 | 6.85 | 6.84 |

At first glance, one can see that the mean values per class are generally higher than the overall, i.e. all the minipatches in the set. This is true for the water and blocks classes, as well as for the road network in world's maps, and the frame in Parisian maps. The median is always higher than the overall, except for the world non-built. This demonstrates the existence of a figurative idiosyncrasy of those classes and the specificity of intra-class figuration, despite the great diversity of corpora. On the other hand, the class mean $\kappa$ is always higher than the median $\kappa$. This means that this intra-class figurative convergence is expressed mainly on some of the figurative features, and not on all of them. In this sense, the classes with a high median $\kappa$ can also be considered to be convergent. However, their description should include a wider collection of features, and not just the most salient ones. The non-built class of the world corpus, however, shows a clear lack of convergence.

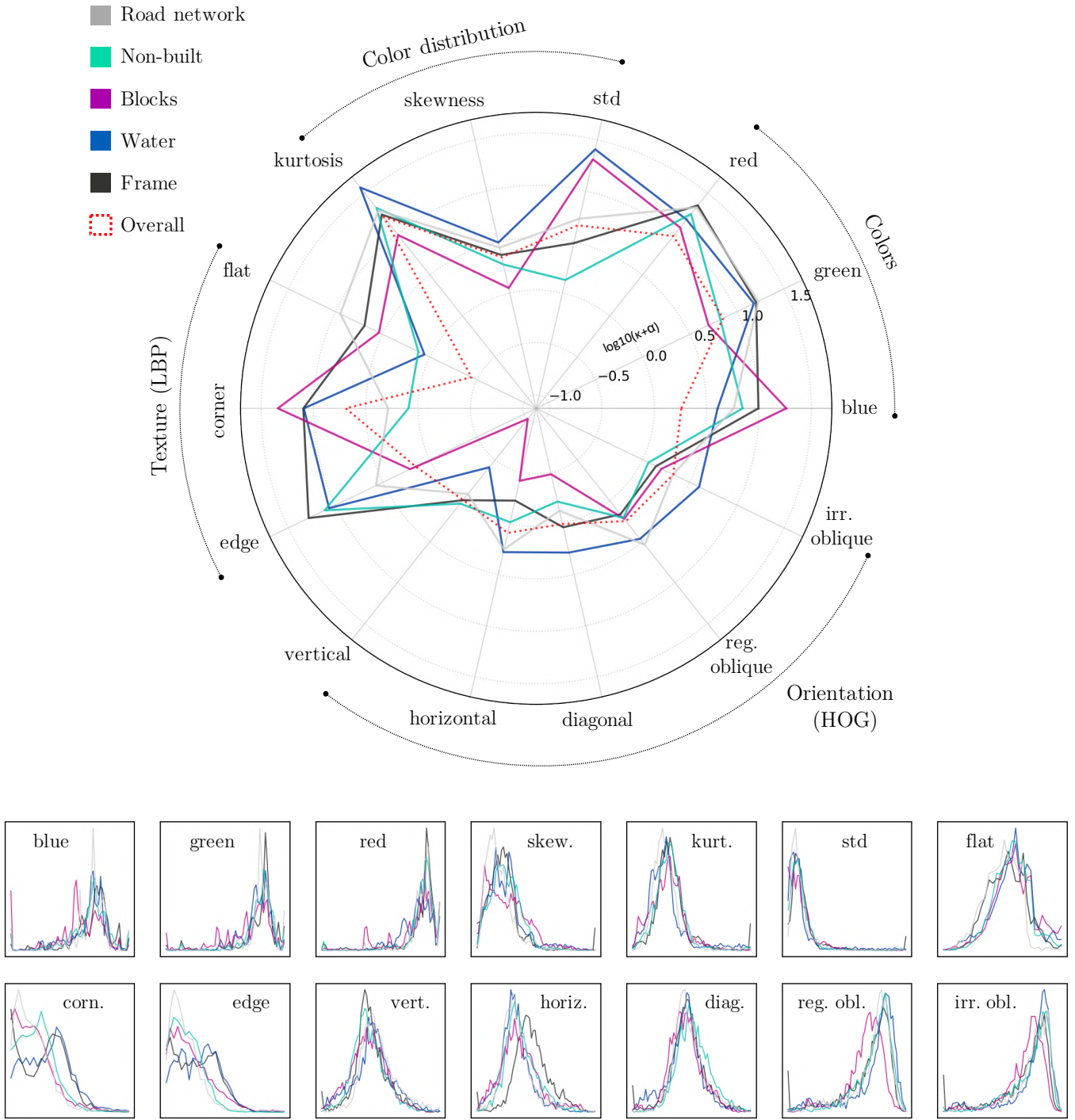

**Fig. 3.3.3** Radial kurtograph of figuration, comparing the $\kappa$-coefficient (see Eq. 1) of the visual features' distributions between the classes of the world corpus, and detailed figurative features histograms. $\alpha \ni \kappa + \alpha \geq 0.1 \; \forall \kappa$

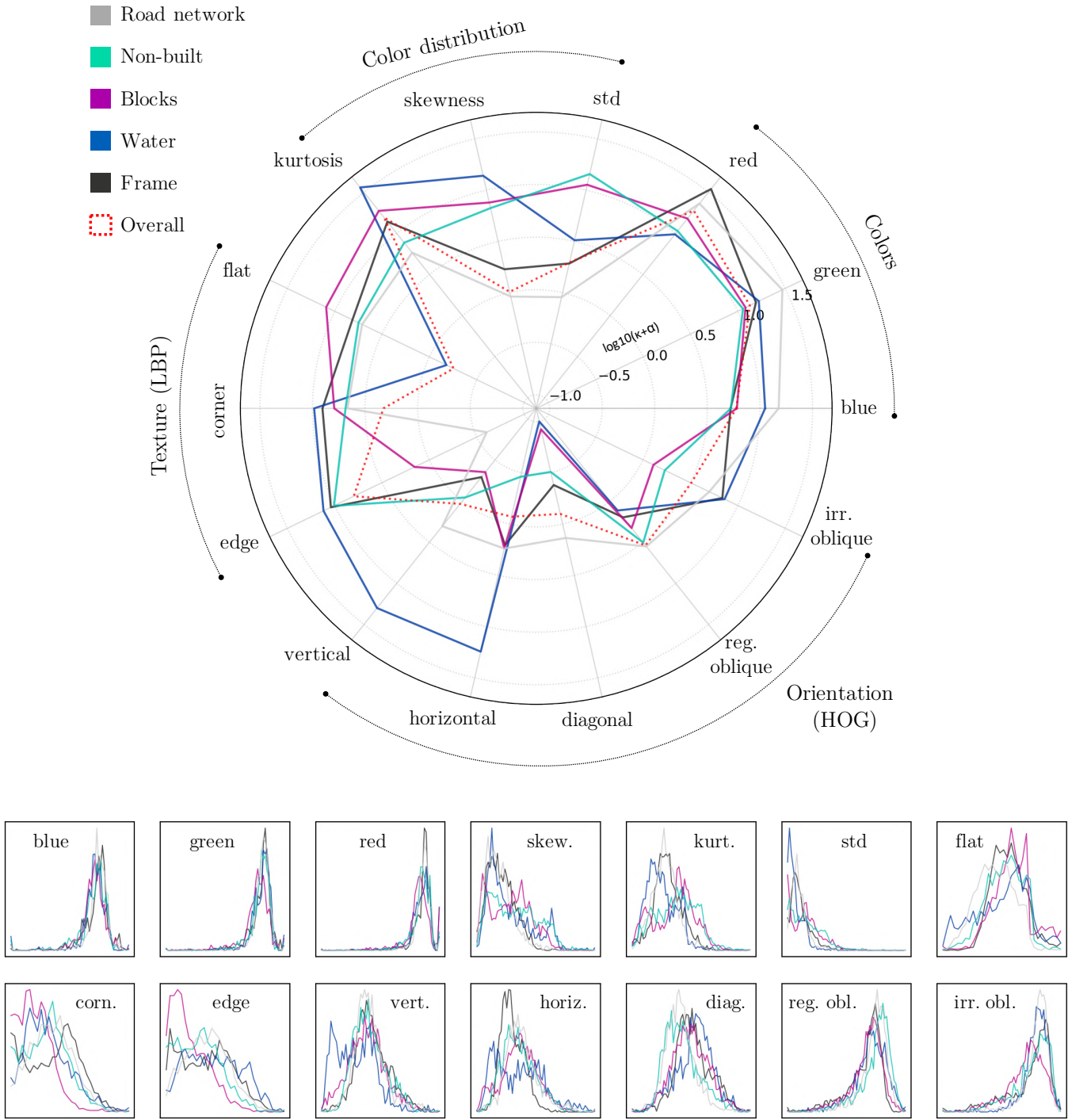

**Fig. 3.3.4** Radial kurtograph of figuration, comparing the $\kappa$-coefficient (see Eq. 1) of the visual features' distributions between the classes of the Paris corpus, and detailed figurative features histograms. $\alpha \ni \kappa + \alpha \geq 0.1\ \forall\ \kappa$

Comparing $\kappa$ between the world corpus and Paris corpus, one can notice that the world corpus has some more characteristic features, but that the specificity of the Paris corpus is expressed in a larger number of features. We recall that the $\kappa$-coefficient is used to determine whether a feature is characteristic. The further the feature is from the center of the kurtograph, the more idiosyncratic and figuratively convergent it is. Histograms help to interpret how this characteristic feature is expressed visually. Compared to the metric of the previous section, the values are increased because the downsampling is more severe. In general, one could conclude that the Parisian corpus expresses a greater figurative richness on small groups of maps, but that this figurative diversity then reaches a plateau. Large-scale convergence analysis is therefore also important.

By observing the classes separately, one can see on the kurtographs that the frame class is characterized mainly by the RGB values, thus its color. Indeed, in the histograms, the main color peaks are particularly marked, which is indicative of the paper color. In relation to this, the prominence of the red will be noted once again for Paris. Another characteristic feature is the presence of angles and edges, probably due to texts and frame lines. Flat textures are also frequent which corresponds, as mentioned above, to the paper.

The water class is clearly the one with the greatest figurative convergence for both corpora. The color of the water is significant in both cases. However, it seems that most of the time it adopts a neutral color. A peak of cold color (under-expression of reds) is however visible for the world corpus. This colored minority could also correspond to the skewness peak. For both corpora in general, the color tends to be homogeneous, with a low standard deviation and kurtosis. Two types of textures can be distinguished. On the one hand, the rather flat textures, which probably correspond to the neutral, sometimes colored peaks we mentioned, and on the other hand the textures involving edges, and sometimes corners in the world corpus. This probably corresponds to the wavy hatching we observed visually in **Fig. 3.3.2**. For Paris, the edges are prioritized, with a rather horizontal orientation, and a low verticality, which is logical, given the orientation of the Seine. For the world, the lines seem to be less oriented and the frequent curvature of the textures is visible in LBP corner indicator.

The representation of the blocks on the world maps uses warm tones, as the blue curve is peaking on the left. In accordance with what was observed above, it seems that the flat colored textures are especially characteristic. The textures use very few corners and relatively few lines. However, hatching textures of buildings may not be detected by the LBPs, as their radius does not allow the detection of very close hatching, as shown in **Fig. 3.2.4**. The figuration of this class is particularly convergent for the Parisian corpus.

The non-built is the only class that is clearly non-convergent for the world. This confirms our first impression. Because of this insignificance, we are not going to detail this class for the world. For Paris, figuration implies a large number of features without specifically relying on salient ones. The color value seems to be high, which with high skewness can mean higher brightness, as well as marked contrasts (high kurtosis and std). Linear and corner textures prevail, as well as regular oblique orientations.

Finally, the last class, the road network, is only moderately convergent for both corpora. Interpretation is difficult, as its features differ little from the features of other classes. The color is close to the paper, perhaps slightly darker, with very few variations and very few colored areas a priori, which is also visible in a low skewness. One can also observe a low prevalence of flat areas. Observing LBPs, one can say that the class is not very textured for the world. It can contain clearly oriented edges for Paris.

## 3.4 Inter-class proximity

We also sought to measure the figurative distance between classes. This allowed to estimate which classes share the same codes. For this part, all the 29 features were used. The different classes were extracted under a 75% area coverage threshold, and downsampled as explained above. The features histograms were correlated with the Pearson method within the classes of the same corpus and between the classes of the world corpus and their counterpart of the Paris corpus.

The aggregated correlation matrix is found in **Fig. 3.4.1** and the associated mean correlations per class in **Tab. 3.4.2**. First, one can notice that the mean inter-class correlation between the classes of the corpus World is significantly higher compared to the inter-class correlation between the classes of the corpus Paris ($\mu_{r\_Paris} = 0.842 < \mu_{r\_World} = 0.917$, $p_{value} = 0.0026 < 0.05$). In fact, the mean inter-class correlation in the World corpus is even higher than the class correlation accross corpora ($\mu = 0.879$). This means that the classes from the World corpus are in average closer from one another than they are from the same classes in the Parisian corpus. This is also visible on the kurtographs (**Fig. 3.3.3** and **Fig. 3.3.4**), where one can easily see that the histogram curves are very intermingled for the corpus world, and less so for the corpus Paris.

The non-built is the class showing the lowest figurative convergence between the two corpora, with a Pearson correlation of only 0.81, which remains however high. This means that the figurative codes used to represent the non-built vary considerably between French maps and maps of the rest of the world. In fact, we had already noticed that the figuration of this element was already non-convergent among world maps (**Tab. 3.3.2**). Oppositely, the frame class is the class showing the highest correlation (0.96) between the two corpora. Logically enough, the representation of the frame of the map varies little. This is also an interesting reference point, which allows to note the non-negligible part of codes shared for the representation of water, roads and urban blocks between the French and the other world's maps.

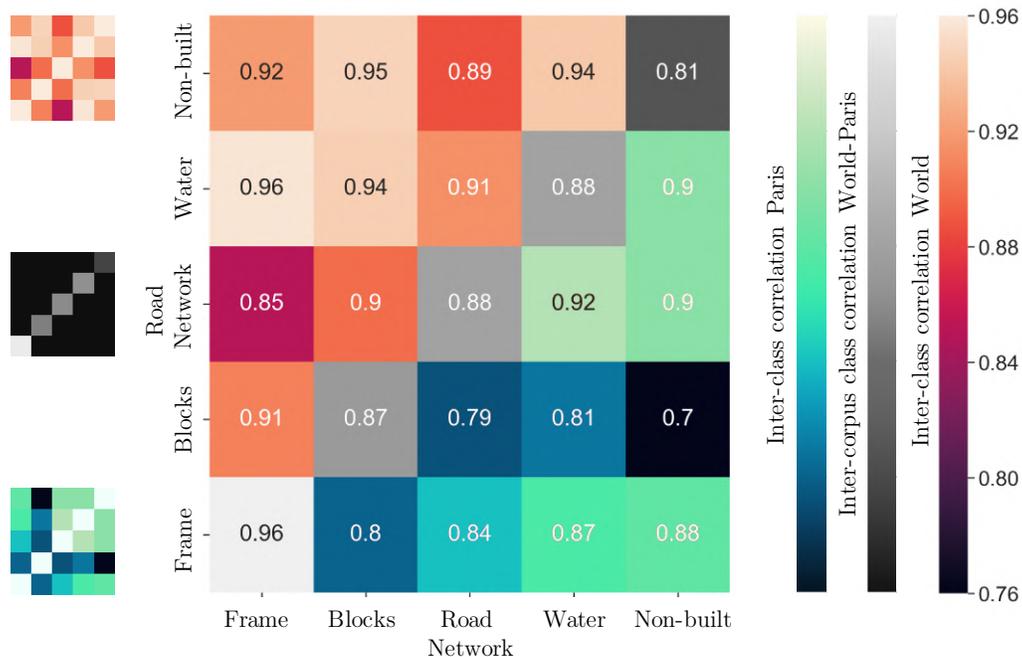

**Figure 3.4.1** Aggregated correlation heatmap matrix of intra-corpus inter-class correlation and inter-corpus class correlation. $r_{pearson} \in [-1, 1]$. All correlations are significant ($p_{value} \ll 0.05$)

**Table 3.4.2** Mean intra-corpus inter-class correlations

| Corpus | **Mean** | Frame | Blocks | Road netw. | Water | Non-built |
|---|---|---|---|---|---|---|
| World | **0.917** | 0.908 | 0.924 | 0.888 | 0.939 | 0.924 |
| Paris | **0.842** | 0.846 | 0.775 | 0.862 | 0.877 | 0.847 |

The class that seems to distinguish itself the best from the others within the Parisian corpus is the block, which only obtains a mean correlation of 0.775. The dissociation between the block class and the non-built class is particularly strong ($r = 0.7$). Conversely, the block and non-built classes of the world corpus are poorly dissociated ($r = 0.95$).

The high correlation between the different classes in the corpus world could be problematic for the segmentation. Indeed, this means that the figurative features used here, such as color and texture have a low discrimination power for the world and could probably not constitute a reliable cue for the neural networks. The classes that are best differentiated from the others, from the point of view of these elementary features, such as for example the urban blocks for the Paris corpus, could in the contrary be better recognized.

## 3.5 Chapter conclusion

In this chapter, a method for operationalizing the cartographic representation was developed. When applied to the world corpus and the Paris corpus, as well as to a USGS map, one can establish that the figurative diversity of our corpora is far superior to the diversity of the data normally used in map processing ($\kappa_{World} = 6.06$, $\kappa_{Paris} = 7.75$, $\kappa_{USGS} = 102.87$). When this operationalization method is used directly on the different semantic classes of the Paris and world corpora, one can observe that the intra-class variance is generally greater than the inter-class variance, demonstrating the existence of a figurative idiosyncrasy of the classes. It was also demonstrated in this chapter that the way in which specific map elements, such as water, blocks, and roads are represented could be investigated and analyzed quantitatively at a large scale. For instance, this allowed to note that the figuration of water is very characteristic ($\overline{\kappa_{World}} = 15.22$, $\overline{\kappa_{World}} = 12.85$), as well as that of urban blocks ($\overline{\kappa_{World}} = 9.81$, $\overline{\kappa_{World}} = 11.79$), to some extent. On the other hand, the figuration of the non-built is not so characteristic ($\overline{\kappa_{World}} = 7.72$, $\overline{\kappa_{World}} = 6.85$). The map frame shows few cultural differences between Paris and the world ($r_{pearson} = 0.96$), whereas the non-built, precisely, seems to be more culturally dependent ($r_{pearson} = 0.81$). On average, the classes of the corpus world are very inter-correlated ($r_{World} = 0.917 \gg r_{Paris} = 0.842$) and their figuration differs little from each other. These findings provided some answers to the first two research sub-questions. In the next chapter, we will focus on the semantic segmentation of cartographic documents using neural networks.

# 4 Model

In sections 1 to 6 of this chapter, we will focus on getting the best possible performance for the semantic segmentation problem. In section 7, we will present the results of these performances, which allow to establish a new performance benchmark in map processing. We will discuss the results of the experiments immediately. However, the final discussion about the last two research sub-questions will take place in the concluding chapter.

## 4.1 Network architecture

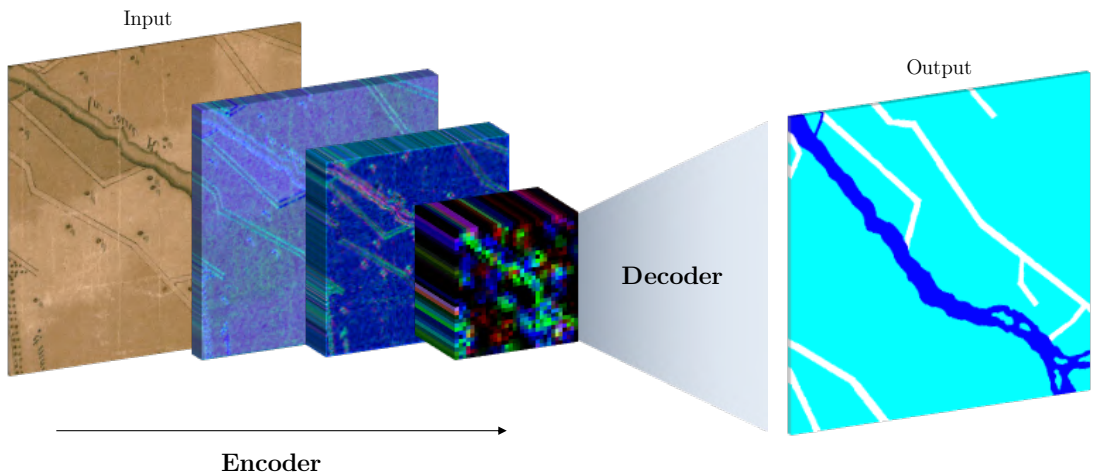

**Figure 4.1.1** Principle of CNN-based semantic segmentation

Neural networks designed for image segmentation generally have an architecture consisting of an encoder and a decoder (see **Fig. 4.1.1**). The general idea behind this architecture is very simple. The input image in full resolution first goes through an encoder. The encoder reduces the spatial dimensions of the image, while increasing the number of layers, and thus the number of features that describe it, using operations that can be convolutions or poolings. In a way, the encoder allows the image to be abstracted to a higher level of understanding than what is possible with mere pixel values, in an effort of description. In a second step, a decoder performs the reverse path. It increases the size of the spatial features again, reduce the depth, and reconstructs an image: the output. The semantic output describes the type of objects recognized in the original image, in an interpretative effort.

For map segmentation problems, Oliveira et al. [11] use a UNet [125] architecture (**Fig. 4.1.2**), with ResNet [126] as encoder. Chiang et al. [44] test three different encoders in an FCN architecture: VGG16, GoogLeNet and ResNet. On a railway segmentation problem, on a homogeneous corpus, VGG16 gets an IoU of 0.048, GoogLeNet 0.118 and ResNet 0.231. They also test a PSPNet [127] architecture (**Fig. 4.1.3**), which obtains an IoU of 0.291. In a third experiment, they use a modified PSPNet architecture with skip-connection in VGG encoder, which they do not detail, and obtain an IoU of 0.622. Skip-connections introduces "contacts" between the first, simpler layers and the more abstract layers to preserve the initial spatial information [2], [44, pp. 79–80]. They are found in the ResNet architecture, and in another form, in UNet.

We're going to run initial tests on three state-of-the-art architectures : UNet, implemented in a forthcoming Pytorch [128] version of dhSegment [43], as well as PSPNet and UperNet, both implemented by Zhou et al [129]–[131].

We will use the UNet architecture (**Fig 4.1.2**) with a ResNet encoder. The intuitive idea of UNet is to store the features extracted at each encoder level, then concatenate these features during decoding. Thanks to ResNet skip connections, the network has a kind of memory of the lower levels of abstraction and this allows to better define the shapes reconstituted by the decoder.

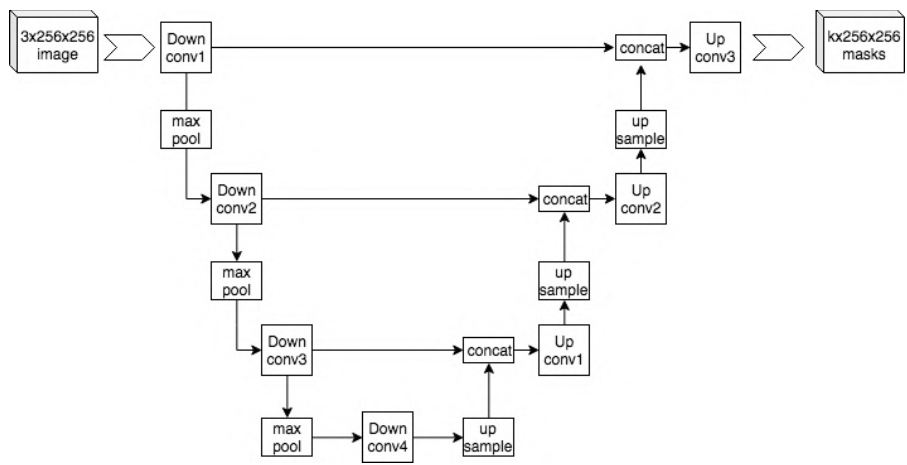

**Figure 4.1.2** Example of a UNet architecture. Source of the image: [132]

PSPNet and UperNet are both based on the principle of spatial pyramid pooling [133]. The pyramid pooling module (PPM) aims to make the network size-agnostic. In fact, these architectures include several levels of image resolution, which aims to give better performance on small objects and fine objects, such as lines [127].

For reasons of resources and computing time and because the last two architectures in particular are heavy to train, an extended grid search could not be performed. The results are therefore indicative and are mainly aimed at selecting a basic architecture to be optimized.

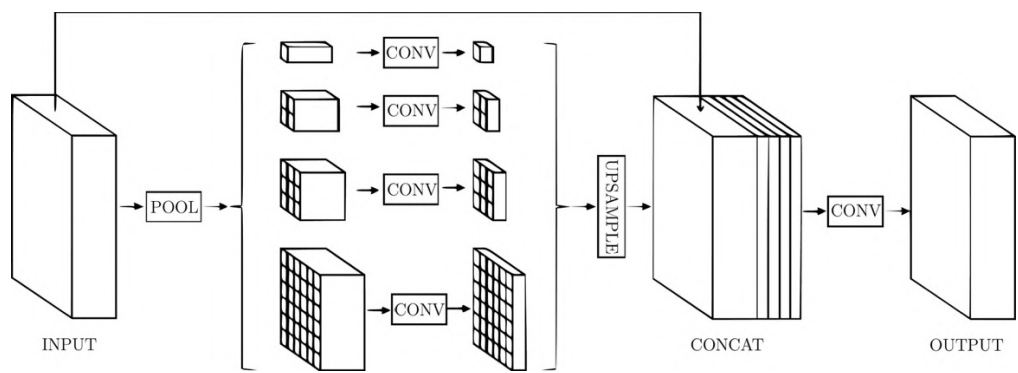

**Figure 4.1.3** Example of a PSPNet architecture. Redrawn and adapted from [127]

The methodology is as follows. Each network was trained during 50 epochs, with a batch size of 2, on 300 training examples from the Paris dataset, and annotated with 3 classes. The data were augmented by side flip and upside-down flip, and with a rotation $r \in [0, \pi]$. The learning rate was $5 \cdot 10^{-5}$, the loss used was cross entropy. A slight weight decay regularization of factor $\lambda = 1 \cdot 10^{-7}$ was applied on the gradient to prevent overfitting. The optimization relied on stochastic gradient descent (SGD). ResNet50, pretrained on ImageNet [134], was used as encoder. PSPNet was using PPM with deep supervision [135].

The training lasted 26h for PSPNet, 9h for UperNet and only 2h for UNet. At the end of each epoch, the network was validated on a set of 30 samples from the Parisian corpus. The performance metric used to validate the results is the mean intersection over union (mIoU), defined as the mean over each class of the overlapping area divided by the intersecting area, between prediction and ground-truth, or $\frac{true\ positive}{true\ positive + false\ positive + false\ negative}$.

For each model, the mIoU of the best epoch is retained. The best performances were obtained by UNet (mIoU = 0.718), followed by UperNet (mIoU = 0.5495) and finally PSPNet (mIoU = 0.4845).

At the end of this experiment, UNet was selected as it reached the best performance in the shortest training time. UNet and the above settings will be used for all subsequent experiments unless otherwise specified.

## 4.2  Batch size and learning rate

For learning purposes, the training examples can be propagated in the network by batches, i.e. "packets". Powerful graphics cards are able to process several images in parallel. The gradient on the loss function is summed for each learning batch, which generally facilitates convergence. When the graphics card is less powerful or to save memory, it is possible to use minibatches. The principle of minibatches is to accumulate the gradient of several smaller batches, before applying a gradient descent to optimize the cost function. A *batch size · minibatch size* > 1 generally favors convergence because the accumulated gradient of several batches allows a "smoother" descent towards an optimum, avoiding crenellations.

The batch size also influences the size of the learning rate (lr), i.e. the size of the steps along the descending gradient. The optimal ratio between the two is subject to debate. According to Goyal et al [136], the dependence becomes linear with a very large batch size. However, at our scale, the following rule from the work of Krizhevsky [137] will be used: a multiplication of the batch-minibatch size by 2 leads to a multiplication of the learning rate by $\sqrt{2}$. We therefore consider a quadratic dependency.

Without grid search, i.e. without iterating dumbly on the two parameters to obtain the best combination at the cost of considerable computing time, the optimization of these two parameters is somewhat tedious. We started by training UNet as described in section 4.1 for 60 epochs with a batch size of 1 and by varying the learning rate from $1 \cdot 10^{-5}$ to $3.2 \cdot 10^{-4} = 32 \cdot 10^{-5}$ by successive doubling steps. The slight increase in the number of epochs aims not to penalize the smaller learning rates, which may be optimal, but might also converge more slowly. At the end of the experiment, it seems that the optimum is found between $4 \cdot 10^{-5}$ and $8 \cdot 10^{-5}$. The hypothesis $H_0$ is that the mIoU obtained for a lr between $4 \cdot 10^{-5}$ and $8 \cdot 10^{-5}$ is equal to the values obtained for a lr between $1 \cdot 10^{-5}$ and $2 \cdot 10^{-5}$, and between $1.6 \cdot 10^{-4}$ and $3.2 \cdot 10^{-4}$. Hypothesis $H_1$ is that the mIoU is higher for a lr between $4 \cdot 10^{-5}$ and $8 \cdot 10^{-5}$. A Student's t-test validates $H_1$ unambiguously ($p_{value} = 0.022 < 0.05$). Since the variance is high, it is difficult to obtain a more accurate though statistically significant estimate of the learning rate. Consequently, we will select a value of $5 \cdot 10^{-5}$, which is closer to the lower bound of the range, for safety, and increase the number of training epochs in the future to ensure a good convergence.

In a second step, several values of batch · minibatch size were tested, namely a batch size of 1 and a minibatch size of 1, 4, 16 or 64. The corresponding learning rate values were $5 \cdot 10^{-5}$, $1 \cdot 10^{-4}$, $2 \cdot 10^{-4}$ and $4 \cdot 10^{-4}$. Each experiment was repeated 10 times. In the end, smaller batch sizes gave significantly better results ($r_{pearson} = -0.4824, p_{value} = .0016 < .05$). The better results were achieved with a minibatch size of 1 or 4, without significant difference between both.

This result may be surprising because, as explained above, gradient accumulation is considered a good practice in deep learning (e.g [138], [139]). However, for semantic segmentation problems, parameter optimization favors larger image size and smaller minibatch size [140]. Thus, Long and Shelhammer [2] when they revolutionized semantic segmentation with FCN architectures, also recommended to use a batch size of 1 and to avoid sample patching. A strategy called online learning [141]. Consequently, a batch size of 1 and learning rate of $5 \cdot 10^{-5}$ will be used for the next experiments.

## 4.3 Activation function

The numerical values coming out of the neurons pass through an activation function, which transforms them into an output. On of the simplest function is the rectified linear unit (ReLu, **Fig. 4.3.1**). While the rectified linear unit is considered reliable for many neural networks, it creates a problem of dying neurons. This occurs when some neurons return negative values too often, which are set to zero by ReLu. During backpropagation, the error gradient cannot be propagated upstream, as the output value of the neuron is zero. If this occurs too often, the neuron is "dead", and cannot learn anymore. Leaky ReLu (**Fig. 4.3.1**) proposes to add a slight slope for negative values, to solve this problem.

The course of this experiment is as follows: the network was trained 10 times during 100 epochs with each configuration. The first configuration is as described in sections 4.1 and 4.2. In the second configuration, the ReLu activation function of the decoder is replaced by LeakyReLu. The number of classes is 3, the metric is mIoU, applied to the validation set. At the end of the experiment, the statistical outliers ($x \notin [\mu \pm 2\sigma]$) were removed. A Student's t-test was used for computing the $p_{value}$.

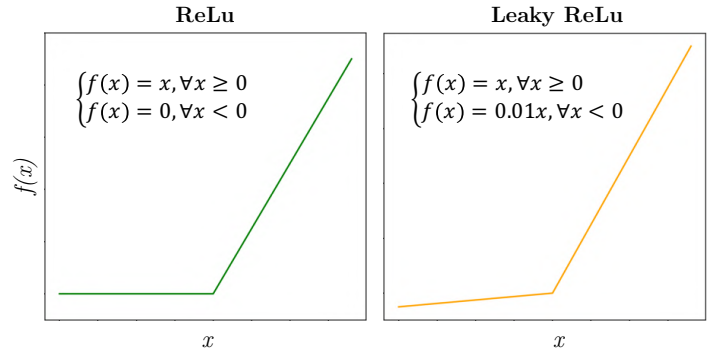

**Figure 4.3.1** Rectified Linear unit and Leaky Rectified Linear unit functions

The $H_0$ hypothesis is the following: no difference between the performance of the network when using ReLu or when using LeakyReLu. The $H_1$ hypothesis is that LeakyReLu is better than ReLu. In the end, $H_0$ is clearly rejected in favor of $H_1$ ($p_{value} = 0.0028 < 0.05$). With ReLu, the average mIoU is 0.746 versus 0.775 for LeakyReLu. We were therefore able to significantly improve network performance with this modification.

## 4.4 ResNet101

The ResNet architecture is declined in several sizes. Until now, in the frame of this work, the ResNet50 architecture, which contains 50 layers, has been used. However, it is also possible to use deeper networks, such as ResNet101 or ResNet152, which have a higher abstraction capacity and can therefore theoretically solve more complex problems. Conversely, more shallow variants also exist, such as ResNet34. Although they can sometimes improve learning, the use of deeper networks also has disadvantages, such as longer training times. The inherent increase in the number of parameters can also lead to overfitting problems. For these reasons, we will limit ourselves to testing an alternative close to ResNet50, namely ResNet101. We will evaluate the benefits of ResNet101 in solving our problem through a very simple experiment, which pits ResNet50 against ResNet101. The network as described in sections 4.1 to 4.3 was trained 6 times for each of the two encoders during 100 epochs on 300 Paris training examples and validated on 30.

At the end of the experiment, outliers were removed when they exceeded twice the standard deviation ($2\sigma$). The mean IoU was then compared using a one-tailed Student t-test. The hypothesis $H_0$ is that the mIoU is similar depending on whether the encoder used is ResNet50 or ResNet101. The hypothesis $H_1$ is that the mIoU is better when ResNet101 is used as an encoder. The experiment clearly demonstrates the superiority of ResNet101, and therefore allows to reject $H_0$, with a $p_{value} = 0.038 < 0.05$. The mIoU for ResNet50 was 0.7787 in average while the mIoU for ResNet101 reached a mean of 0.7994.

## 4.5 Pre-segmentation of the frame

The detection of the frame class, which includes all the non-cartographic content of the image, is not trivial. This is partly due to the frame lying at the scale of the map object, whereas predictions are made at the scale of the patches, thus somehow out of context. While the position and topology of the frame is very clear on the whole map, it is not so obvious on patches. Therefore, an additional annotation was performed on whole maps, which were reduced in size to 1000 vertical pixels. The annotation consisted of annotating only the frame and the map content with two different labels. The goal is to detect the frame at the map scale, rather than at the patch scale. The 256 maps of the World corpus were all reduced in size and annotated. This annotation work took 2 days, which is much less than the time needed to annotate the map patches.

The UNet was then parameterized as described in sections 4.1 to 4.3 and trained for 100 epochs on 230 map samples. The validation set consisted of 26 maps. At the end of the experiment, the mean IoU measured on the validation set is 0.8057, including an IoU of 0.7287 for the frame class and 0.8726 for the map content.

This result is comparable to what was obtained for patches. In comparison, the average IoU obtained in experiment 4.3 on the frame class was 0.776. However, one should take into account the fact that the number of learning examples is lower for this experiment as well as the fact that the annotation time needed to produce learning examples is much shorter.

Therefore, we believe that this approach of pre-segmentation of the framework is promising. A few weeks of annotation could be sufficient to obtain very large datasets, and excellent performance. We have also tried, in an experiment not detailed here, to apply transfer learning by pre-training the network on a page segmentation dataset included in dhSegment [43], without significant success.

The result of the pre-segmentation is in lower resolution compared to the map full resolution, resulting from the map size reduction. However, it could be resized and encoded as a 4th color channel, such as a transparency channel $\alpha$, to be added to the map patches. This would allow the second channel to refine the outline of the frame in higher resolution, at the pixel scale.

As the time for this work was limited and the frame segmentation was not the most central point of the work, we decided not to spend more time on annotating map frames. We therefore directly designed a validation experiment to test the feasibility and relevance of using an approximate presegmentation of the frame, encoded as a 4th color channel for images passed through the main segmentation network. For this experiment, the ground-truth labels of the Parisian corpus patches were binarized, according to the frame class. They were then reduced by a factor of 10, and increased again by a factor of 10, in order to simulate an uncertainty of 10 square pixels in the frame presegmentation, due to the necessity to reduce the map resolution in the previous step. This layer was then added as the 4th channel of the image patches.

The neural network described in sections 4.1 to 4.3 was trained during 100 epochs on 300 samples from the Paris corpus and then evaluated on 30 samples from the same corpus. While the mean mIoU obtained in section 4.3 was 0.775, the mIoU obtained for this experiment is 0.8814. The IoU of the frame goes up from 0.776 in average to 0.9948. For the following experiments, we will refer to the use of this trick as a 2+1 class, or 4+1 class problem, instead of 3, respectively 5 classes.

The idea of pre-segmenting the framework seems appropriate and the creation of generic databases for segmentation of the framework would probably be beneficial in the future. This is all the more justified as the map frame appears to be little influenced by cultural differences. As a matter of fact, it was noticed in Chapter 3 that the figuration of this class was very strongly correlated between the Parisian and world corpora. Topologically, there is not much difference either, since the position of the frame is roughly identical by definition.

## 4.6  Pretraining data and opportunities of transfer learning

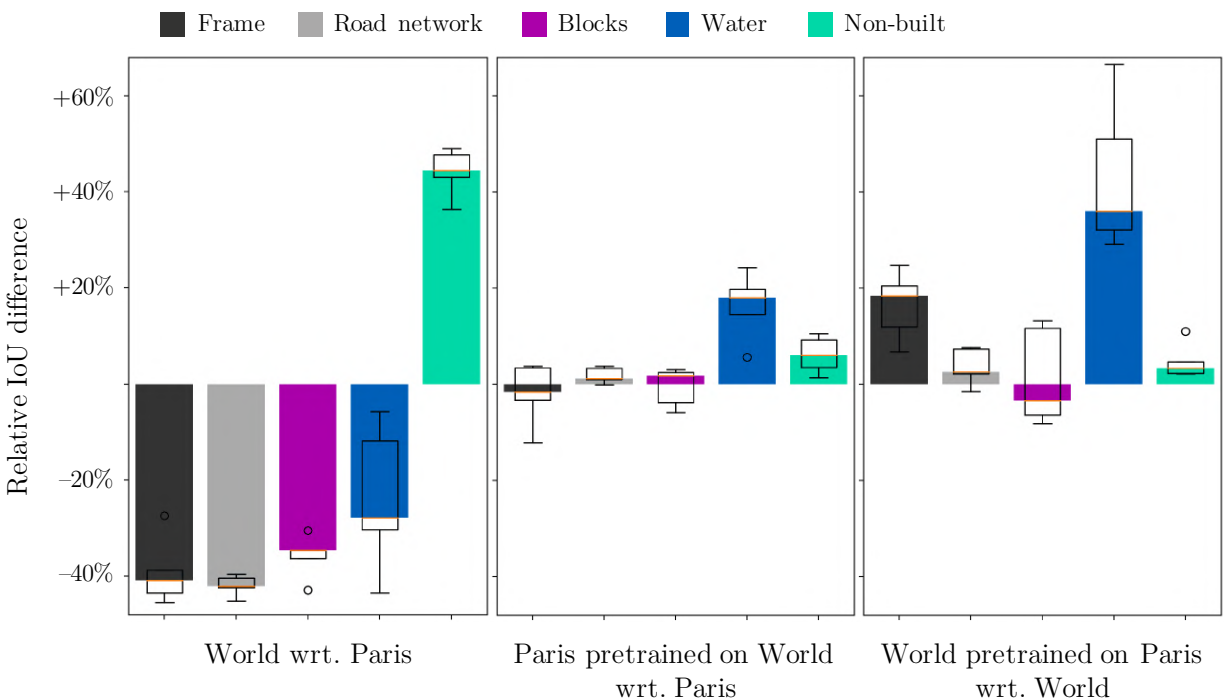

**Figure 4.6.1** Per-class relative performance of regular training between World and Paris corpora (left), and per-class relative performance of transfer learning between World and Paris corpora (middle and right). The relative IoU is computed with regard to the median of the reference IoU. When the pretraining is not specified, the CNN is generically pretrained on ImageNet. Each experiment is repeated 5 times.

The neural networks can be pre-trained on existing training sets. For CNNs specialized in semantic segmentation, the default pre-training is done on ImageNet [134], which has been used so far. Pre-training allows the encoder weights to be initialized to values that are already effective in solving problems of a similar nature. Transfer learning, in the same way, seeks to pre-train the networks on a closely related task. For instance, if a researcher wishes to train a neural network to recognize wolves, he could use an existing and potentially larger set, designed for dog recognition, as a pre-training set and use transfer learning to solve his own problem. For map processing, the learning transfer opportunities are less obvious. One might think of aerial images datasets. However, the basic features and textures, encoded in the first layers of the neural networks, are very different between a map and an aerial photography. Alternatively, Ignjatic [37] suggests using Liu's [42] training set for segmenting apartments rooms. In the present case, two separate sets for world maps and Paris were already annotated. As it appears that these are probably the two sets at our disposal that are the most relevant for a learning transfer, we will use them to feed each other.

First of all, the network presented in chapters 4.1 to 4.2 was trained during 100 epochs on the Paris dataset, and separately on the world. The improvements presented in section 4.3-4.4 were not included, as they occurred chronologically later than the present experiment. The improvement presented in section 4.5 was not included because it would confuse the performance on the frame class for this qualitative and comparative experiment. After the first training, the network was re-trained on the Paris set, using the weights trained in the previous step on the world corpus as initialisation. Reciprocally, the world set was re-trained by initializing this time the weights on the Paris corpus. Relative performances (mIoU) were then compared. The results are shown in **Fig. 4.6.1**.

Class recognition performance shows significant disparities between the two datasets. If, in general, the Parisian corpus achieves much higher performances, the world corpus seems to be better at recognizing the non-built lands. This will be further investigated in the next chapter.

The transfer learning is very successful for the World corpus, when pretrained on Paris. The overall increase in mIoU is not yet significant according to conventional criteria ($p_{value} = 0.062 > 0.05$), but the probability of a positive impact is very high. The water ($p_{value} = 0.0063$), and to some extend the frame ($p_{value} = 0.060$) and the non-built (0.062) in particular seem to be better recognized. The transfer learning from the World corpus to Paris is a bit less successful. The only noticed improvement in performance is for water class (0.0076).

Overall, learning transfer seems to be successful. The most significant improvement occurs for the water class, which is also the less represented class in both training sets (**Tab. 2.4.3**) and particularly for the Parisian corpus where it represents only 2.6% of the set. The pre-training therefore probably compensates for an unsufficient representation. The pre-training of the world corpus on the Paris corpus also allows to clearly improve the frame recognition. In this case, the class is well-represented for both datasets. However, we had seen that the frame class was figuratively the closest class between the two corpora (**Fig. 3.4.1**).

Moreover, the Parisian corpus obtains in average much better performances (+67% mIoU) on this class. This may explain why the transfer is so efficient and why it is working only one-way.

The decoder weights are not pre-trained. However, there are a number of initialization functions existing and not all of them give equally good results. In an experiment that will not be described extensively here, we compared five layer weights initializers, namely Saxe et al orthogonal [142], He et al [143] normal, and uniform, as well as Glorot and Bengio [144] normal, and uniform. The best performances were obtained through a uniform weight initialization proposed by Glorot and Bengio.

## 4.7 Benchmark performance

To achieve the best performance, the advances in sections 4.1 to 4.6 were combined. The settings will now be briefly summarized. Neural networks use a UNet architecture, with ResNet 101 as encoder and a LeakyReLu activation function for the decoder. The data are augmented by side flip and upside-down flip, and with a rotation $r \in [0, \pi]$. The learning rate is $5 \cdot 10^{-5}$, for a batch size of 1. The loss used is cross entropy. The optimisation relies on stochastic gradient descent (SGD).

In this experiment, 4 training sets were used. The first two sets, the Parisian sets, consist of 300 training examples and 30 validation examples. The last two sets, from the world corpus, are made up of 256 training examples and 49 validation examples. For Paris as for the world, two types of ontologies were tested: 2+1 classes and 4+1 classes. The frame of the maps (+1) was pre-segmented with an approximate error of 10 pixels.

The training was carried out in two stages. In the first step, each network, pretrained on ImageNet, was trained for 150 epochs on one of the 4 sets. In a second step, the networks were trained again during 150 epochs, by initializing the weights on the models of the previous step, in a crossed way (on Paris for the world and on the world for Paris).

Four metrics were used to quantify performance: IoU, accuracy, precision and recall. Accuracy indicates the proportion of correct predictions. Precision is an index of the confidence of the prediction. It indicates the proportion of a class that has been correctly predicted, among all pixels that have been predicted as belonging to that class. Conversely, recall indicates sensitivity. It is used to check, among all the pixels actually belonging to a class, the proportion of pixels that could be identified. The numerical results are presented in **Tab. 4.7.1**. In a second step, the confusion matrices between the different classes were computed. These were normalized with respect to the proportion of pixels belonging to the class, according to the ground-truth (denominator of the recall). These matrices are shown in **Fig. 4.7.2**. Examples of segmented map patches from the validation sets are also shown in **Fig. 4.7.4** at the end of the section.

**Table 4.7.1** Performance achieved, per class and classes mean, on the two datasets, for 2+1 and 4+1 classes problems

| Metric | Class | Paris 2+1 | World 2+1 | Paris 4+1 | World 4+1 |
| --- | --- | --- | --- | --- | --- |
| IoU | **Mean** | **0.8905** | **0.8055** | **0.6363** | **0.5595** |
|  | Frame | 0.9953 | 0.9924 | 0.9810 | 0.9881 |
|  | Blocks | 0.9181 | 0.9114 | 0.5657 | 0.3559 |
|  | Road Netw. | 0.7580 | 0.5147 | 0.7132 | 0.4682 |
|  | Water | – | – | 0.4682 | 0.3318 |
|  | Non-built | – | – | 0.4235 | 0.6538 |
| Accuracy | **Mean** | **0.9624** | **0.9556** | **0.8392** | **0.7986** |
|  | Frame | 0.9992 | 0.9985 | 0.9909 | 0.9964 |
|  | Blocks | 0.9438 | 0.9336 | 0.7495 | 0.8173 |
|  | Road Netw. | 0.9441 | 0.9348 | 0.8982 | 0.8211 |
|  | Water | – | – | 0.9452 | 0.6471 |
|  | Non-built | – | – | 0.6123 | 0.7113 |
| Precision | **Mean** | **0.9292** | **0.8544** | **0.7292** | **0.6986** |
|  | Frame | 0.9959 | 0.9935 | 0.9838 | 0.9893 |
|  | Blocks | 0.9689 | 0.9730 | 0.6872 | 0.4353 |
|  | Road Netw. | 0.8229 | 0.5967 | 0.7874 | 0.5348 |
|  | Water | – | – | 0.4856 | 0.7098 |
|  | Non-built | – | – | 0.7017 | 0.8240 |
| Recall | **Mean** | **0.9456** | **0.9062** | **0.8175** | **0.7187** |
|  | Frame | 0.9996 | 0.9989 | 0.9971 | 0.9988 |
|  | Blocks | 0.9448 | 0.9350 | 0.7618 | 0.6611 |
|  | Road Netw. | 0.8924 | 0.7848 | 0.8832 | 0.7898 |
|  | Water | – | – | 0.9288 | 0.3839 |
|  | Non-built | – | – | 0.5165 | 0.7599 |

The clearest finding is the impact of the increase in the number of classes. This difference is also visible in the result examples shown in **Fig. 4.7.4**. While the "complete" ontology contains 5 (or 4+1) classes, the performance drops massively when increasing from 2+1 to 4+1 classes. The resulting proportion of misclassified pixels (1-accuracy) is multiplied by 4.3 for Paris and by 4.5 for the world. The world corpus also gets poorer global performances compared to the Paris corpus. For the 2+1 classes problem, this is mainly expressed by a worse recognition of the road network. Recall and most importantly precision are much lower for the world set.

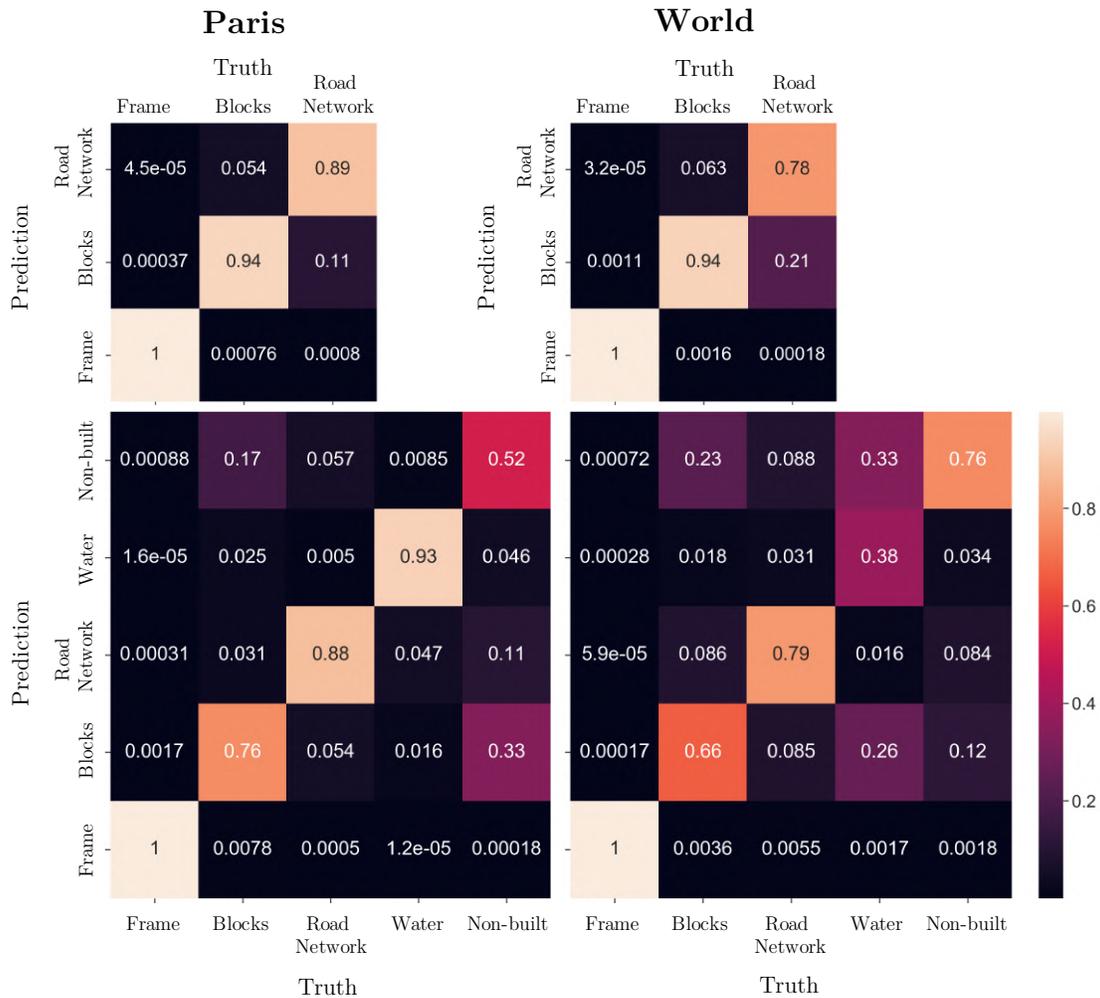

**Figure 4.7.2** Confusion matrix, normalized according to ground-truth, i.e TP + FP (true positives + false positives). The diagonal correspond to recall.

For the 4+1 classes problem, the differences are more alternate. While the IoU for the block, road network and water classes is better for Paris, the IoU for the non-built class is much better for the world. This was already noticed in the previous section. It is expressed here as a concomitant collapse of precision and recall for this class.

The drop in performance between the 2+1 classes problem and the 4+1 class problem is not equal for all classes. While the road network remains almost stable, performance collapses for the blocks. In particular, it is the recall, i.e. the ability to recognize this class, that drops. Confusion matrices can be used to investigate these performance differences. One can notice that the proportion of confused pixels remains almost identical for the road network: about 11% of the pixels for Paris and 21% for the world, which is not very high. The average confusion of the road network by other classes is also stable. For Paris, the water class, when added, also benefits from a very low proportion of confusion. On the other hand, the non-built is correctly identified in only 52% of cases. Nearly a third of the pixels representing the non-built are wrongly attributed to the blocks. This is the biggest source of confusion for Paris. Reverse confusion from blocks to non-built also exists, but to a lesser extent (17%). For the world, water is the most confused class. This is problematic because this class is more represented in the world corpus, compared to the Paris corpus. Indeed, water occupies about 12% of the surface of the world maps compared to 2.6% of the Parisian maps (**Tab. 2.4.3**). For the world, water is correctly identified in only 38% of cases. It is confused in a third of cases with non-built and in a quarter of cases with urban blocks. Surprisingly, while the non-built is very easily recognized for the world (76% recall), pixels belonging to other classes tend to be wrongly attributed to it. 9% of pixels representing the road network and 23% of pixels representing the urban blocks are wrongly identified as non-built. This may indicate an imbalance in the classification, which tends to over-represent the non-built.

Despite these limitations, performance remains very exciting. In the 2+1 classes problem, the average accuracy per class, i.e. the average of correctly classified pixels per class, reaches 96.2% for Paris and 95.6% for the world. In the Parisian maps, 94.5% of the pixels representing blocks are recognized, and the classification precision for this class is high, at 97%. The IoU of road network recognition reaches 75.8%, whereas Chiang et al [123], for a binary railroad recognition problem, and on an extremely homogeneous and clean map (see **Figs. 3.2.5** and **3.2.6**), only achieved 62.2%, with a training corpus of 4800 examples, whereas ours contains only 300 examples at most. The comparison with Oliveira et al [11] is more complicated because they calculate precision and recall at the cadastral plot level and not at the pixel level. Under a threshold of 50% of the mIoU, the plot recognition precision they obtain is 0.557, while the recall is 0.944. Under the same threshold, the precision per pixel of our model trained in Paris for the recognition of urban blocks is 0.977, and the recall is 0.974. We thus demonstrate a real technological advantage compared to previous studies, moreover on much more diverse data.

The analysis of performance below a threshold, as Oliveira et al do, is relevant because maps can be completely misinterpreted, as shown in the following section. Thus, there is a long imperforming tail. Moreover, performance thresholding allows to estimate the industrialization capacity of the model on the part of the maps that work best. For the best half of the maps, i.e. for the maps whose mIoU exceeds the median mIoU of our validation sets, the precision and recall are detailed in **Tab. 4.7.3**. Under this condition, the percentage of misclassified pixels (1-accuracy) is only 2.65% for Paris and 3.56% for the world.

**Table 4.7.3** Achieved performance, per class and classes mean, on the best half (mIoU > median(mIoU)) of the two datasets, for 2+1 classes problems

| Metric | Class | Paris 2+1 | World 2+1 |
|---|---|---|---|
| Precision | **Mean** | **0.9679** | **0.9205** |
| | Frame | 0.9969 | 0.9935 |
| | Blocks | 0.9774 | 0.9816 |
| | Road Netw. | 0.9295 | 0.7863 |
| Recall | **Mean** | **0.9456** | **0.9445** |
| | Frame | 0.9990 | 0.9992 |
| | Blocks | 0.9736 | 0.9957 |
| | Road Netw. | 0.9368 | 0.8686 |

These results highlight a very promising lead group. On at least half of our corpora's maps, the block vectorization process, now almost entirely manual, would probably already be partly automatable. Such an automatization would make it possible to segment thousands of maps in a few days, whereas for large cities, it currently takes several weeks of manual work to vectorize a single map. However, these results may still require manual correction, as can be seen in the images in **Fig. 4.7.4**. Moreover, the use of the 5-classes ontology, or 4+1 classes, which we consider the smallest complete ontology, still requires further progress. Indeed, confusion between classes remains too high. We will discuss this point in more detail in the conclusion chapter, where we will also relate the performance obtained here to the findings on figuration from the previous chapter.

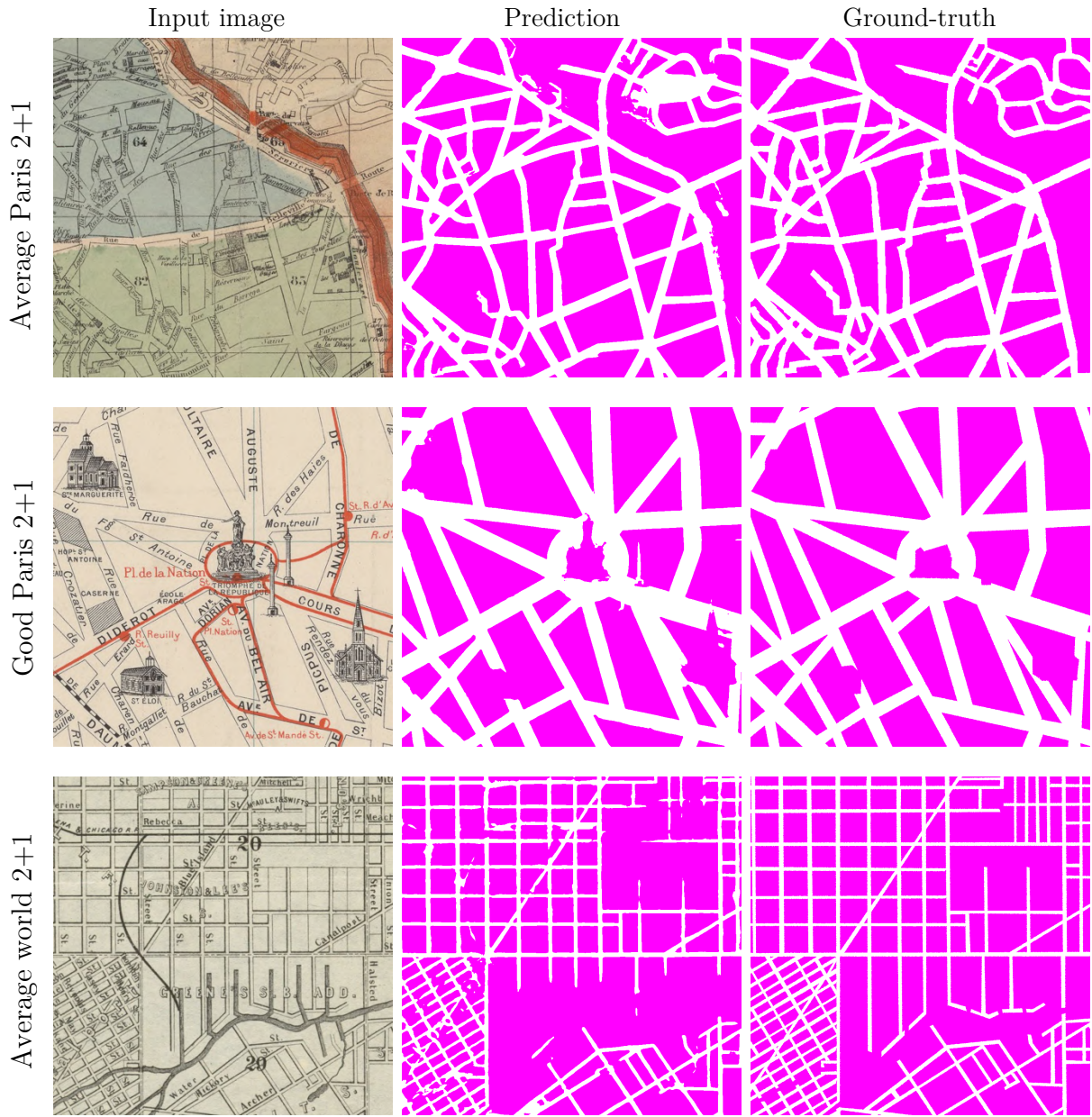

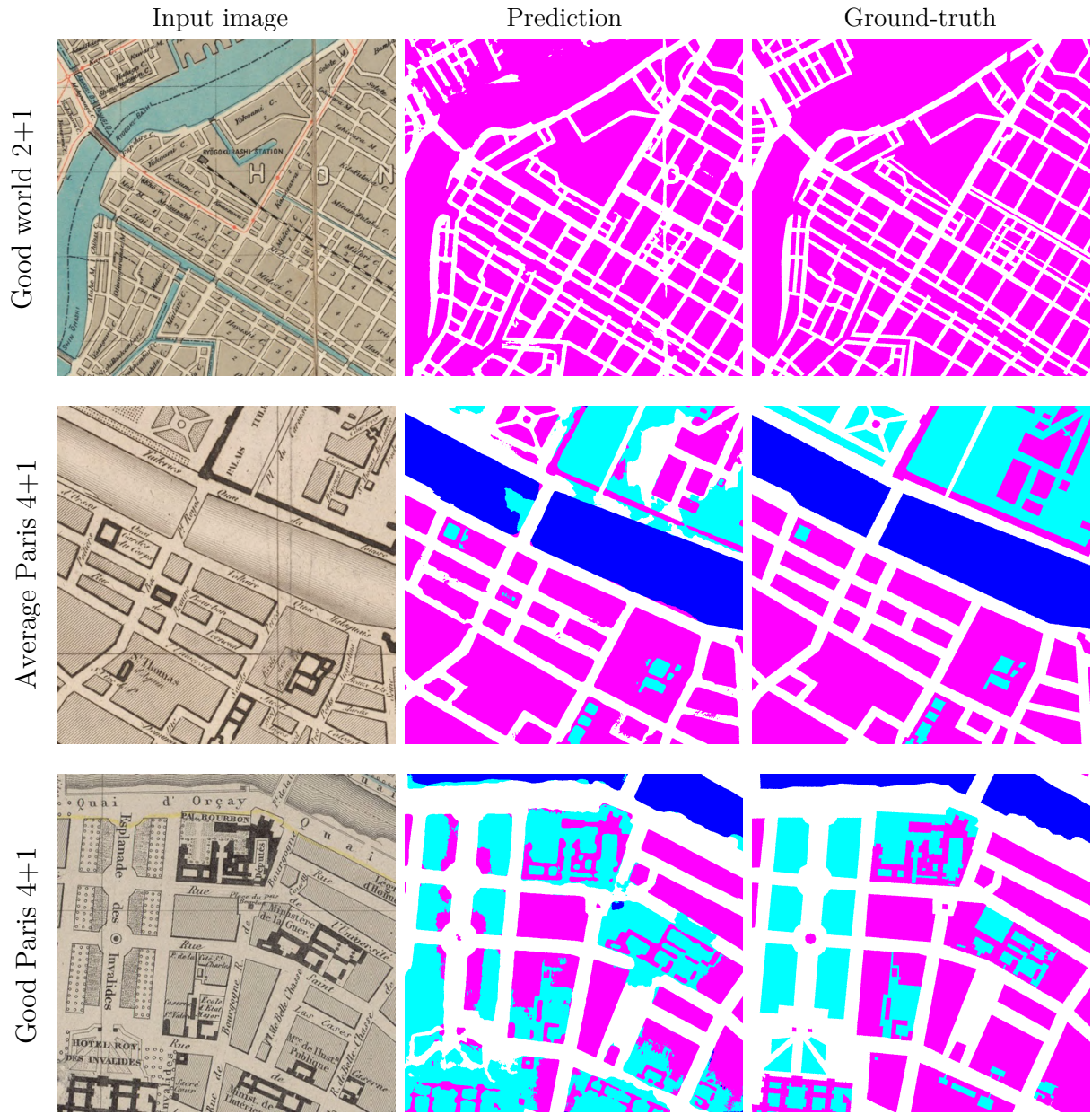

**Figure 4.7.4** Examples of results. See World 4+1 in the Figs 4.8.1 and 4.8.2. Each time, the first example is close to the median mIoU and the second is part of the better half. Values of the mean IoU from top to bottom : 0.8965, 0.9506, 0.8079, 0.8622, 0.6091, and 0.7934. [68], [145]–[149]

## 4.8 K-fold cross validation

In order to estimate the bias of the validation set, and to investigate the cross-cultural performance, a cross validation was performed on the World 4+1 dataset. The parameters used are as in section 4.7, except for the number of epochs, which was decreased to 100, as it was noticed that this already allows convergence. For the k-fold cross validation, the training and the validation sets are reunited together, and randomly shuffled. In the first iteration, the first 1/k portion of the reunited set is assigned to validation set, while the remaining (k–1)/k portion is considered to be the training set. The network is trained and validated. Then, in the second iteration, the validation is set to the second 1/k portion of the reunited set, while all the remaining samples as used for training. Etc. In total, the network is trained k times and all samples are used once as validation and k-1 times as training sample. The advantage is that the bias in the constitution of the validation and training set is minimized, as the method allows to successively validate on the totality of the samples found in the set.

In the present case, k=8, meaning that the network was trained 8 times using $1/8^{th}$ of the set as validation and $7/8^{th}$ as training, iteratively. In total, the mIoU over the 8 experiments is 0.6112, which is noticeably better than the 0.5595 score obtained in the previous section for the same 4+1 world set. That means that the average performance is likely slightly better than the previously observed performance, meaning that the regular validation set of the world corpus may be somehow more difficult to segment than the average sample.

The mIoU can also be computed on each patch separately, which corresponds to an average of 0.5424. The average mIoU on the 5 images that were augmented by super-resolution was also computed and compared with the latter, in order to estimate the impact of the super-resolution step. The resulting mIoU is 0.2374. Further experiments would be needed to validate this result, since the number of samples is low, and since there is no valid control set presently. However, it seems that this preprocessing step had a negative impact on the performance for those patches.

As the title of this work indicates, cross-cultural performance is very important for understanding the generalization potential of neural networks. One can notice here that maps that have been at least co-published by a Western country score 0.5608, while other maps score 0.4911. The region of the city represented is also impactful, with Subsaharian (0.6322) and North African (0.6280) cities scoring best, followed closely by Eastern Europe and Central Asia (0.5931), South America (0.5913), Western Europe (0.5891), and North America (0.5711). At the end of the line are the South Asian cities (0.3985). In the middle, one would find Middle East (0.5048), East Asia (0.4896), Oceania (0.4685), and Central America (0.4635). The urban form also has a strong impact, as cities with a more regular (0.5453) or mixed (0.5535) urban form score much better than cities with an irregular (0.5088) urban form. This performance drop is also especially noticeable on the block class for regular (0.3548), mixed (0.3027), and irregular (0.2179) urban form.

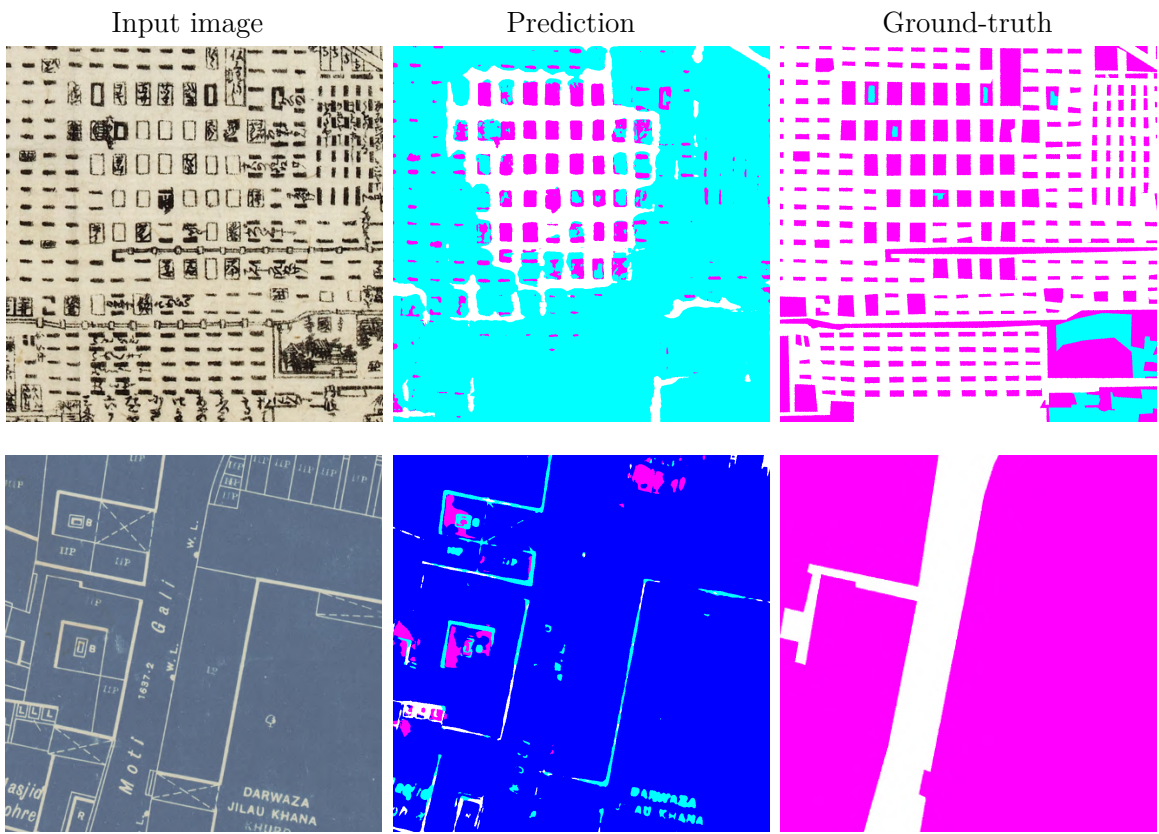

**Fig. 4.8.1** Examples of confusing figurations [100], [150].

These results lead to some very important findings. First, the poor performance obtained in the end on images that have been subjected to a super-resolution network pleads for the use of images with a naturally good scanning resolution. Images with poor resolution might therefore probably not be used in meta-analyses. In general, the easiest city maps to segment seem to be the Western colonies (see **Fig. 2.3.2**, left), which is consistent with the ease of segmenting maps of regular cities. The more recent African colonies are particularly easy to segment. We recall that they are essentially South African cities. The same good performance is observed for Slavic cities, which were also sometimes built more recently. Asia, on the other hand presents more problems. The difference in performance between East Asia and South Asia is probably also influenced by urban form, which at the time tended to be regular in East Asia, but irregular in South Asia (mainly India), influenced by the Islamic world. As a result, South Asia suffers from two complexity factors. Several reasons could explain these poor performances. The condition of the Asian maps in the corpus may be worse than average. However, we noticed in Chapter 2 that several Asian maps were incongruously figured, although in the corpus these uncommon figuration conventions were often shared between several maps of this region (**Fig. 2.3.6**). These particular figurations can be hard to interpret for neural networks (see **Fig. 4.8.1**).

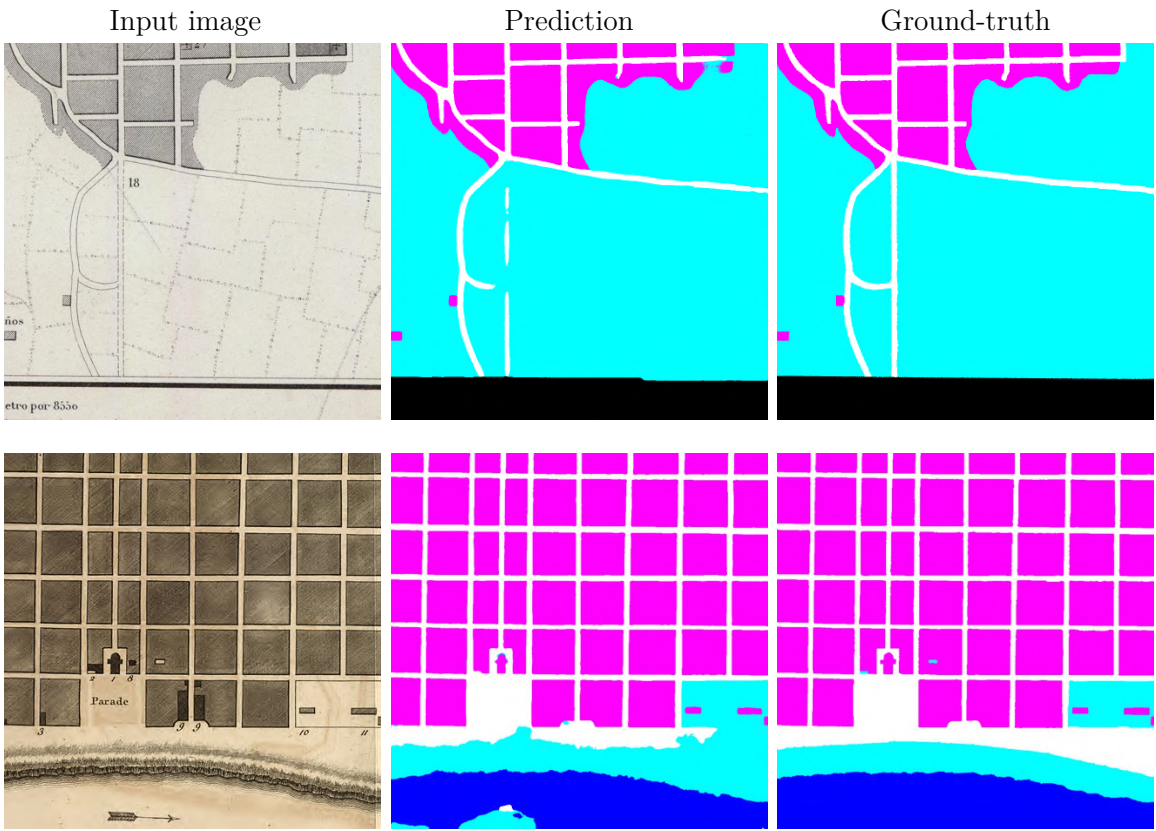

**Fig. 4.8.2** Examples of congruent figurations [151], [152].

The representation of Asian maps in the corpus, on the other hand, is probably not in question (**Fig. 2.3.2**, right). We are therefore not dealing with a fundamental Western bias, but more probably with a map grammar that is more difficult to segment.

The investigation of the underperformance obtained for this type of map could be the subject of a much more complete study, which will not be carried out here, however. Surprising is also the poor performance of the Oceanian, mainly Australian, maps. In this case, the low representation of the region (less than 3%) could be a cause, and simply create a measurement bias.

## 4.9 Confidence prediction

Establishing a confidence index is a challenge in deep learning. Indeed, it is often difficult to estimate the quality of the output. We have shown in section 4.7 that the performances reached industrial performances on the best maps. However, how can one identify these promising maps, since neural networks do not a priori provide any confidence assessment? For the problem of semantic segmentation, theory and scientific literature are lacking. We have therefore attempted, in a series of highly experimental trials, to create an estimator of the confidence of prediction, at the patch scale.

In the first point, the prediction error can be measured for each patch using cross-validation, as implemented in the previous section. In the present case, k-fold cross validation was performed on the Paris 3 classes set, with k=10. The network described in chapters 4.1 and 4.2 was trained during 60 epochs. The reason for this setting is that these experiments took place chronologically before section 4.3 and following. The average mIoU over the 10 folds is 0.6993. The output of the network is compared to the ground-truth and an accuracy map, in which the pixels take the value 1 if the prediction is correct and 0 if the prediction is wrong, is created. The use of k-fold cross-validation ensures that all accuracy maps are created on patches used for validation.

Logically, the first experimental setting attempted to use the figurative descriptors from Chapter 3. We will not describe this experiment here, as the machine learning algorithms applied showed no predictive power during validation.

In a second step, a UNet network as described in sections 4.1 and 4.2 was trained on the 3-classes Paris dataset. Deep features were extracted from its trained ResNet encoder and were used as descriptors. Classical machine learning models were then deployed. Alternatively, an approach reintegrating the decoder for a new training phase was also tested. In both cases, once again, no true predictive power was observed.

The third experimental setting, although much simpler, proved to be the best. In the latter, a second network, identical to the first one, was simply trained on 300 pairs of images and accuracy maps, and validated on another 30 pairs. The goal of this network was to predict the accuracy map corresponding to the image. The predictor is defined as the patch accuracy, and the reference as the accuracy previously measured by k-fold cross-validation. Instead of applying the argmax operator on the output of the neural network, a specific and global threshold was determined with respect to the percentile corresponding to the average accuracy of the reference accuracy maps.

For this third setting, the Pearson correlation between the obtained predictor and the reference is 0.571 ($p_{value} = 1.2 \cdot 10^{-3}$), on the validation set, which represents an intermediate to high dependency. One can thus consider that, for the problem investigated in this work, and potentially for semantic segmentation problems in general, a neural network can provide a reasonably good estimate of the accuracy of the predictions delivered by another neural network, and thus establish a reliable confidence index. This result would benefit from being validated on larger datasets, and in other contexts. However, it represents a very interesting finding. In the present case, the confidence index could be used to discard upstream the maps that are unlikely to be segmented correctly, or on the contrary, to select the maps of a city that are most likely to be correctly vectorized. Other application perspectives exist, e.g. for active learning.

## 4.10 Active learning

We therefore investigated whether the confidence index could be used in an active learning context [153]. Active learning is a machine learning strategy that aims to optimize the number of learning examples required by choosing them properly to fill in the learning gaps of the network. This scheme can be particularly relevant for problems that require time-consuming data annotation, such as the one we are dealing with. The idea of active learning is that the network itself asks a teacher for the ground truth when the prediction of one of the samples seems uncertain. The teacher is typically a human annotator. This can be done iteratively, one sample after another, or in larger steps. The development of efficient active learning strategies could be very beneficial in map processing, to refine the performance of a generic network on a new corpus of maps by automatically choosing a set of optimal learning patches. In general, the patches considered optimal are those on which the error is potentially the highest. Various strategies for estimating the confidence index are therefore developed.

The experimental setting is as follows. The neural network, as described in sections 4.1 to 4.3 has been trained 10 times during 100 epochs on 4 different training sets, to solve a 3-classes problem. The validation set in the contrary is invariable. It simply consists of the 30 Paris validation samples. The first training set (set 200) is the reference. It consists of the first 200 patches of the Paris training set. The second training set (set 200 + 50 random) is also used as a reference. It consists of the first 200 patches of the Paris training set, and of 50 randomly selected patches from the remaining 100 patches. The random selection varies for each repetition of the experiment. At this step, the network is trained with the first configuration. Indeed, the 3$^{rd}$ and 4$^{th}$ sets were drawn from worse, respectively best results of the network trained on the 1$^{st}$ set. More precisely, the 3$^{rd}$ set (set 200 + 50 worse, or active learning set) was made up of the first 200 patches of the Paris training set, and the 50 patches (out of the remaining 100) that had obtained the worst mIoU. Finally, the 4$^{th}$ set (set 200 + 50 best, or anti-active learning set) consisted of the first 200 patches, plus the 50 patches that had obtained the best mIoU.

The outcome of the experiment is undetermined. Unfortunately, no statistically significant difference was found between the 4 conditions when using a one-tailed Student t-test, with a threshold $p_{value}$ at .05. Outliers were removed when they exceeded twice the standard deviation ($2\sigma$). The mean (median) mIoU by set is 0.735 (0.751) for the set 200, 0.743 (0.746) for the set 200 + 50 random, 0.751 (0.766) for the set 200 + 50 best, and 0.748 (0.754) for the set 200 + worse. Overall, no trend is observed. Active learning might not be very effective at small-scale on our model.

These results are not very encouraging about the potential of active learning in this research field. However, further experiments using different selection metrics, such as pixel-level metrics (e.g. accuracy) can be carried out in the future. A limitation of the present experiment setting is the low choice for the active learning set. Indeed, the choice is restricted to the already annotated samples and half of them must be selected. In theory, it would be possible to use the confidence prediction established earlier to identify, on all the maps, the most relevant patches, extract them and annotate them. However, the correlation obtained previously still has a margin of error, compared to ground-truth. Moreover, the annotation process would be time-consuming. Other variants of active learning, such as single-pass [154] active learning, in which only one new sample is added at the time might also be considered.

## 4.11 Curriculum learning

Another similar approach is curriculum learning [155]. Differently from active learning, curriculum learning proposes to train the neural network starting with the simplest examples. One way to do this is to sort the batches from the simplest to the most difficult, and then present them to the neural network in that order. Intuitively, this allows the CNN to first learn the basic examples before generalizing with more complex examples.

This approach was tested on the world dataset, sorting each cross-validated patch by mIoU thanks to the cross validation performed in section 4.8. The patches normally belonging to the validation set were removed from the cross-validated patches and used again for validation. The network described in section 4.7 was trained 3 times, during 100 epochs, with and without curriculum learning.

The experiment was validated with a Student t-test. $H_0$ is that the performance is similar with or without curriculum learning. $H_1$ is that curriculum learning is better. In the end, $H_0$ could clearly not be rejected ($p_{value} = 0.74$). Moreover, curriculum learning is scoring worse, with a mean drop in mIoU of –0.05. Other sorting procedures and metrics could be considered. However, this low impact was expected, since learning transfer was shown to be efficient. Indeed, the gist of curriculum learning is to roughly direct learning towards a minimum, before refining performance with more complex examples. In this case, the pre-training step already allows to steer learning in the right direction. Moreover, the unremarkable effect of active learning, which was based on similar principles was foreshadowing a failure for curriculum learning.

## 4.12 Size of the training set

In order to estimate and validate the impact of the size of the training set on performance, a series of experiments was performed on the benchmark performance model (section 4.7) by gradually resampling the Paris and the World training set, from 300 samples in normal times to 60, in increments of 60. The network was then trained separately 100 epochs on each resampled training set. The experiment was reproduced with 5 different seeds for each step, where seeds were used both for resampling and for initializing the network parameters.

The experiment was also reproduced identically on the Paris dataset using a simpler model, namely the model described in sections 4.1 and 4.2.

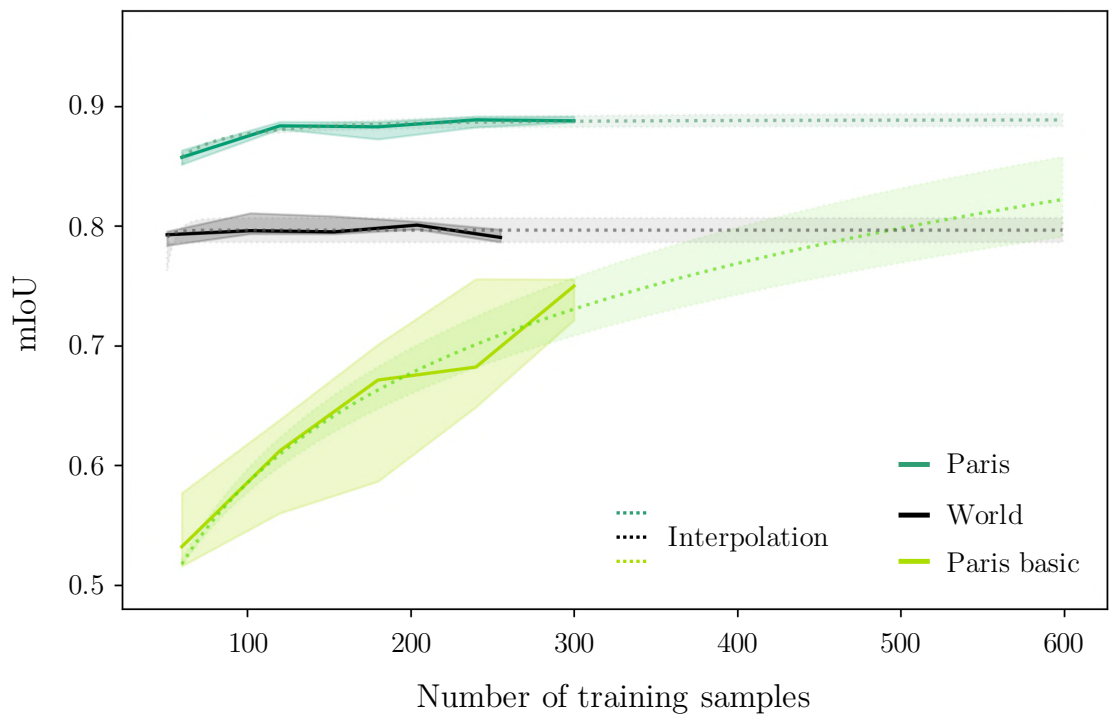

**Figure 4.12.1** Extrapolated impact of the number of training samples on the mIoU on a 2+1-classes (respectively 3-classes) segmentation problem, for Paris and the World (and again Paris, with a more basic segmentation model). The plain area boundaries are the minimum and maximum mIoU over 5 experiments, and the plain line is the mean mIoU over the 5 experiments. The dotted lines are the power law interpolations $f(x) = b + a \cdot x^{-c}$ and the corresponding areas with dotted boundaries are the confidence areas for $a, b, c \pm 5\%$. For Paris, $a = 70.06, b = 0.1110, c = 1.884$, for the World, $a = 9.312 \cdot 10^{15}, b = 2.185 \cdot 10^{-17}, c = 10.67$, and for Paris basic, $a = 942.5, b = -941.5, c = 1.406 \cdot 10^{-4}$

Subsequently, the mIoU was computed on the test data, and interpolated with an extended power law function $f(x) = b + a \cdot x^{-c}$ and linear item weighting as described by Johnson [156]. Thanks to the interpolation/extrapolation model developed by Johnson, it is possible to estimate the number of annotations needed to achieve a performance target. Tested on 4 distinct machine learning problems, with training sets of barely 10% of the total, Johnson's method allowed to extrapolate the performance with an average root-mean-square (RMS) error $< 0.05$, and often much lower. The **Fig. 4.12.1** shows the performance curves for the 3 experiments: Paris, the World, and Paris using the basic segmentation model.

As one can notice, the impact of the size of the training set on performance is crucial for the basic model. For the latter, 300 training samples are not enough to achieve maximum performance. However, the improvements made to the model in sections 4.3 to 4.6 result in a very marked increase in performance, to the point where the predictive power reaches a plateau. This is observed for Paris and for the world. Another striking feature is the distribution, which is reduced for the benchmark model, indicating less uncertainty in extrapolation.

The composition of the training set remains an important parameter and it should not be forgotten that CNNs can sometimes have unpredictable behaviors. However, if the extrapolation model is correct, a performance plateau will be reached at a mIoU of circa 0.9 for Paris and 0.8 for the world. This is good news as it means that the model developed is likely to exploit the data optimally. It also means that, mainly thanks to learning transfer, optimal performance can be achieved with as few as 100 training examples for new corpora. On the other hand, it also implies that new technologies and models will have to be deployed to improve performance in the future, as additional training data may not be sufficient per se. Moreover, it seems that the greater cultural diversity of the corpus world does not allow the latter to progress better or to learn more than the corpus Paris.

## 4.13 Impact of texture and color

The research questions we have asked address the impact of figuration on CNN's performance. The following experiment aims to determine what role color, texture and morphology really play in neural networks performance.

The images of the training and validation sets were subjected to 4 different treatments that will be called reference, gray, binary, and textureless binary (see **Fig. 4.13.1**). The CNN was trained for 60 epochs separately with each of the 4 variants, for 5 times. For the reference, the images were not modified in any way. The second treatment was the gray treatment. In the latter, the RGB color channels of the image were transformed into a grayscale. For the third treatment, the images were transformed into a grayscale, then binarized by Otsu thresholding [117]. For the fourth treatment, the images were transformed into a grayscale, binarized, and texture were extracted by LBP [113], with a radius $r=3$. Finally, a second Otsu thresholding was applied.

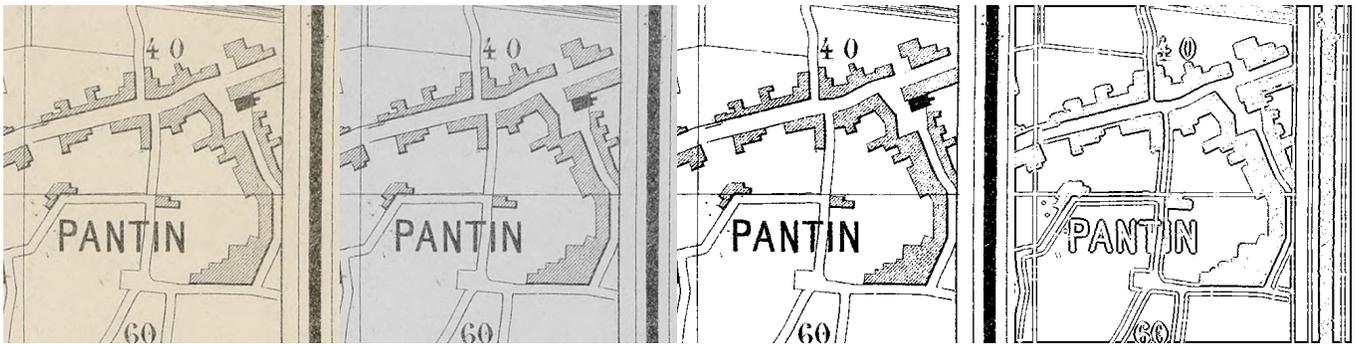

**Figure 4.13.1** Images processed by phasing out visual characteristics. From left to right: 1) reference, 2) gray, 3) binary, 4) textureless binary

The results were evaluated on the validation examples, using the mIoU metric, and are presented in **Fig. 4.13.2**. The removal of color had noticeably almost no impact on performance. The median loss is only 1.3%. The binarization of the values resulted in a drop of 7.2%. Finally, the disappearance of colors and textures led to a 10.4% decrease in performance.

One can conclude from this experiment that color and texture have a minimal impact on the performance of neural networks. Removing all the colors results in almost no impact. This is justified, as explained in the introduction, by the fact that the majority of the maps were printed in gray tones. The color is rare, sometimes added a posteriori with the brush and the explanative potential of hue is thus poor for the neural network. The binarization of the image has slightly more impact. This is due to the loss of grey levels, and therefore of value. In addition, binarization can also have an impact on the texture, which often also includes grey levels. The last step removes most of the remaining texture information. Yet performance integrity is maintained at nearly 90%. Thus, morphology and topology seem to account for most of the performance of the neural network. The latter can probably even compensate part of the loss of information with these other cues. The possibility to reach such high performances (here a mIoU of 0.673 with 3 classes) without using color or texture, while the power of the model is not yet fully mobilized in this experiment, also shows the incredible potential of that family of algorithms in hierarchizing spatial information itself and thus offer great robustness when facing high figurative diversity.

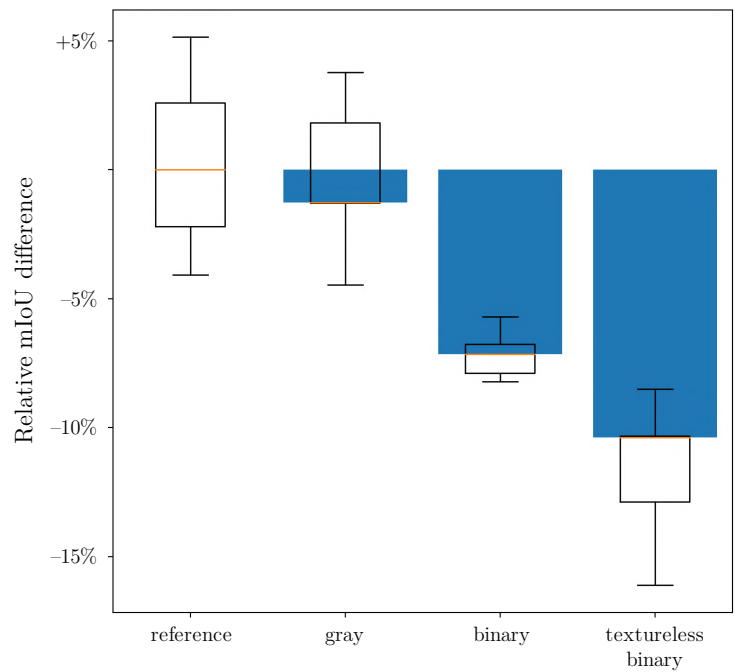

**Figure 4.13.2** Relative loss in performance (mIoU) due to the removal of visual characteristics. Each experiment is repeated 5 times. The median reference mIoU is 0.7515

# 5 General discussion

In this discussion chapter, we will try to link together the elements brought in through the different experiments and answer the two last research subquestions. In particular, we are going to study the links between Chapter 3, which mainly concerned the analysis of figuration, and Chapter 4, which presented a semantic segmentation model for urban maps. This is the first point we are going to discuss.

## 5.1 Impact of figuration on segmentation

To elaborate the link between the many elements of this work, we will begin by summarizing all the results that provide clues and keys to explain performance. The first of these is the proportion of areas covered by the different classes in the training examples (**Tabs. 2.4.3** and **2.4.4**). This factor can have a very strong impact on performance, depending on whether certain classes are under-represented or over-represented. The second element is the figurations diversity for the two corpora, in particular the diversity by class (**Figs. 3.3.3** and **3.3.4**). The third element considered, closely related to the second, is the inter-class and inter-corpora correlation of the visual features histograms (**Fig. 3.4.1** and **Tab. 3.4.2**). This correlation matrix is capital because it allows one to estimate, to a certain extent, which classes share the same figurative codes. From Chapter 4, we will mention the impact of learning transfer by class (**Fig. 4.6.1**), as well as, of course, the benchmark performance of **Tabs. 4.7.1** and **4.7.3** and the associated confusion matrix (**Fig. 4.7.2**). Finally, we are going to examine those results from the perspective of **Fig. 4.13.2**, which evaluated the impact of different visual cues on neural network performance. All the figures above mentioned can be found in the core text but are also repeated at the very end of this work, in Annex 1, so that they can be easily detached to follow the discussion.

The first highlight was the non-built class, which scores much better for the world (0.6528), compared to Paris (0.4235), even though the world generally scores lower. This could be explained by a higher areal representation (36% for the world compared to 19% for Paris), however the difference is not sufficient to justify such a discrepancy. Moreover, the figuration of this class is rather strongly correlated to the other classes, for the world (in average $r_{pearson} = 0.924$). The difference in performance can therefore not be explained by a figurative identity that stands out from the others. Moreover, it had been verified that the figurative convergence of this class was weak (median $\kappa = 3.13$, mean $= 7.72$), compared to the other classes (overall median $\kappa = 4.19$, mean $= 9.10$), for the world corpus. This class is somewhat of a special case, since it is the "remaining" or catch-all class, which includes many different elements: courtyards, fields, parks, forests, etc. This effect is visible in the weak specificity of its figuration, and in the rather strong correlation with practically all the other classes, except the road network. Its good performance for the corpus world could thus be partly explained by the poorer performance of the other classes. It benefits from uncertainty and a catch-all effect that boosts its recall. Indeed, by observing the confusion matrix, one can see that this class is often confused. A third (0.33) of the water surfaces and nearly a quarter (0.23) of the block surfaces are wrongly classified as non-built, while the non-built class itself does not confuse these classes (0.03, respectively 0.12). And since this class is superficially more present than the others, this does not strongly impact its precision. The catch-all effect plead in favor of dividing this class into several sub-classes in the future. The ontology adopted does not seem to be sufficient for the world. One could for instance imagine dividing this class into courtyards, woods, fields and arable land, and meadows and wasteland. Alternatively, in the short term, one could modify the weights of cross entropy, which is the loss used by the neural network, to give less importance to the performance of this class. This would undoubtedly amplify the catch-all effect, but would probably increase the performance of the other classes. For Paris, this class obtains lower results, but quite close to those obtained for water, for example. This class is less catch-all, and the balance between confusing and being confused is shifted. In particular, the non-built is abundantly confused with blocks (0.33), even though the figuration of the latter is the least correlated (0.7). Conversely, blocks are also sometimes confused with the non-built (0.17). This confusion is by far the most important reason for this poor performance. Several explanations are possible. Notably, in Paris, this class mainly describes parks, as well as courtyards. Compared to the rest of the world, the non-built class is topologically less differentiated. Some parks adopt forms that are quite similar to building blocks, especially in Paris where the urban form is mixed, with corner buildings and various trapezoids. Moreover, the spatial juxtaposition of inner courtyards and buildings promotes confusion, especially in this direction, since an unrecognized inner courtyard will be classified as a non-built confused with blocks. The detection of inner courtyards seems indeed difficult, as one can see in **Fig. 4.7.4**, where one can also observe some examples of the difficulty to identify and separate the morphologies of blocks and non-built. The Parisian corpus also suffers from poor learning transfer from the world, which may be caused by a relatively large figuration distance between Paris and the world for this class ($r_{pearson} = 0.81$). This seems obvious, considering that the elements represented in one and the other corpus are also semantically quite different, as explained.

The drop in performance is also significant for block class recognition, between the 2+1 classes problem and the 4+1 class problem. This was observed very clearly, both for Paris and for the world, with a more significant drop for the world. It is again suspected that the overperformance in the first problem is due to the catch-all effect, which again manifests itself in the confusion matrices, where this class confounds little, but is confused a lot. 11% of the road network for Paris and 21% for the world are erroneously attributed to blocks. Without any significant impact on precision, again, as this class represents, in the smallest ontology, 65% of the world's surface and 56% of the surface of Paris.

The water class mainly shows a spectacular increase in performance thanks to learning transfer (+18% for Paris, +36% for the World). Its figuration is very convergent, which translates into very high mean $\kappa$ for both Paris (12.85) and the world (15.22). This very homogenous figuration, supported by a high correlation between Paris and the world ($r = 0.88$), probably reinforces the performance transfer by a better recognition of the figuration. The initial performances are moreover probably impacted by the low surface proportion of water in the training corpus (12% for the world, 3% for Paris), which explains that the insights brought by cross pretraining also allows the network to better characterize this class.

Another interesting element is the road network. Despite its difficult morphology, because sometimes linear, it is the second best IoU for the world (0.4682, frame excluded), and the best IoU for Paris (0.7132, idem), for the 4+1 class problem. However, it is difficult to compare this class to the others because morphological and hierarchical decision factors are also taken into account in neural networks and this class is outstanding in this respect. Indeed, it is partially linear, whereas the other classes are rather surface, except for water, which is another case apart. On the other hand, one can compare the world to Paris and observe that the performance is much better for the latter (0.7132) than for the world (0.6538). Here also, the surface representation can be taken into account (12% for the world compared to 21% for Paris), but is not in itself a sufficient explanation. Road network is a class apart, which is very little confused (recall = 0.8832 for Paris, 0.7898 for the world). For the corpus world, it is the least confused class, and interestingly enough, it is also the class that is figuratively the least correlated to the other classes (mean $r = 0.888$). For Paris, the level of confusion is also low. As it does not seem to be mainly caused by figurative differences ($r_{Paris\_world} = 0.88$), it is likely that this discrepancy in performance is rather driven by topological reasons, and a contribution of the extremely idiosyncratic morphology of roads. Actually, several traditional map processing studies have already tackled the problem of road segmentation using these topological criteria [7], [29], [45]. This explanation could also justify the discrepancy between Paris and the World. Indeed, being geographically homogeneous, the Parisian corpus has a characteristic urban form, which could facilitate the identification of roads, and potentially other elements, and justify the difference in performance with the World corpus.

These conclusions are reinforced when looking at the results of the experiment on the impact of texture and color (**Fig. 4.13.2**). It can be seen that by removing the major figuration features of color, value and texture, performance drops only slightly, by about 10%. This doesn't necessarily mean that color and texture are only marginally used in normal times by the neural network. In fact, it means that when these visual cues are removed, the network is likely to be able to refer to morphological and topological criteria, and to hierarchize information.

This conclusion is also reinforced by our findings above. For example, even if the important learning transfer for water seems to be supported by figurative criteria, the low performance in absolute values, in addition to the low surface representation of water, can also be explained by a greater morphological and topological diversity for this class, a fortiori for the world. Indeed, water can take the form of a narrow river, a wide river, but also a rectangular basin, a small lake, a coast, a port, etc. This diversity probably makes this class more difficult to characterize according to morphological criteria and seems to impact its absolute recognition performance, especially for the world corpus. Moreover, as suggested above, this explanation justifies the frequent confusions between blocks and non-built, which essentially have a similar morphological and topological identity. Both are in fact areal classes surrounded by roads.

The arguments discussed above do not pretend to be exhaustive because neural networks have many as yet unknown springs. However, they do shed some light on a few points. We consider that the framework of the maps is ideal for this discussion on the ability of neural networks to combine figurative and topological clues and to open up avenues of understanding on their performance. Indeed, maps have a "textbook" figuration of computer vision, with basic textures, such as hatching, some colors, and partly geometric morphologies, squares, rectangles, trapezoids. They are therefore easier to characterize from this point of view. These conditions are met in very few fields of application of deep learning. In addition, for semantic segmentation, there is a very specific and comprehensible metric, the mIoU, and the ability to visualize the results in order to understand the errors. This field therefore brings together all the elements that can help to make progress on these questions and better understand the performance of neural networks.

Speaking of questions, we now finally have the keys in hand to answer part of the third research sub-question: How robust are trained neural networks when facing high figurative variance? In view of the above, it would seem that neural networks have a capacity to abstract and hierarchize information that allows them to cope with high figurative variance. It is difficult to justify differences in performance solely on figurative criteria, as it seems that many more important elements are taken into account, including the completeness of the ontology and how it fits the two corpora, differences in class representation in the training set, as well as many morphological and topological criteria. We noticed that, in the end, the absolute figurative diversity of the two corpora was not so different, despite the difference in cultural diversity. On the other hand, the cultural factors are more pronounced than could have been predicted, which is expressed in particular by substantial differences in performance according to regions and the urban form of the cities. In conclusion, neural networks can be considered as very promising tools for map processing research, due to their capacity to go beyond, without excluding, traditional figurative features. On the other hand, a cultural limitation appears, since it seems that some map grammars are difficult to reconcile, and some cultural conventions are very difficult to analyze for networks trained on generic corpora (see examples in **Fig. 4.8.1**). However, for more specific corpora, one could envisage models trained separately, as is done for example in optical character recognition (OCR), where handwriting must be recognized by specific networks. The figurative robustness of the networks is also demonstrated by the learning transfer performances, which allow, as we have shown in the experiment on the size of the training corpus, to obtain excellent performances with only a very small number of training examples, once the network has been pre-trained on a larger generic corpus. In this respect, the generic training sets that have been created as part of this work could be very useful for further research.

## 5.2 Avenues of improvement

The fourth and final research sub-question is more open-ended than the previous ones. It questions the avenues for further increasing the performance of neural networks: 4) What avenues might increase their representational flexibility? How can the composition of training set contribute?

The answers to this question are myriad and we shall confine ourselves to presenting a few leads. First of all, in view of the previous conclusions, it seems fair to mention that the increase in representational flexibility is not simply an increase in figurative flexibility, but it should also allow a better understanding of the different morphologies, typically due to differences in urban architecture. To improve the understanding of topology, size agnosticism would be desirable. This could be brought by a PSPNet-inspired architecture, if the latter can be improved for better results, or alternatively by some size-related data augmentation.

We already presented our experiments concerning active learning learning, which were very close to this fourth research sub-question. To briefly recall the principle, active learning proposes to feed the network training with especially difficult examples, which should in theory allow to increase this famous representational flexibility, and to increase performances. The results of this experiment were disappointing, and the opposite experiment (presenting the network with easier examples) was also proved to be disappointing. This path of active learning therefore seems unpromising. However, the use of other selection metrics, and the extension of the patch choice for the network could lead to better results.

In a similar way to active learning, curriculum learning proposes to transpose concepts of human pedagogy to neural networks. As a reminder, the idea of curriculum learning is to train the network with examples of increasing difficulty, starting with the simplest ones. In the proof of concept we implemented, the samples were simply sorted by mIoU. The results proved to be disappointing. However, other methodologies could be tested. For example, by first training the network on maps of cities with a regular urban shape, and then increasing its representational flexibility by adding cities with a less regular morphology. Other methodologies applied to related problems, such as Zhang et al's [157] curriculum learning for street view segmentation, could also lead to improvements.

Furthermore, we tried to use insights from a close and much more advanced field, namely the segmentation of satellite images. In an experiment that we will not detail in this work, we implemented an edge loss reinforced semantic segmentation network, as described by Liu et al [52]. This paper proposed to add an intermediate loss to the semantic loss, comparing the contours of the ground-truth with the edges extracted in the intermediate deep layers of the encoder, respectively the decoder. Thus, such a method could theoretically allow to emphasize morphology and pattern recognition, and thus make the network more independent from other figuration features. However, the results obtained have been rather poor. It therefore seems that this track cannot be easily transposed to maps, despite the edges being very sharp in map data compared to aerial images.

In neurology, modulation is a key concept. Modulatory neurons are responsible for activating or inhibiting their neighbours to facilitate or prevent the onset of an electrical signal, or action potential. Vecoven et al [158] therefore proposes to use neuro-modulated networks (NMNs), in which a neuromodulatory network analyses the context and then acts on the main network via a $z$ factor that multiplies the weights. Vecoven et al have demonstrated a better adaptation to changing contextual constraints, on three different navigation problems. This field is still very experimental and the applications could be very beneficial to map processing problems, where context is obviously very important. As the implementation of Vecoven's NMNs in Pytorch would be very time-consuming and, for CNNs, the resulting network would probably be huge, we have tried to implement a simplified version, which we will not detail extensively here, in which each layer from the encoder was multiplied by a $z$ factor before being added in the decoder. The $z$-factors were calculated by a minimal VGG encoder in charge of neuromodulation. However, the results of this experiment were not conclusive and further basic research is probably necessary before these concepts can be applied to CNNs. In any case, contextual neuromodulation could theoretically allow to significantly increase the representational flexibility of the networks and remains an avenue to be monitored in the future.

# 6 Conclusion

In this work we tried to answer the following core question, which is divided into 4 research sub-questions:

1) **How do figuration impact CNNs performance when they are used for semantic segmentation of historical city maps?**
    i. How to measure the figurative diversity in a map corpus?
    ii. How does the representation of specific map elements, such as buildings, streets, water, or crops vary?
    iii. How robust are trained neural networks when facing high figurative variance?
    iv. What avenues might increase their representational flexibility? How can the composition of training set contribute?

To this end, we compared more than 1500 maps of Paris with a corpus of 256 historical maps of cities around the world which was set up for this work, based on literature and databases of heritage institutions. A 5-classes annotation ontology, compatible with our previous works was created and extended to 330 patches of Paris that were already annotated with a minimal ontology. Another 305 additional samples from the world were also fully annotated. In total, we thus built a library of 635 training examples.

In a second step, we based ourselves on existing methods, which we adapted and improved in order to operationalize the map figuration. We created a metric that allows to measure the acuity of multimodal distributions, and thus provide an index for measuring the figurative convergence or diversity within a corpus. This allowed to address the first subquestion. The diversity of both corpora, Paris and the world, was compared to a map used in a related research and it was found that the diversity found in both our corpora was considerably greater. Thanks to this method and the annotated labels, we were also able to characterize specifically the figuration of the main elements depicted on maps, such as water, city blocks, and road network, and thus answer the second subquestion.

Thirdly, we adapted a fully convolutional neural network designed for semantic segmentation to historical city maps. We made several specific improvements to this model and have succeeded in presenting a new performance benchmark, which far surpasses previously established records. At a maximum, we achieved a mean intersection over union (mIoU) of 0.8905 for Paris and 0.8055 for the World, on a 3-classes problem with frame presegmentation. On a 5-classes problem, the best mIoU is 0.6363 for Paris, and 0.5595 for the World. This helped to answer the third sub-question. Indeed, it seems that neural networks are extremely robust in the face of figurative diversity, even if some map grammars seem more difficult to segment. This high performance is likely due to the fact that neural networks are able to integrate many concepts, such as presumably morphology, topology, and semantic hierarchy, to supplement figurative features, without excluding the latter.

Fourthly, we have proposed a new method for estimating predictive confidence, based on neural networks themselves, and capable of predicting potential performance on patches that have not participated in learning.

Finally, in order to answer the fourth and last research subquestion, we explored innovative and more traditional methods and evaluated future development paths for improving the performance and representational flexibility of neural networks, in the context of CNN-based map semantic segmentation. Among the avenues we have explored are the increase in the number of training samples, active learning, curriculum learning, neuromodulation, and technology transfer from related fields, such as the field of aerial image segmentation.

# Annex 1: Main figures and tables

**Table 2.4.3** Proportion of areas covered in the different corpora and sets by the 5 annotated classes.

| Corpus | Set | Frame | Road netw. | Blocks | Water | Non-built |
|---|---|---|---|---|---|---|
| World | train | 0.2219 | 0.1226 | 0.1749 | 0.1212 | 0.3594 |
| World | eval | 0.1851 | 0.1140 | 0.1783 | 0.0728 | 0.4499 |
| Paris | train | 0.2358 | 0.2079 | 0.3381 | 0.0262 | 0.1919 |
| Paris | eval | 0.1659 | 0.2036 | 0.3601 | 0.0448 | 0.2256 |

**Table 2.4.4** Proportion of areas covered in the different corpora and sets, in the 3-classes ontology.

| Corpus | Set | Frame | Road netw. | Map content |
|---|---|---|---|---|
| World | train | 0.2258 | 0.1234 | 0.6507 |
| World | eval | 0.1806 | 0.2041 | 0.7009 |
| Paris | train | 0.2358 | 0.2056 | 0.5586 |
| Paris | eval | 0.1659 | 0.2014 | 0.6328 |

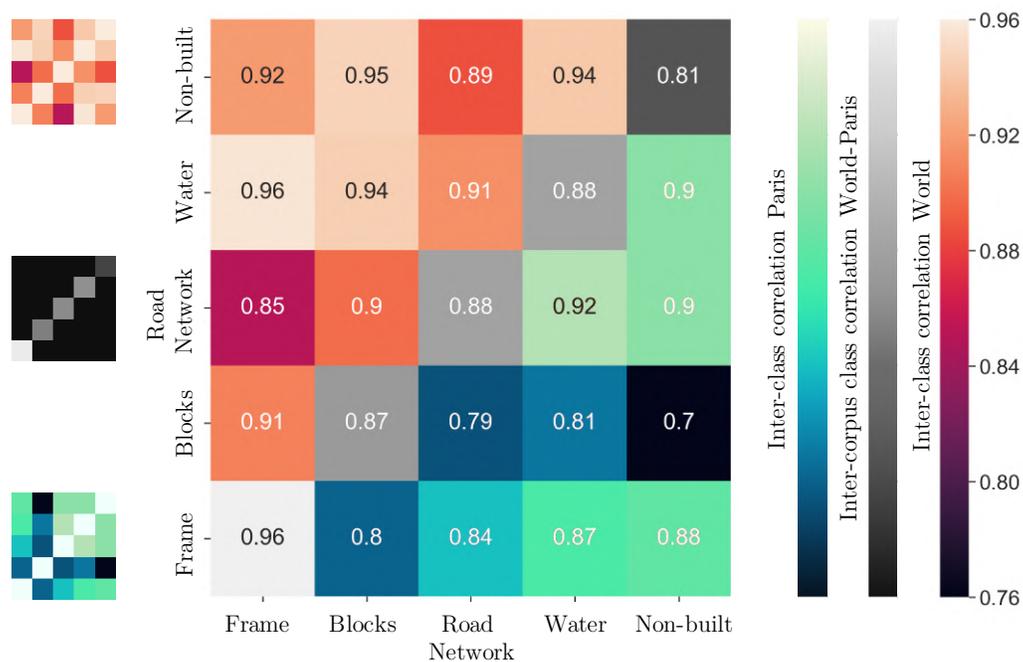

**Figure 3.4.1** Aggregated correlation heatmap matrix of intra-corpus inter-class correlation and inter-corpus class correlation. $r_{pearson} \in [-1, 1]$. All correlations are significant ($p_{value} \ll 0.05$)

**Table 3.4.2** Mean intra-corpus inter-class correlation

| Corpus | **Mean** | Frame | Blocks | Road netw. | Water | Non-built |
|---|---|---|---|---|---|---|
| World | **0.917** | 0.908 | 0.924 | 0.888 | 0.939 | 0.924 |
| Paris | **0.842** | 0.846 | 0.775 | 0.862 | 0.877 | 0.847 |

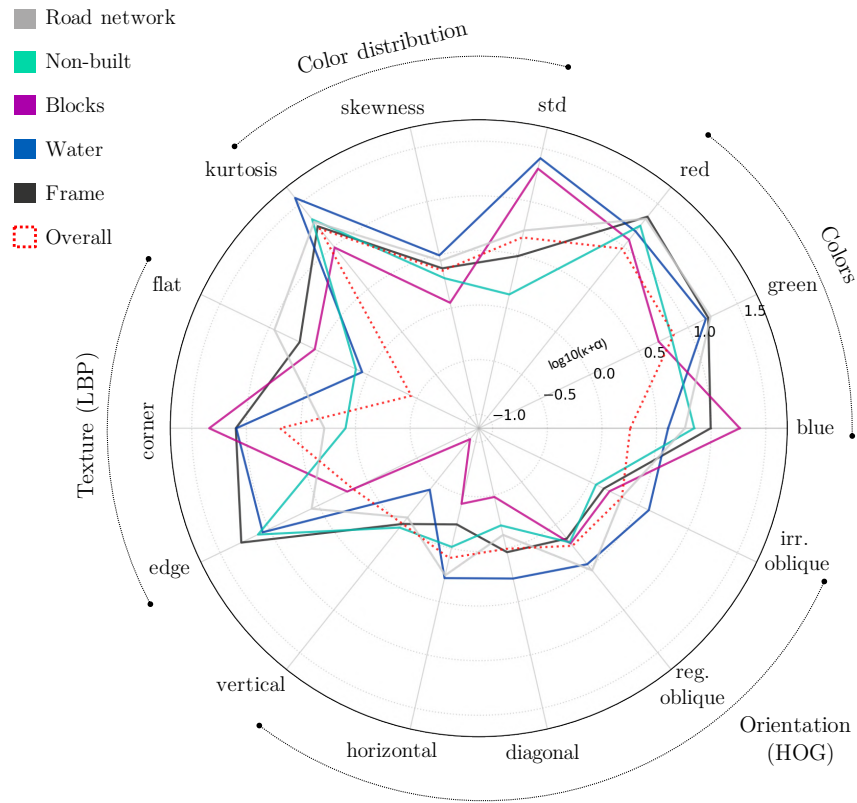

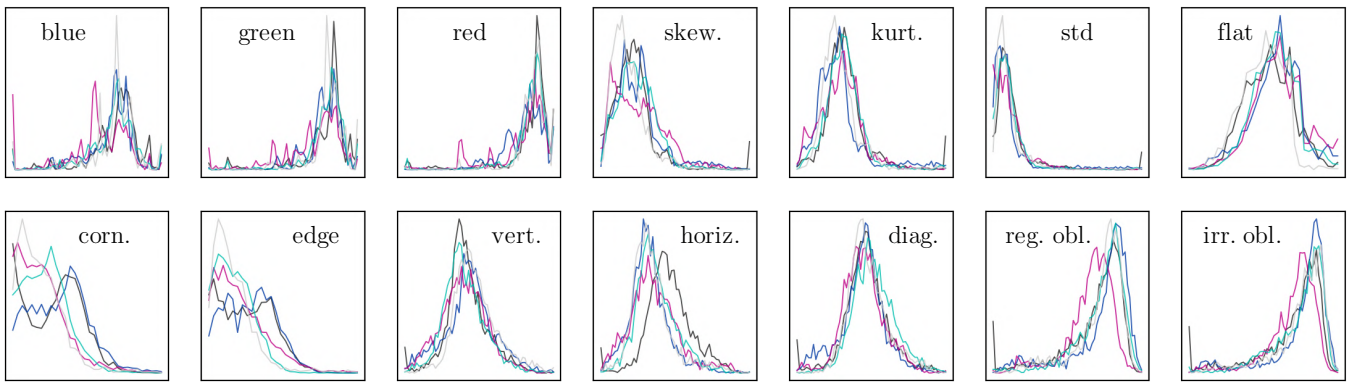

**Fig. 3.3.3** Radial kurtograph of figuration, comparing the $\kappa$-coefficient (see Eq. 1) of the visual features' distributions between the classes of the world corpus. $\alpha \ni \kappa + \alpha \geq 0.1 \; \forall \; \kappa$

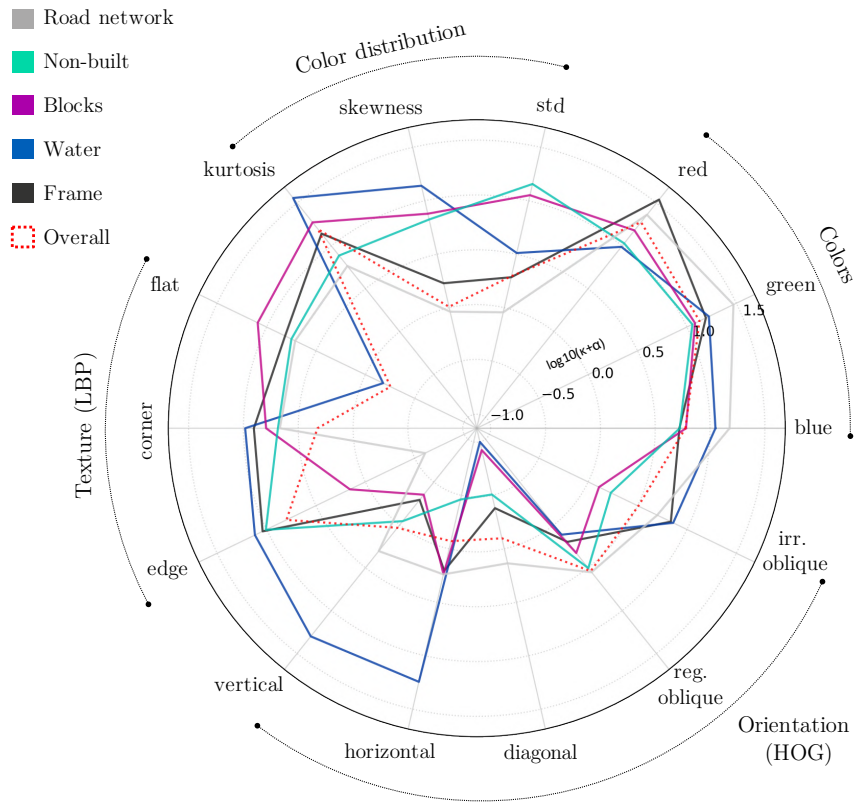
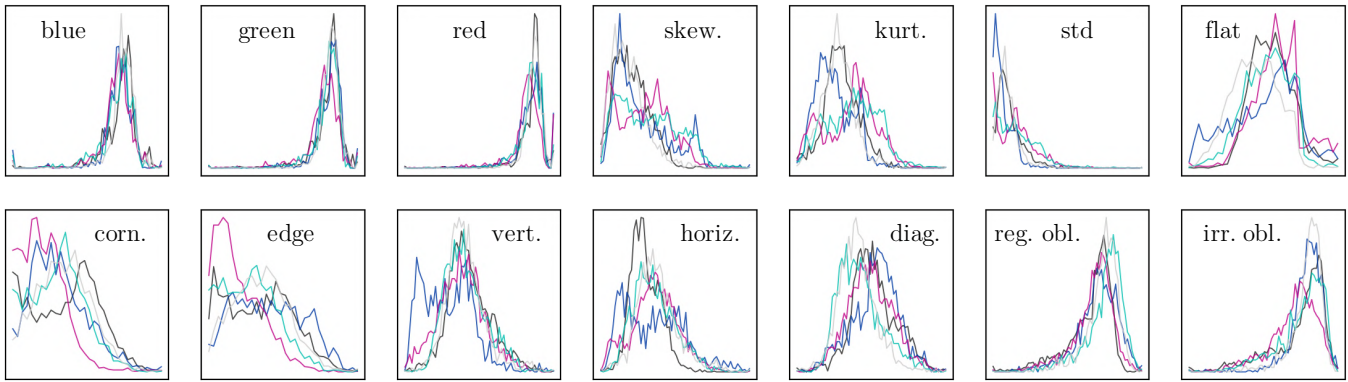

**Fig. 3.3.4** Radial kurtograph of figuration, comparing the $\kappa$-coefficient (see Eq. 1) of the visual features' distributions between the classes of the Paris corpus. $\alpha \ni \kappa + \alpha \geq 0.1 \; \forall \; \kappa$

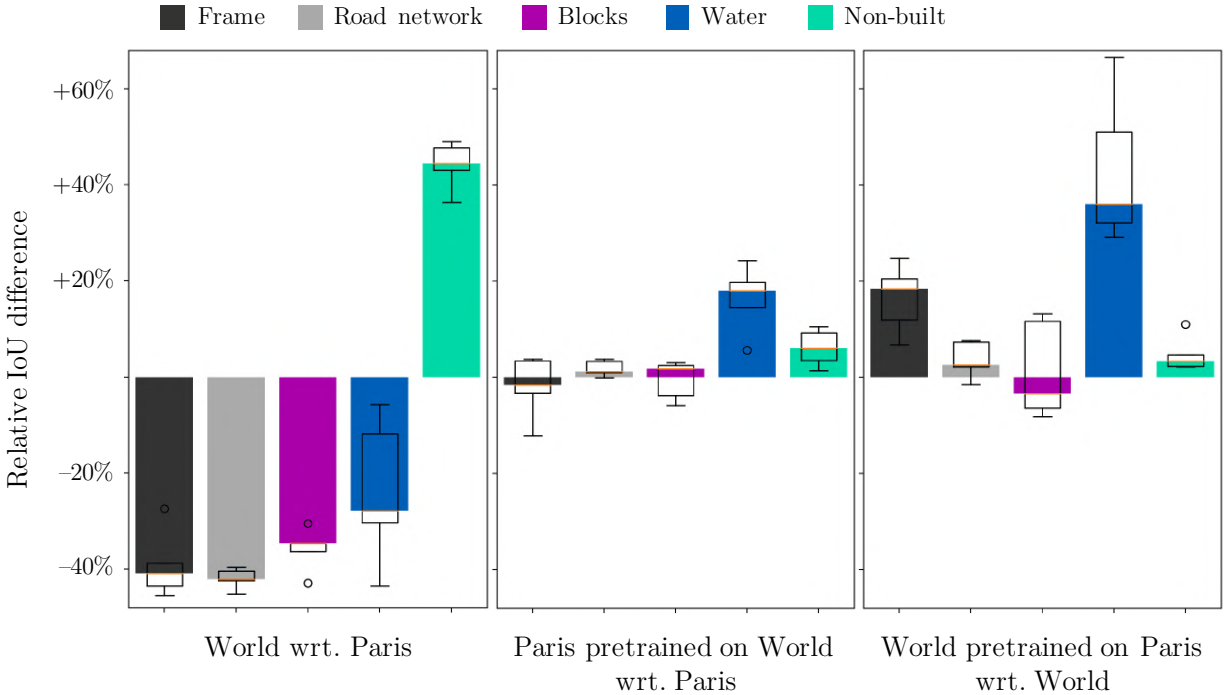

**Figure 4.6.1** Per-class relative performance of regular training between World and Paris corpora (left), and per-class relative performance of transfer learning between World and Paris corpora (middle and right). The relative IoU is computed with regard to the median of the reference IoU. When the pretraining is not specified, the CNN is generically pretrained on ImageNet. Each experiment is repeated 5 times.

**Table 4.7.1** Performance achieved, per class and classes mean, on the two datasets, for 2+1 and 4+1 classes problems

| Metric | Class | Paris 2+1 | World 2+1 | Paris 4+1 | World 4+1 |
|---|---|---|---|---|---|
| IoU | **Mean** | **0.8905** | **0.8055** | **0.6363** | **0.5595** |
| | Frame | 0.9953 | 0.9924 | 0.9810 | 0.9881 |
| | Blocks | 0.9181 | 0.9114 | 0.5657 | 0.3559 |
| | Road Netw. | 0.7580 | 0.5147 | 0.7132 | 0.4682 |
| | Water | – | – | 0.4682 | 0.3318 |
| | Non-built | – | – | 0.4235 | 0.6538 |
| Accuracy | **Mean** | **0.9624** | **0.9556** | **0.8392** | **0.7986** |
| | Frame | 0.9992 | 0.9985 | 0.9909 | 0.9964 |
| | Blocks | 0.9438 | 0.9336 | 0.7495 | 0.8173 |
| | Road Netw. | 0.9441 | 0.9348 | 0.8982 | 0.8211 |
| | Water | – | – | 0.9452 | 0.6471 |
| | Non-built | – | – | 0.6123 | 0.7113 |
| Precision | **Mean** | **0.9292** | **0.8544** | **0.7292** | **0.6986** |

|        | Frame     | 0.9959 | 0.9935 | 0.9838 | 0.9893 |
|--------|-----------|--------|--------|--------|--------|
|        | Blocks    | 0.9689 | 0.9730 | 0.6872 | 0.4353 |
|        | Road Netw.| 0.8229 | 0.5967 | 0.7874 | 0.5348 |
|        | Water     | –      | –      | 0.4856 | 0.7098 |
|        | Non-built | –      | –      | 0.7017 | 0.8240 |
| Recall | **Mean**  | **0.9456** | **0.9062** | **0.8175** | **0.7187** |
|        | Frame     | 0.9996 | 0.9989 | 0.9971 | 0.9988 |
|        | Blocks    | 0.9448 | 0.9350 | 0.7618 | 0.6611 |
|        | Road Netw.| 0.8924 | 0.7848 | 0.8832 | 0.7898 |
|        | Water     | –      | –      | 0.9288 | 0.3839 |
|        | Non-built | –      | –      | 0.5165 | 0.7599 |

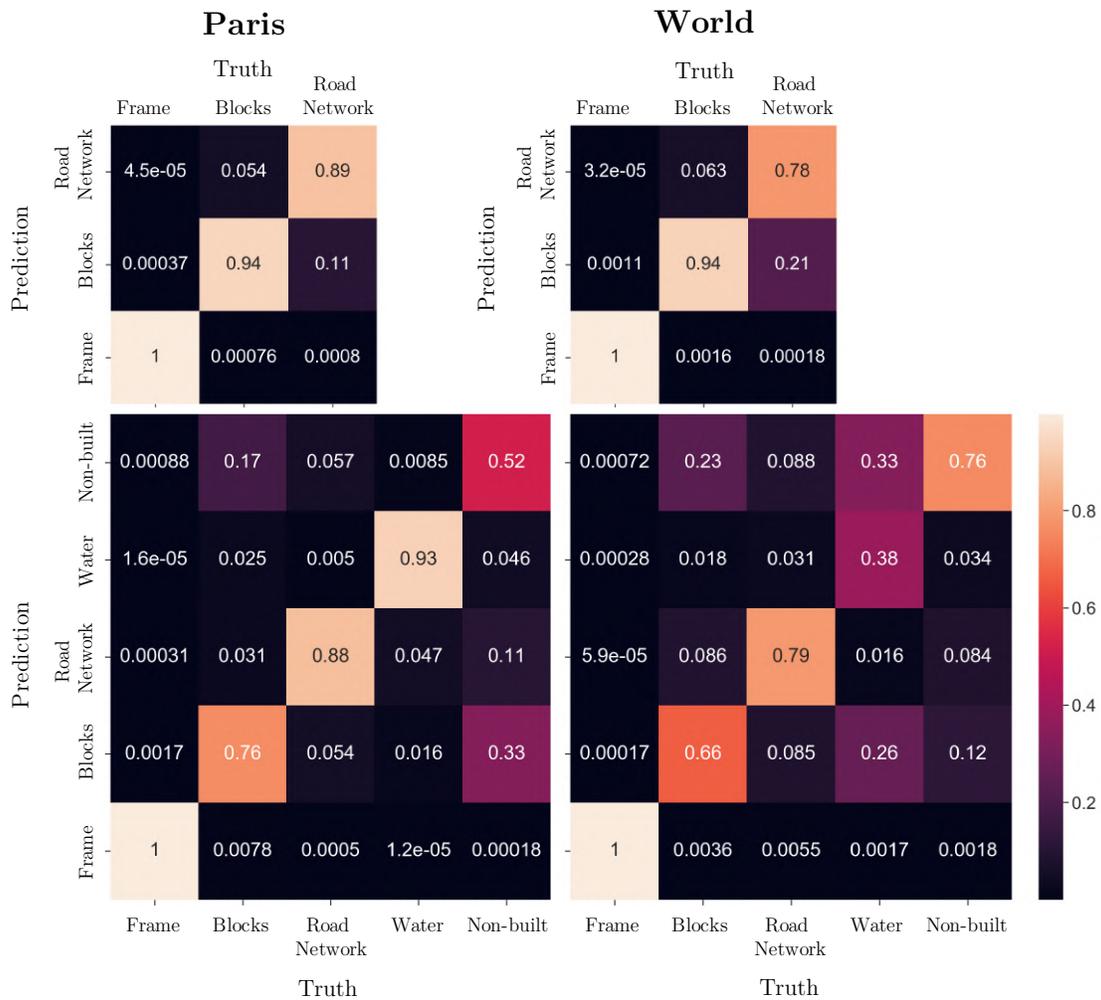

**Figure 4.7.2** Confusion matrix, normalized according to ground-truth, i.e TP + FP (true positives + false positives). The diagonal correspond to recall.

**Table 4.7.3** Achieved performance, per class and classes mean, on the best half (mIoU > median(mIoU)) of the two datasets, for 2+1 classes problems

| Metric | Class | Paris 2+1 | World 2+1 |
|---|---|---|---|
| Precision | **Mean** | **0.9679** | **0.9205** |
| | Frame | 0.9969 | 0.9935 |
| | Blocks | 0.9774 | 0.9816 |
| | Road Netw. | 0.9295 | 0.7863 |
| Recall | **Mean** | **0.9456** | **0.9445** |
| | Frame | 0.9990 | 0.9992 |
| | Blocks | 0.9736 | 0.9957 |
| | Road Netw. | 0.9368 | 0.8686 |

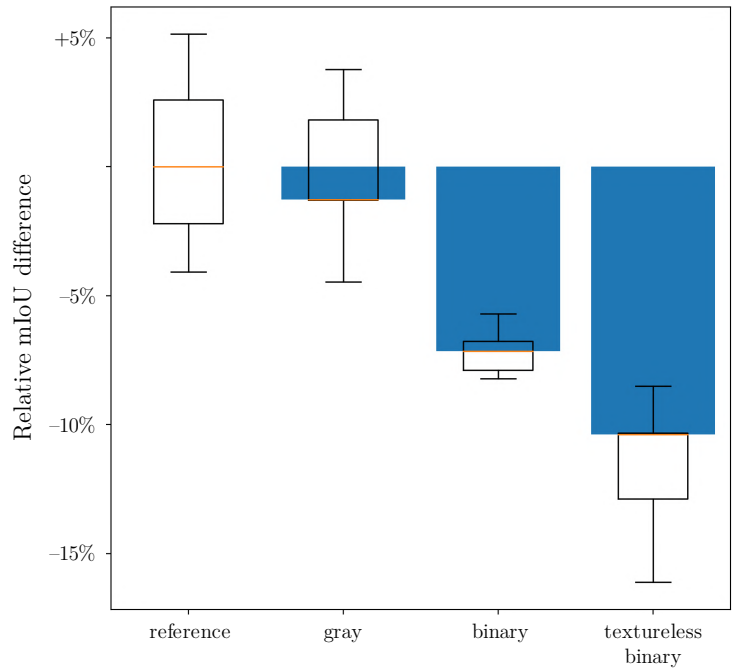

**Figure 4.13.2** Relative loss in performance (mIoU) due to the removal of visual characteristics. Each experiment is repeated 5 times. The median reference IoU is 0.7515